%% file: main.tex
\documentclass[letterpaper, 10 pt, journal, twoside]{ieeetran} 

\usepackage[USenglish]{babel}
\usepackage{graphicx} 
\graphicspath{{./images/}{./plots/}} 
\usepackage[usenames,svgnames,table,rgb,dvipsnames]{xcolor} 
\usepackage[]{mathtools} 
\mathtoolsset{showonlyrefs} 
\usepackage[labelformat=parens]{subcaption} 
\usepackage[hang]{footmisc}
\usepackage[export]{adjustbox} 
\usepackage{wrapfig}
\usepackage{tabularx}
\usepackage{colortbl}
\usepackage{booktabs}
\usepackage{multirow}
\usepackage{leftidx} 
\usepackage{csquotes}
\usepackage{pgf}
\let\pgfimageWithoutPath\pgfimage 
\renewcommand{\pgfimage}[2][]{\pgfimageWithoutPath[#1]{plots/#2}}

\usepackage[lined,linesnumbered,ruled]{algorithm2e}

\usepackage[numbers]{natbib}

\usepackage[nolist,nohyperlinks]{acronym}
\input{fMRT_acronyms}
\input{fMRT_macros}

\usepackage{xr}
\usepackage[hidelinks,bookmarks=true,breaklinks]{hyperref} 

\title{Data-efficient Domain Randomization\\with Bayesian Optimization}
\date{}
\author{Fabio Muratore and Christian Eilers and Michael Gienger and Jan Peters
\thanks{Fabio~Muratore, Christian~Eilers and Jan~Peters are with the Intelligent Autonomous Systems Group, Technical University Darmstadt, Germany.}%
\thanks{Fabio~Muratore, Christian~Eilers and Michael~Gienger are with the Honda Research Institute Eustring, Offenbach am Main, Germany.}%
\thanks{Correspondence to fabio@robot-learning.de}%
}

\begin{document}

\maketitle


\begin{abstract}
When learning policies for robot control, the required real-world data is typically prohibitively expensive to acquire, so learning in simulation is a popular strategy.
Unfortunately, such polices are often not transferable to the real world due to a mismatch between the simulation and reality, called `reality gap'.
Domain randomization methods tackle this problem by randomizing the physics simulator (source domain) during training according to a distribution over domain parameters in order to obtain more robust policies that are able to overcome the reality gap.
Most domain randomization approaches sample the domain parameters from a fixed distribution.
This solution is suboptimal in the context of \simtoreal transferability, since it yields policies that have been trained without explicitly optimizing for the reward on the real system (target domain). Additionally, a fixed distribution assumes there is prior knowledge about the uncertainty over the domain parameters.
In this paper, we propose \acf{BayRn}, a black-box \simtoreal algorithm that solves tasks efficiently by adapting the domain parameter distribution during learning given sparse data from the real-world target domain.
\ac{BayRn} uses Bayesian optimization to search the space of source domain distribution parameters such that this leads to a policy which maximizes the real-word objective, allowing for adaptive distributions during policy optimization.
We experimentally validate the proposed approach in \simtosim as well as in \simtoreal experiments, comparing against three baseline methods on two robotic tasks. Our results show that \ac{BayRn} is able to perform \simtoreal transfer, while significantly reducing the required prior knowledge.
\end{abstract}

\IEEEpeerreviewmaketitle

\section{Introduction}
Physics simulations provide a possibility of generating vast amounts of diverse data at a low cost.
However, sample-based optimization has been known to be optimistically biased~\cite{Hobbs_Hepenstal_1989}, which means that the found solution appears to be better than it actually is.
The problem is worsened when the data used for optimization does not originate from the same environment, also called domain. In this case, we observe a simulation optimization bias, which leads to an overestimation of the policy's performance~\cite{Muratore_etal_2019}.
Generally, there are two ways to overcome the gap between simulation and reality. One can improve the generative model to closely match the reality, \eg by using system identification. Increasing the model's accuracy has the advantage of leading to controllers with potentially higher performance, since the learner can focus on a single domain.
On the downside, this goes in line with a reduced transferability of the found policy, which is caused by the previously mentioned optimistic bias, and aggravated if the model does not include all physical phenomena.
\begin{figure}[t]
	\centering
	\includegraphics[height=3.cm,keepaspectratio]{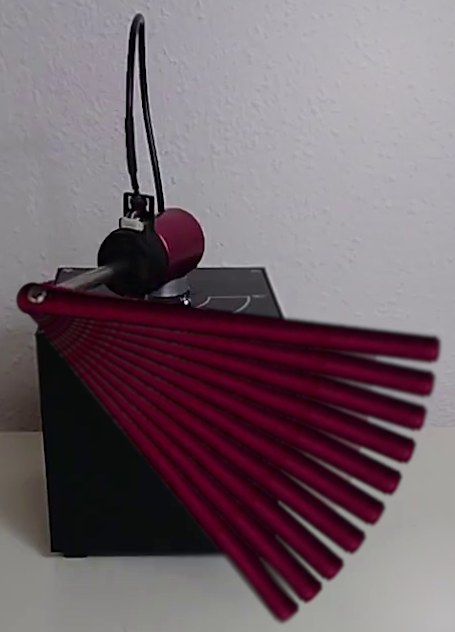}
	\hspace{0.2cm}
	\includegraphics[height=3.cm,keepaspectratio]{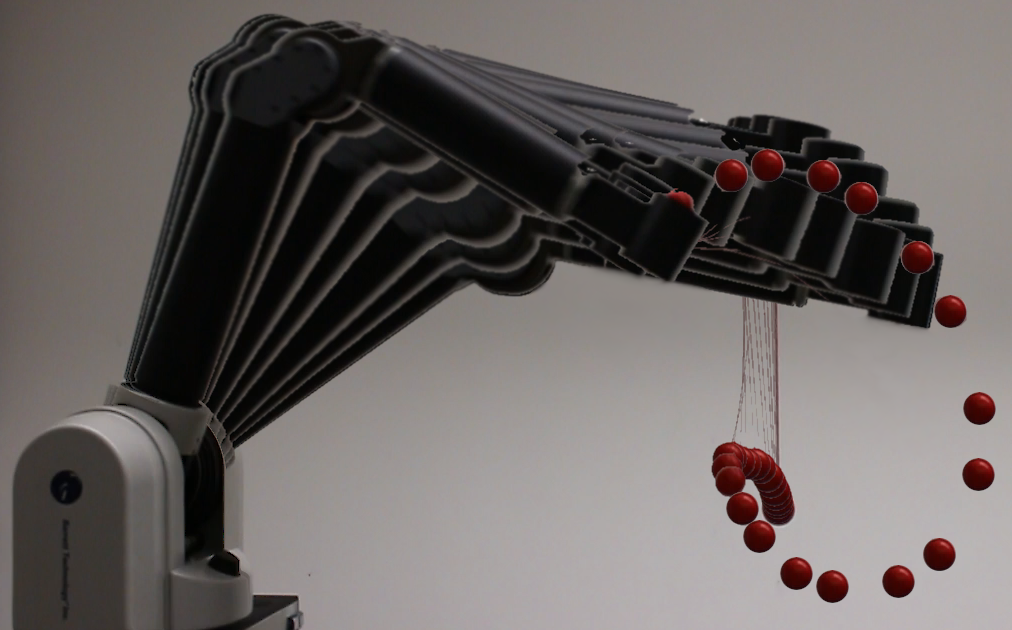}
	\caption{%
		Evaluation platforms: (left) underactuated \sub task on the Quanser Furuta pendulum, (right) \bic task on the Barrett WAM robotic arm.
	}
	\label{fig_QQ_WAM}
	\vspace{-\baselineskip}
\end{figure}
Moreover, we might face a situation where it is not affordable to improve the model.
Alternatively, one can add variability to the generative model, \eg by turning the physics simulator's parameters into random variables. Learning from randomized simulations poses a harder problem for the learner due to the additional variability of the observed data.
But the recent successes in the field of \simtoreal transfer argue for domain randomization being a promising method~\cite{OpenAI_etal_2018,Chebotar_etal_2019}.

\Sota approaches commonly randomize the simulator according to a static handcrafted distribution~\cite{Mordatch_etal_2015,Lowrey_etal_2018,Peng_etal_2018,Tobin_etal_2018}.
Even though static randomization is in many cases sufficient to cross the reality gap, it is desirable to automate the process as far as possible.
Moreover, using a fixed distribution does not allow to update the prior knowledge or incorporate the uncertainty over domain parameters.
Most importantly, closing the feedback loop over the real system will lead to policies with higher performance on the target domain since the feedback enables the optimization of the domain parameter distribution.
However, approaches which adapt an distribution over simulators, yield to additional challenges. For example algorithms that intertwine system identification and policy optimization, \eg,~\cite{Chebotar_etal_2019,Yu_etal_2019b}, introduce a circular dependency since both subroutines depend on the sensible outputs of the other. One possible failure case is a policy which does not excite the system well enough, resulting in bad updates the simulator's parameters.
The \simtoreal algorithm presented in this paper does not require any system identification.

\paragraph*{Contributions} We advance the \sota by introducing \acf{BayRn}, a method which is able to close the reality gap by learning from randomized simulations and adapting the distribution over simulator parameters based solely on real-world returns. The use usage of \ac{BO} for sampling the next training environment (source domain) makes \ac{BayRn} sample efficient \wrt real-world data.
The proposed algorithm can be seen as a way to vastly automate the finding of a source domain distribution in \simtoreal settings, which is typically done by trial and error.
We validate our approach by conducting a \simtosim as well as two \simtoreal experiments on an underactuated nonlinear swing-up task, and on a \bic task (Figure~\ref{fig_QQ_WAM}).
The \simtosim setup examines the domain parameter adaptation mechanism of \ac{BayRn}, and shows that the belief about the domain distribution parameters converges to a specified ground truth parameter set.
In the \simtoreal experiments, we compare the performance of policies trained with \ac{BayRn} against multiple baselines based on a total number of 700 real-world rollouts. 
Moreover, we demonstrate that \ac{BayRn} is able to work with step-based as well as episodic \ac{RL} algorithms as policy optimization subroutines.

The remainder of this paper is organized as follows: first, we introduce the necessary fundamentals (Section~\ref{sec_adr_prob_statement}) for \ac{BayRn} (Section~\ref{sec_adr_BayRn}). Next, we evaluate the devised method experimentally (Section~\ref{sec_adr_experiments}). Subsequently, we put \ac{BayRn} into context with the related work (Section~\ref{sec_adr_related_work}). Finally, we conclude and mention possible future research directions (Section~\ref{sec_adr_conclusion}).

\section{Background and Notation}
\label{sec_adr_prob_statement}
Optimizing control policies for \acp{MDP} with unknown dynamics is generally a hard problem (Section~\ref{sec_adr_MDP}).
It is specifically hard due to the simulation optimization bias~\cite{Muratore_etal_2019}, which occurs when transferring the polices learned in one domain to another.
Adapting the source domain based on real-world data requires a method suited for expensive objective function evaluations. \acs{BO} is a prominent choice for these kind of problems (Section~\ref{sec_adr_BO}).

\subsection{\acl{MDP}}
\label{sec_adr_MDP}
Consider a time-discrete dynamical system
\begin{equation}
\begin{gathered}
\label{eq_sys_dyn}
\fs_{t+1} \sim \transprob[\domparams]\left( \given{\fs_{t+1}}{\fs_t, \fa_t, \domparams} \right), \quad
\fs_0 \sim \initstatedistr[,\domparams]( \given{\fs_0}{\domparams}), \\
\fa_t \sim \pol{\given{\fa_t}{\fs_t; \polparams}}, \quad
\domparams \sim \domparamdistr{\domparams; \domdistrparams},
\end{gathered}
\end{equation}
with the continuous state ${\fs_t \in \stateset[\domparams] \subseteq \RR^{\dimstate}}$, and continuous action ${\fa_t \in \actionset[\domparams] \subseteq \RR^{\dimact}}$ \mbox{at time step $t$}. The environment, also called domain, is instantiated through its parameters ${\domparams \in \RR^{\dimdomparam}}$ (\eg, masses, friction coefficients, or time delays), which are assumed to be random variables distributed according to the probability distribution ${\domparamdistrsym \colon \RR^{\dimdomparam} \to \RR^{+}}$ parametrized by $\domdistrparams$.
These parameters determine the transition probability density function ${\transprob[\domparams] \colon \stateset[\domparams] \times \actionset[\domparams] \times \stateset[\domparams] \to \RR^{+}}$ that describes the system's stochastic dynamics.
The initial state $\fs_0$ is drawn from the start state distribution ${\initstatedistr[,\domparams] \colon \stateset[\domparams] \to \RR^{+}}$.
Together with the reward function ${\rewsym \colon \stateset[\domparams] \times \actionset[\domparams] \to \RR}$, and the temporal discount factor ${\gamma \in [0,1]}$, the system forms a \ac{MDP} described by the set ${\mdp[\domparams] = \set{\stateset[\domparams], \actionset[\domparams], \transprob[\domparams], \initstatedistr[,\domparams], r, \gamma}}$.
The goal of a \acf{RL} agent is to maximize the expected (discounted) return, a numeric scoring function which measures the policy's performance. The expected discounted return of a stochastic domain-independent policy $\pol{\given{\fa_t}{\fs_t; \polparams}}$, characterized by its parameters $\polparams \in \Theta \subseteq \RR^{\dimpolparam}$, is defined as
\begin{equation}
\label{eq_edr}
\edr{\polparams,\domparams,\fs_0} = \EsubBig{\traj\sim p(\traj)}{\sum_{t=0}^{T-1} \gamma^t r(\fs_t, \fa_t) \Big| \polparams,\domparams, \fs_0}.
\end{equation}
While learning from experience, the agent adapts its policy parameters.
The resulting state-action-reward tuples are collected in trajectories, \aka rollouts, $\traj = \set[t=0][T-1]{\fs_t,\fa_t,\rewsym_t}$, with $\rewsym_t = \rewfcn{\fs_t, \fa_t}$.
To keep the notation concise, we omit the dependency on $\fs_0$.

\subsection{\acl{BO} with \aclp{GP}}
\label{sec_adr_BO}
\acf{BO} is a sequential derivative-free global optimization strategy, which tries to optimize an unknown function $f \colon \boinputspace \rightarrow \RR$ on a compact set $\boinputspace$~\cite{Snoek_etal_2012}.
In order to do so, \ac{BO} constructs a probabilistic model, typically a \ac{GP}, for $f$. 
\acp{GP} are distributions over functions
${f \sim \GP{\GPmean}{\GPcovar}}$ defined by a prior mean $\GPmean \colon \boinputspace \rightarrow \RR$ and positive definite covariance function $\GPcovar \colon \boinputspace \times \boinputspace \rightarrow \RR$ called kernel.
This probabilistic model is used to make decisions about where to evaluate the unknown function next.
A distinctive feature of \ac{BO} is to use the complete history of noisy function evaluations $\data = \set[i=0][n]{\fx_i, y_i}$ with $\fx_i \in \boinputspace$ and $y_i \sim \distrnormal{y}{f(\fx_i),\varepsilon}$ where $\varepsilon$ is the variance of the observation noise.
The next evaluation candidate is then chosen by maximizing a so-called acquisition function $\acqfcnsym\colon \boinputspace \rightarrow \RR$, which typically balances exploration and exploitation. Prominent acquisition functions are Expected Improvement and Upper Confidence Bound.
Through the use of priors over functions, \ac{BO} has become a popular choice for sample-efficient optimization of black-box functions that are expensive to evaluate.
Its sample efficiency plays well with the algorithm introduced in this paper where a \ac{GP} models the relation between domain distribution's parameters and the resulting policy's return estimated from real-world rollouts, \ie $\fx \equiv \domdistrparams$ and $y \equiv \estedr[][\text{real}]{\polparamsopt}$.
For further information on \ac{BO} and \acp{GP}, we refer the reader to~\cite{Snoek_etal_2012} as well as~\cite{Rasmussen_Williams_2006}.

\section{\acf{BayRn}}
\label{sec_adr_BayRn}
The problem of source domain adaptation based on returns from the target domain can be expressed in a bilevel formulation
\begin{align}
\domdistrparams\opt &= \argmax_{\domdistrparams \in \domdistrparamspace} \edr[][\text{real}]{\polparamsopt(\domdistrparams)} \quad \text{with} \label{eq_BayRn_upper_problem}\\
\polparamsopt(\domdistrparams) &= \argmax_{\polparams \in \polparamspace} \Esub{\domparams \sim \domparamdistrsym(\domparams; \domdistrparams)}{\edr{\polparams,\domparams}} \label{eq_BayRn_lower_problem},
\end{align}
where we refer to \eqref{eq_BayRn_upper_problem} and \eqref{eq_BayRn_lower_problem} as the upper and lower level optimization problem respectively.
Thus, the two equations state the goal of finding the set of domain distribution parameters $\domdistrparams\opt$ that maximizes the return on the real-world target system $\edr[][\text{real}]{\polparamsopt(\domdistrparams)}$, when used to specify the distribution $\domparamdistr{\domparams; \domdistrparams}$ during training in the source domain.
The space of domain parameter distributions is represented by $\domdistrparamspace$.
In the following, we abbreviate $\polparamsopt(\domdistrparams)$ with $\polparamsopt$.
\begin{algorithm}[t!]
	\caption{\acl{BayRn}} 
	\label{algo_BayRn}
	\input{pseudocode/BayRn.tex}
\end{algorithm}
At the core of \ac{BayRn}, first a policy optimizer, \eg, an \ac{RL} algorithm, is employed to solve the lower level problem \eqref{eq_BayRn_lower_problem} by finding a (locally) optimal policy $\pol{\polparamsopt}$ for the current distribution of stochastic environments.
This policy is evaluated on the real system for $n_\tau$ rollouts, providing an estimate of the return $\estedr[][\text{real}]{\polparamsopt}$.
Next, the upper level problem \eqref{eq_BayRn_upper_problem} is solved using \ac{BO}, yielding a new domain parameter distribution which is used to randomize the simulator.
In this process the relation between the domain distribution's parameters $\domdistrparams$ and the resulting policy's return on the real system $\estedr[][\text{real}]{\polparamsopt}$ is modeled by a \ac{GP}.
The \ac{GP}'s mean and covariance is updated using all recorded inputs $\domdistrparams$ and the corresponding observations $\estedr[][\text{real}]{\polparams[][\optsym]}$.
Finally, \ac{BayRn} terminates when the estimated performance on the target system exceeds $\tholdsucc$ which is the task-specific success threshold. 
Since the \ac{GP} requires at least a few (about 5 to 10) samples to provide a meaningful posterior, \ac{BayRn} has an initialization phase before the loop.
In this phase, $\ninit$ source domains are randomly sampled from $\domdistrparamspace$, and subsequently for each of these domains a policy is trained. After evaluating the $\ninit$ initial policies, the \ac{GP} is fed with the inputs $\domdistrparams_{1:\ninit}$ and the corresponding observations $\estedr[][\text{real}]{\polparams[1:n_{\text{init}}][\optsym]}$.
The complete \ac{BayRn} procedure is summarized in Algorithm~\ref{algo_BayRn}.
In principle, there are no restrictions to the choice of algorithms for solving the two stages \eqref{eq_BayRn_upper_problem} and \eqref{eq_BayRn_lower_problem}.
For training the \ac{GP}, we used the \ac{BO} implementation from BoTorch~\cite{BoTorch_2019} which expects normalized inputs and standardized outputs.
Notably, we decided for the expected improvement acquisition function and a zero-mean \ac{GP} prior with a Mat\'{e}rn 5/2 Kernel.

\paragraph*{Connection to System Identification}
Unlike related methods (Section~\ref{sec_adr_related_work}), \ac{BayRn} does not include a term in the objective function that drives the system parameters to match the observed dynamics. Instead, the \ac{BO} component in \ac{BayRn} is free to adapt the domain distribution parameters $\domdistrparams$ (\eg, mean or standard deviation of a body's mass) while learning in simulation such that the resulting policies perform well in the target domain.
This can be seen as an indirect system identification, since with increasing iteration count the \ac{BO} process will converge to sampling from regions with high real-world return.
There is a connection to control as inference approaches which interpret the cost as a
log-likelihood function under an optimality criterion using a Boltzmann
distribution construct~\cite{Toussaint_2009,Rawlik_etal_2013}.
Regarding \ac{BayRn}, the sequence of sampled domain distribution parameter sets strongly depends on the acquisition function and the complexity of the given problem.
We argue that excluding system identification from the upper level objective \eqref{eq_BayRn_upper_problem} is sensible for the presented \simtoreal algorithm, since it learns from a randomized physics simulator, hence attenuates the benefit of a well-fitted model.

%
%

\section{Experiments}
\label{sec_adr_experiments}
We study \acf{BayRn} on two different platforms:
1) an underactuated rotary inverted pendulum, also known as Furuta pendulum, with the task of swinging up the pendulum pole into an upright position, and
2) the tendon-driven 4-\acs{DoF} robot arm WAM from Barrett, where the agent has to swing a ball into a cup mounted as the end-effector.
First, we set up a simplified \simtosim experiment on the Furuta pendulum to check if the proposed algorithm's belief about the domain distribution parameters converges to a specified set of ground truth values.
Next, we evaluate \ac{BayRn} as well as the baseline methods SimOpt~\cite{Chebotar_etal_2019}, \acf{UDR}, and \acf{PPO}~\cite{Schulman_etal_2017} or \ac{PoWER}~\cite{Kober_Peters_2011} in two \simtoreal experiments.
Additional details on the system description can be found in Appendix~\ref{secapdx_adr_setup}.
Furthermore, an extensive list of the chosen hyper-parameters can be found in Appendix~\ref{secapdx_adr_experiment_hparams}.
A video demonstrating the \simtoreal transfer of the policies learned with \ac{BayRn} can be found at \texttt{\url{www.ias.informatik.tu-darmstadt.de/Team/FabioMuratore}}. 
Moreover, the source code of \ac{BayRn} and the baselines is available at~\cite{Muratore_SimuRLacra}.

\subsection{Experimental Setup}
\label{sec_adr_experimental_setup}
\begin{wrapfigure}[9]{r}{0.48\columnwidth} 
	\vspace*{-0.3em}
	\def\svgwidth{0.48\columnwidth}
	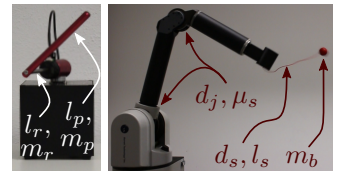
	\caption{%
		Platforms with annotated domain parameters
	}
	\label{fig_QQ_WAM_domain_param}
\end{wrapfigure}
All rollouts on the Furuta pendulum ran for \SI{6}{\second} at \SI{100}{\hertz}, collecting 600 time steps with a reward $\rewsym_t \in ]0, 1]$.
We decided to use a \ac{FNN} policy in combination with \ac{PPO} as policy optimization (sub)routine (Table~\ref{tab_adr_exp_hparams_QQ}).
Before each rollout, the platform was reset automatically. On the physical system, this procedure includes estimating the sensors' offsets as well as running a controller which drives the device to its initial position with the rotary pole centered and the pendulum hanging down. In simulation, the reset function causes the simulator to sample a new set of domain parameters $\domparams$ (Figure~\ref{fig_QQ_WAM_domain_param}).
Due to the underactuated nature of the dynamics, the pendulum has to be swung back and forth to put energy into the system before being able to swing the pendulum up.

The Barrett WAM was operated at $\SI{500}{\hertz}$ with an episode length of $\SI{3.5}{\second}$, \ie, 1750 time steps.
For the ball-in-cup task, we chose a \acs{RBF}-policy commanding desired deltas to the current joint angles and angular velocities, which are passed to the robots feed-forward controller. Hence, the only input to the policy is the normalized time.
At the beginning of each rollout, the robot is driven to an initial position. When evaluating on the physical platform, the ball needs to be manually stabilized in a resting position.
Once the rollout has finished, the operator enters a return value (Appendix~\ref{secapdx_adr_setup}).

\begin{table}[t]
	\centering
	\caption{%
		Range of domain distribution parameter values $\domdistrparams$ used during the experiments. 
		All domain parameters were randomized such that they stayed physically plausible. 
	}
	\label{tab_dom_distr_param_QQ_WAM}
	\begin{subtable}{\columnwidth}
		\centering
		\caption{\sub}
		\label{tab_dom_distr_param_QQ}
		\input{tables/domain_distribution_parameters_QQ.tex}
	\end{subtable}%
	\\
	\vspace{1.5em}
	\begin{subtable}{\columnwidth}
		\centering
		\caption{\bic}
		\label{tab_dom_distr_param_WAM}
		\input{tables/domain_distribution_parameters_WAM.tex}
	\end{subtable}%
\end{table}

In the \simtoreal experiments, we compare \ac{BayRn} to SimOpt, \ac{UDR}, and \ac{PPO} or \ac{PoWER}.
For every algorithm, we train 20 polices and execute 5 evaluation rollouts per policy.
\ac{PPO} as well as \ac{PoWER} are set up to learn from simulations where the domain parameters are given by the platforms' data sheets or \acs{CAD} models. These sets of domain parameters are called nominal.
Hence, \ac{PPO} and \ac{PoWER} serve as a baseline representing step-based and episodic \ac{RL} algorithms without domain randomization or any real-world data.
\ac{UDR} augments an \ac{RL} algorithm, here \ac{PPO} or \ac{PoWER}, and can be seen as the straightforward way of randomizing a simulator, as done in~\cite{Peng_etal_2018}.
Each domain parameter $\domparam$ is assigned to an independent probability distribution, specified by its parameters $\domdistrparams$, \ie mean and variance, (Table~\ref{tab_dom_distr_param_QQ_WAM}).
Thus, we include \ac{UDR} as a baseline method for static domain randomization.
Note that \ac{UDR} can, in contrast to \ac{BayRn} and SimOpt, be easily parallelized which reduces the time to train a policy significantly.
With SimOpt, \citet{Chebotar_etal_2019} presented a trajectory-based framework for closing the reality gap, and validated it on two \sota~\simtoreal robotic manipulation tasks.
SimOpt iteratively adapts the domain parameter distribution's parameters by minimizing discrepancy between observations from the real-world system and the simulation.
While \ac{BayRn} formulates the upper level problem~\eqref{eq_BayRn_upper_problem} solely based on the real-world returns, SimOpt minimizes a linear combination of the L1 and L2 norm between simulated and real trajectories. Moreover, SimOpt employs \ac{REPS}~\cite{Peters_etal_2010} to update the simulator's parameters, hence turning \eqref{eq_BayRn_upper_problem} into an \ac{RL} problem.
The necessity of real-world trajectories renders SimOpt unusable for the \bic task since the feed-forward policy is executed without recording any observations. Thus, there are no real-world trajectories with which to update the simulator.
\ac{BayRn} (Section~\ref{sec_adr_BayRn}), SimOpt and \ac{UDR} randomize the same domain parameters with identical nominal values.
At the beginning of each \simtoreal experiment (Section~\ref{sec_adr_sim2real_results}), the domain distribution parameters $\domdistrparams$ are sampled randomly from their ranges (Table~\ref{tab_dom_distr_param_QQ_WAM}).
The main difference is that \ac{BayRn} and SimOpt adapt the domain distribution parameters, while \ac{UDR} does not.
We chose normal distributions for masses and lengths as well as uniform distributions for parameters related to friction and damping.

\subsection{\Simtosim Results}
\label{sec_adr_sim2sim_results}
\begin{figure}[t]
	\centering
	\vspace{-8pt}
	\input{plots/eval_sim2sim_1row2col.tex}
	\caption{%
		Target domain returns \subref{fig_sim_eval_QQ_mean} and the associated standard deviation \subref{fig_sim_eval_QQ_std} modeled by the \ac{GP} learned with \ac{BayRn} in a \simtosim setting (brighter is higher).
		The ground truth domain parameters as well as the maximum a posteriori domain distribution parameters found by \ac{BayRn} are displayed as a red and orange star, respectively.
		The circles mark the sequence of domain parameter configurations (darker is later).
	}
	\label{fig_sim_eval_QQ}
\end{figure}
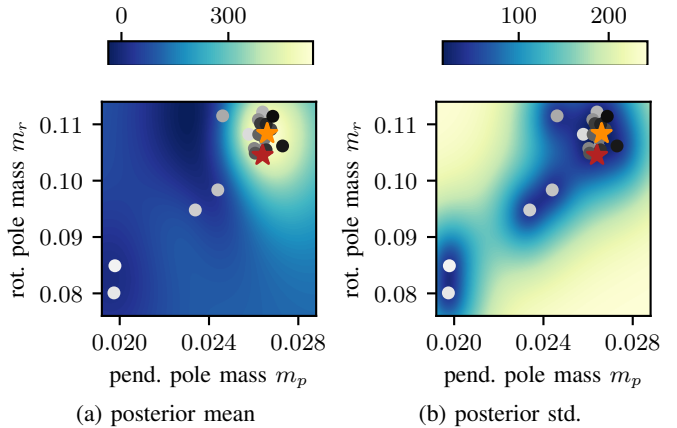

Before applying \ac{BayRn} to a physical system, we examine the domain distribution parameter sampling process of the \ac{BO} component in simulation.
In order to provide a (qualitative) visualization, we chose to only randomize the means of the poles' masses, \ie, $\domdistrparams = [\E{m_r}, \E{m_p}]\tran$.
Thus, for this \simtosim experiment the domain distribution parameters $\domdistrparams$ are synonymous to the domain parameters $\domparams$.
Apart from that, the hyper-parameters used for executing \ac{BayRn} are identical to the ones used in the \simtoreal experiments (Appendix~\ref{secapdx_adr_experiment_hparams}).
\begin{figure*}[t]
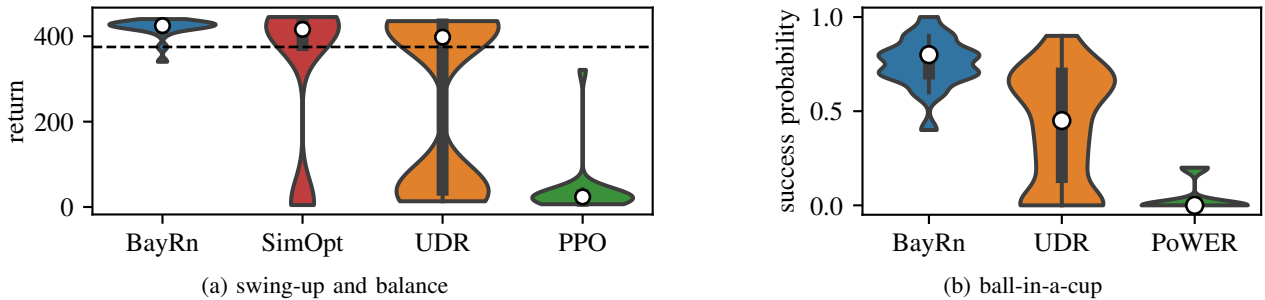

	\centering
	\begin{subfigure}[t]{\columnwidth}
		\centering
		\input{plots/returns_qq-su_violinplot.pgf}
		\vspace*{-1.2\baselineskip}
		\caption{\sub}
		\label{fig_real_eval_QQ}
	\end{subfigure}%
	\hfill
	\begin{subfigure}[t]{\columnwidth}
		\centering
		\input{plots/returns_wam-bic_violinplot.pgf}
		\vspace*{-0.2\baselineskip}
		\caption{\bic}	
		\label{fig_real_eval_WAM}
	\end{subfigure}%
	\caption{%
		Performance of the different algorithms across both \simtoreal tasks.
		For each algorithm 20 policies have been trained, varying the random seed, and evaluated 5 times to estimate the mean return per policy (700 rollouts in total).
		The median performance per algorithm is displayed by white circles, and the inner quartiles are represented by thick vertical bars. 
		A dashed line in \subref{fig_real_eval_QQ} marks an approximate threshold where the task is solved, \ie, the rotary pole is stabilized on top in the center.
		SimOpt was not applicable to our open-loop \bic task \subref{fig_real_eval_WAM} because of its requirement for recorded observations.
	}
	\label{fig_real_eval_QQ_WAM}
\end{figure*}
As stated in Section~\ref{sec_adr_BayRn}, \ac{BayRn} was designed without an (explicit) system identification objective. However, we can see from Figure~\ref{fig_sim_eval_QQ_mean} that the maximizer of the \ac{GP}'s mean function ${\domdistrparams\opt = [0.0266, 0.1084]\tran}$ closely match the ground truth parameters $\domdistrparams_{\text{GT}} = [0.0264, 0.1045]\tran$.
Moreover, Figure~\ref{fig_sim_eval_QQ_std} displays how the uncertainty about the target domain return is reduced in the vicinity of the sampled parameter configurations.  
There are two decisive factors for the domain distribution parameter sampling process: the acquisition function (Algorithm~\ref{algo_BayRn} Line~\ref{line_BayRn_optimize_acqf}), and the quality of the found policy (Algorithm~\ref{algo_BayRn} Line~\ref{line_BayRn_optimize_policy}).
Concerning the latter, a failed training run is indistinguishable to a successful one which fails to transfer to the target domain, since the \acs{GP} only observes the estimated real-world return $\estedr[][\text{real}]{\polparamsopt}$.

\subsection{\Simtoreal Results}
\label{sec_adr_sim2real_results}
Figure~\ref{fig_real_eval_QQ_WAM} visualizes the results of the \simtoreal experiment described in Section~\ref{sec_adr_experimental_setup}.
The discrepancy between the performance of \ac{PPO} and \ac{PoWER} and the other algorithms reveals that domain randomization was the decisive part for \simtoreal transferability.
To verify that the \ac{PPO} and \ac{PoWER} learned meaningful policies, we checked them in the nominal simulation environments (not reported) and observed that they solve the tasks excellently.
In Figure~\ref{fig_real_eval_QQ}, we see that each median performance of \ac{BayRn}, SimOpt, and \ac{UDR} are above the success threshold. However, \ac{UDR} has a significantly higher variance.
SimOpt solves the \sub task in most cases. However, we noticed that the system identification subroutine can converge to extreme domain distribution parameters, rendering the next policy useless, which then yields a collection of poor trajectories for the next system identification, resulting in a downward spiral.
\ac{BayRn} on the other side relies on the policy optimizer's ability to robustly solve the simulated environment (Section~\ref{sec_adr_sim2sim_results}). This problem can be mediated by re-running the policy optimization in case a certain return threshold in simulation has not been exceeded.
For the \bic task, Figure~\ref{fig_real_eval_WAM} shows an improvement of \simtoreal transfer for \ac{BayRn}, especially since the tasks open-loop design amplifies domain mismatch.
During the experiments, we noticed that \ac{UDR} sometimes failed unexpectedly. We suspect the a high dependency on the initial state to be the reason for that.

Comparing the Furuta pendulum's nominal domain parameters $\domdistrparams_{\text{nom}} = [m_p, m_r, l_p, l_r]\tran = [0.024, 0.095, 0.129, 0.085]\tran$ to the means among \ac{BayRn}'s final estimate $\domdistrparams\opt_{\text{mean}} = [0.023, 0.098, 0.123, 0.087]\tran$, we see that the domain parameters' means changed by less than \SI{10}{\percent} each. Complementary the variances among \ac{BayRn}'s final estimate are $\domdistrparams\opt_{\text{var}} = [\num{6.29e-8}, \num{5.67e-6}, \num{4.10e-5}, \num{1.19e-5}]\tran$, indicating a higher uncertainty on the link lengths (relative to the means). 
Thus, the final domain parameters are well within the boundaries of the \ac{BO} search space (Table~\ref{tab_dom_distr_param_QQ_WAM}).
In combination, these small differences result in significantly different system dynamics. We believe this to be the reason why the baselines without domain randomization completely failed to transfer.

\section{Related Work}
\label{sec_adr_related_work}
We divide the related research on robot reinforcement learning from randomized simulations into approaches which use static (Section~\ref{sec_adr_related_work_static}) or adaptive (Section~\ref{sec_adr_related_work_adaptive})  distributions for sampling the physics parameters. \acf{BayRn} as introduced in Section~\ref{sec_adr_BayRn} belongs to the second category.

\subsection{Domain Randomization with Static Distributions}
\label{sec_adr_related_work_static}
Learning from a randomized simulator with fixed domain parameter distributions has bridged the reality gap in several cases~\cite{OpenAI_etal_2018,Lowrey_etal_2018,Muratore_etal_2019}.
Most prominently, the robotic in-hand manipulation reported in~\cite{OpenAI_etal_2018} showed that domain randomization in combination with careful model engineering and the usage of recurrent neural networks enables direct \simtoreal transfer on an unprecedented difficulty level.
Similarly, \citet{Lowrey_etal_2018} employed \acl{NPG} 
to learn a continuous controller for a positioning task, after carefully identifying the system's parameters. Their results show that the policy learned from the identified model was able to perform the \simtoreal transfer, but the policies learned from an ensemble of models were more robust to modeling errors.
\citet{Mordatch_etal_2015} used finite model ensembles to run trajectory optimization on a small-scale humanoid robot.
In contrast, \citet{Peng_etal_2018} combined model-free \ac{RL} with recurrent neural network policies trained on experience replay in order to push an object by controlling a robotic arm.
The usage of risk-averse objective function has been explored on MuJoCo tasks in~\cite{Rajeswaran_etal_2016}. The authors also provide a Bayesian point of view.

\citet{Cully_etal_2015} can be seen as an edge case of static and adaptive domain randomization, where a large set of policies is learned before execution on the physical robot and evaluated in simulation. Every policy is associated to one configuration of the so-called behavioral descriptors, which are related but not identical to domain parameters.
In contrast to \ac{BayRn}, there is no policy training after the initial phase. Instead of retraining or fine-tuning, the algorithm suggested in~\cite{Cully_etal_2015} reacts to performance drops, \eg due to damage, by using \ac{BO} to sequentially select a pretrained policy and measure its performance on the robot. The underlying \ac{GP} models the mapping from behavior space to performance.
This method demonstrated impressive damage recover abilities on a robotic locomotion and a reaching task. However, applying it to \ac{RL} poses big challenges. Most notably, the number of policies to be learned in order to populate the map, scales exponentially with the dimension of the behavioral descriptors, potentially leading to a very large number of training runs. 

Aside from to the previous methods, \citet{Muratore_etal_2019} propose an approach to estimate the transferability of a policy learned from randomized physics simulations. Moreover, the authors propose a meta-algorithm which provides a probabilistic guarantee on the performance loss when transferring the policy between two domains form the same distribution.

Static domain randomization has also been successfully applied to computer vision problems. A few examples that are:
(i) object detection~\cite{Tobin_etal_2017a},
(ii) synthetic object generation for grasp planning~\cite{Tobin_etal_2018}, and
(iii) autonomous drone flight~\cite{Sadeghi_Levine_2017}.

\subsection{Domain Randomization with Adaptive Distributions}
\label{sec_adr_related_work_adaptive}
\citet{Ruiz_etal_2018} proposed the meta-algorithm which is based on a bilevel optimization problem highly similar to the one of \ac{BayRn} (\ref{eq_BayRn_upper_problem}, \ref{eq_BayRn_lower_problem}).
However, there are two major differences. First, \ac{BayRn} uses Bayesian optimization on the acquired real-wold data to adapt the domain parameter distribution, whereas \enquote{learning to simulate} updates the domain parameter distribution using REINFORCE. 
Second, the approach in~\cite{Ruiz_etal_2018} has been evaluated in simulation on synthetic data, except for a semantic segmentation task. Thus, there was no dynamics-dependent interaction of the learned policy with the real world. 

With SPRL, \citet{Klink_etal_2019} derived a relative entropy \ac{RL} algorithm that endows the agent to adapt the domain parameter distribution, typically from easy to hard instances.
Hence, the overall training procedure can be interpreted as a curriculum learning problem. The authors were able to solve \simtosim goal reaching problems as well as a robotic \simtoreal \bic task, similar to the one in this paper.
One decisive difference to \ac{BayRn} is that the target domain parameter distribution has to be known beforehand.

The approach called \ac{ADR}~\cite{Mehta_etal_2019} also formulates the adaption of the domain parameter distribution as an \ac{RL} problem where different simulation instances are sampled and compared against a reference environment based on the resulting trajectories. This comparison is done by a discriminator which yields rewards proportional to the difficulty of distinguishing the simulated and real environments, hence providing an incentive to generate distinct domains. Using this reward signal, the domain parameters of the simulation instances are updated via \acl{SVPG}. 
\citet{Mehta_etal_2019} evaluated their method in a \simtoreal experiment where a robotic arm had to reach a desired point.
The strongest contrast between \ac{BayRn} and \ac{ADR} is they way in which new simulation environments are explored. While \ac{BayRn} can rely on well-studied \ac{BO} with an adjustable exploration-exploitation behavior, \ac{ADR} can be fragile since it couples discriminator training and policy optimization, which results in a non-stationary process where distribution of the domains depends on the discriminator's performance.

\citet{Paul_etal_2018} introduce \acl{FPO} which, like \ac{BayRn}, employs \ac{BO} to adapt the distribution of domain parameters such that using these for the subsequent training maximizes the policy's return.
At first glance the approaches look similar, but there is a major difference in how the upper level problem \eqref{eq_BayRn_upper_problem} is solved.
\acl{FPO} models the relation between the current domain parameters, the current policy and the return of the updated policy with a \ac{GP}. This design decision requires to feed the policy parameters into the \ac{GP} which is prohibitively expensive if done straightforwardly. Therefore, abstractions of the policy, so-called fingerprints, are created. These handcrafted features, \eg, the Gaussian approximation of the stationary state distribution, replace the policy to reduce the input dimension. 
The authors tested \acl{FPO} on three \simtosim tasks.
Contrarily, \ac{BayRn} has been designed without the need to approximate the policy. Moreover, we validated the presented method in \simtoreal settings.
\citet{Yu_etal_2019b} intertwine policy optimization, system identification, and domain randomization. The proposed method first identifies bounds on the domain parameters which are later used for learning from the randomized simulator. The suggested policy is conditioned on a latent space projection of the domain parameters. After training in simulation, a second system identification step is executed to find the projected domain parameters which maximize the return on the physical robot. This step runs \ac{BO} for a fixed number of iterations and is similar to solving the upper level problem in \eqref{eq_BayRn_upper_problem}. The algorithm was evaluated on the bipedal walking robot Darwin OP2.

In \citet{Ramos_etal_2019}, likelihood-free inference in combination with mixture density random Fourier networks is employed to perform a fully Bayesian treatment of the simulator's parameters.
Analyzing the obtained posterior over domain parameters, \citeauthor{Ramos_etal_2019} showed that BayesSim is, in a \simtosim setting, able to simultaneously infer different parameter configurations which can explain the observed trajectories.
The key difference between \ac{BayRn} and BayesSim is the objective for updating the domain parameters. While BayesSim maximizes the model's posterior likelihood, \ac{BayRn} updates the domain parameters such that the policy's return on the physical system is maximized.
The biggest advantage of \ac{BayRn} over BayesSim is its ability to work with very sparse real-world data, i.e. only the scalar return values.

\section{Conclusion} 
\label{sec_adr_conclusion}
We have introduced \acf{BayRn}, a policy search algorithm tailored to crossing the reality gap.
At its core, \ac{BayRn} learns from a randomized simulator while using Bayesian optimization for adapting the source domain distribution during learning.
In contrast to previous work, the presented algorithm constructs a probabilistic model of the connection between domain distribution parameters and the policy's return after training with these parameters in simulation.
Hence, \ac{BayRn} only requires little interaction with the real-world system.
We experimentally validated that the presented approach is able to solve two nonlinear robotic \simtoreal tasks.
Comparing the results against baseline methods showed that adapting the domain parameter distribution lead to policies with higher median performance and less variance.
In order to improve the scalability of the Bayesian optimization subroutine to higher numbers of domain distribution parameters, one could for example incorporate quantile Gaussian processes~\cite{Moriconi_etal_2020}, which have shown to scales up to problems with 60-dimensional input.

\vspace*{-0.4\baselineskip}
\section*{Acknowledgments}
Fabio Muratore gratefully acknowledges the financial support from \acl{HRIE}.

Jan Peters received funding from the European Union’s Horizon 2020 research and innovation programme under grant agreement No 640554.

\vspace*{-0.5\baselineskip}
\bibliographystyle{IEEEtranN}

\bibliography{IEEEabrv,main}

\appendices

\section{Modeling Details on the Platforms}
\label{secapdx_adr_setup}
The Furuta pendulum (Figure~\ref{fig_QQ_WAM}) is modeled as an underactuated nonlinear second-order dynamical system given by the solution of
\begin{align}
\label{eq_EoM_QQ}
&\begin{bmatrix}
J_r + m_p l_r^2 + \onefourth m_p l_p^2 (\cos{\alpha})^2 & \onehalf m_p l_p l_r \cos{\alpha} \\
\onehalf m_p l_p l_r \cos{\alpha} & J_p + \onefourth m_p l_p^2
\end{bmatrix}
\begin{bmatrix}
\ddot{\theta}\\ 
\ddot{\alpha}
\end{bmatrix} = \nonumber\\
&\begin{bmatrix}
\tau - \onehalf m_p l_p^2 \sin{\alpha} \cos{\alpha} \dot{\theta} \dot{\alpha} - \onehalf m_p l_p l_r \sin{\alpha} \dot{\alpha}^2 - d_r \dot{\theta}\\
-\onefourth m_p l_p^2 \sin{\alpha} \cos{\alpha} \dot{\theta}^2 - \onehalf m_p l_p g \sin{\alpha} - d_p \dot{\alpha}\\
\end{bmatrix}, 
\end{align}
with the rotary angle $\theta$ and the pendulum angle $\alpha$, which are defined to be zero when the rotary pole is centered and the pendulum pole is hanging down vertically.
While the system's state is defined as ${\fs = [\theta, \alpha, \dot{\theta}, \dot{\alpha}]\tran}$, the agent receives observations ${\fo = [\sin{\theta}, \cos{\theta}, \sin{\alpha}, \cos{\alpha}, \dot{\theta}, \dot{\alpha}]\tran}$.
The horizontal pole is actuated by commanding a motor voltage (action) $a$ which regulates the servo motor's torque ${\tau = k_m ( a - k_m \dot{\theta} ) / R_m}$.
One part of the domain parameters is sampled from distributions specified by in Table~\ref{tab_dom_distr_param_QQ}, while the remaining domain parameters are fixed at their nominal values given in~\cite{Muratore_SimuRLacra}.
We formulate the reward function based on an exponentiated quadratic cost
\begin{equation}
\begin{gathered}
\label{eq_rewfcn_QQ}
\rewfcn{\fs_t, a_t} = \exp{- \left( \fe_t\tran \fQ \fe_t + a_t R a_t \right) } \quad \text{with} \\
\fe_{t} = \left( \begin{bmatrix} 0 & \pi & 0 & 0 \end{bmatrix} - \fs_t\right) \hspace{-.9em} \mod 2\pi.
\end{gathered}
\end{equation}
Thus, the reward is in range $]0, 1]$ for every time step.

The 4-\acs{DoF} Barrett WAM (Figure~\ref{fig_QQ_WAM}) is simulated using MuJoCo
, wrapped by mujoco-py~\cite{mujoco_py}. The ball is attached to a string, which is mounted to the center of the cup's bottom plate. We model the string as a concatenation of 30 rigid bodies with two rotational joints per link (no torsion).
This specific \bic instance can be considered difficult, since the cups's diameter is only about twice as large as the ball's, and the string is rather short with a length of \SI{30}{\centi\meter}.
Similar to the Furuta pendulum, one part of the domain parameters is sampled from distributions specified by in Table~\ref{tab_dom_distr_param_WAM}, while the remaining domain parameters are fixed at their nominal values given in~\cite{Muratore_SimuRLacra}.
Since the feed-forward policy is executed without recording any observations, we define a discrete ternary reward function 
\begin{equation}
\rewfcn{\fs_T, \fa_T} = 
\begin{cases}
1   & \text{if the ball is in the cup,}\\
0.5 & \text{if the ball hit the cup's upper rim,}\\
0   & \text{else}
\end{cases}
\end{equation}
where the final reward given by the operator after the rollout ($\rewfcn{\fs_t, \fa_t} = 0$ for $t<T$) when running on the real system.
We found the separation in three cases to be helpful during learning and easily distinguishable from the others.
While training in simulation, successful trials are identified by detecting a collision between the ball and a virtual cylinder inside the cup.
Moreover, we have access to the full state, hence augment the reward function with a cost term that punishes deviations from the initial end-effector position. 

\section{Parameter Values for the Experiments}
\label{secapdx_adr_experiment_hparams}
Table~\ref{tab_adr_exp_hparams_QQ_WAM} lists the hyper-parameters for all training runs during the experiments in Section~\ref{sec_adr_experiments}.
The reported values have been tuned but not fully optimized.
\begin{table}[h]
	\centering
	\caption{%
		Hyper-parameter values for training the policies in Section~\ref{sec_adr_experiments}.
		The domain distribution parameters $\domdistrparams$ are listed in Table~\ref{tab_dom_distr_param_QQ_WAM}.
	}
	\label{tab_adr_exp_hparams_QQ_WAM}
	\begin{subtable}{\columnwidth}
		\centering
		\caption{\sub}
		\label{tab_adr_exp_hparams_QQ}
		\input{tables/experiment_hparam_QQ.tex}
	\end{subtable}%
	\\
	\vspace{\baselineskip}
	\begin{subtable}{\columnwidth}
		\centering
		\caption{\bic}
		\label{tab_adr_exp_hparams_WAM}
		\input{tables/experiment_hparam_WAM.tex}
	\end{subtable}%
\end{table}

\end{document}

%% file: fMRT_acronyms.tex
\acrodef{ADaPT}{Adaptive Policy Transfer for Stochastic Dynamics}
\acrodef{ADN}{Activation Dynamics Network}
\acrodef{ADR}{Active Domain Randomization}
\acrodef{ARC-t}{Asymmetric Regularized Cross-domain transformation}
\acrodef{A2RP}{Averaged Two-Replication Procedure}
\acrodef{ARPL}{Adversarially Robust Policy Learning}
\acrodef{BO}{Bayesian Optimization}
\acrodef{BayRn}{Bayesian Domain Randomization}
\acrodef{CAD}{Computer Aided Design}
\acrodef{CMA}{Covariance Matrix Adaptation}
\acrodef{CMA-ES}{Covariance Matrix Adaptation - Evolutionary Strategies}
\acrodef{CNN}{Convolutional Neural Network}
\acrodef{CoM}{Center of Mass}
\acrodef{CVaR}{Conditional Value at Risk}
\acrodef{DA}{Domain Adaptation}
\acrodef{DAN}{Deep Adaptation Network}
\acrodef{DDPG}{Deep Deterministic Policy Gradient}
\acrodef{DMP}{Dynamic Movement Primitive}
\acrodefplural{DMP}[DMPs]{Dynamic Movement Primitives}
\acrodef{DNN}{Deep Neural Network}
\acrodef{DoF}{Degree of Freedom}
\acrodefplural{DOF}[DOF]{Degrees of Freedom}
\acrodef{DR}{Domain Randomization}
\acrodef{DS}{Dynamical System}
\acrodef{EoM}{Equations of Motion}
\acrodef{EPOpt}{Ensemble Policy Optimization}
\acrodef{FNN}{Feedforward Neural Network}
\acrodef{FPO}{Fingerprint Policy Optimization}
\acrodef{GAN}{Generative Adversarial Network}
\acrodef{GMM}{Gaussian Mixture Model}
\acrodef{GP}{Gaussian Process}
\acrodefplural{GP}[GPs]{Gaussian Processes}
\acrodef{GPS}{Guided Policy Search}
\acrodef{GPU}{Graphics Processing Unit}
\acrodef{HER}{Hindsight Experience Replay}
\acrodef{HRI}{Honda Research Institute}
\acrodef{HRIE}{Honda Research Institute Europe}
\acrodef{I2RP}{Independent Two-Replication Procedure}
\acrodef{KL}{Kullback-Leibler}
\acrodef{LTI}{Linear Time-Invariant}
\acrodef{LQR}{Linear-Quadratic Regulator}
\acrodef{LSDA}{Large Scale Detection through Adaptation}
\acrodef{LSTM}{Long Short-Term Memory}
\acrodef{LWPR}{Locally Weighted Projection Regression}
\acrodef{MCMC}{Markov Chain Monte Calro}
\acrodef{MDP}{Markov Decision Process}
\acrodef{MP}{Movement Primitive}
\acrodefplural{MDP}{Markov Decision Processes}
\acrodef{MRP}{Multiple Replications Procedure}
\acrodef{NN}{Neural Network}
\acrodef{NPG}{Natural Policy Gradient}
\acrodef{ODE}{Ordinary Differential Equation}
\acrodef{ODEphys}[ODE]{Open Dynamics Engine}
\acrodef{OG}{Optimality Gap}
\acrodef{PNN}{Progressive Neural Network}
\acrodef{PD}{Proportional-Derivative}
\acrodef{POMDP}{Partially-Observable Markov Decision Process}
\acrodef{PoWER}{Policy learning by Weighting Exploration with the Returns}
\acrodef{PPO}{Proximal Policy Optimization}
\acrodef{ProMP}{Probabilistic Movement Primitive}
\acrodef{QBB}{Quanser Ball-Balancer}
\acrodef{QCP}{Quanser Cart-Pole}
\acrodef{QQ}{Quanser Qube}
\acrodef{RA}{Retrospective Approximation}
\acrodef{RARL}{Robust Adversarial Reinforcement Learning}
\acrodef{RBF}{Radial Basis Function}
\acrodef{REPS}{Relative Entropy Policy Search}
\acrodef{RGB}{Red Green Blue}
\acrodef{RNN}{Recurrent Neural Network}
\acrodef{RL}{Reinforcement Learning}
\acrodef{SAA}{Sample Average Approximation}
\acrodef{SOB}{Simulation Optimization Bias}
\acrodef{SP}{Stochastic Program}
\acrodefplural{SP}{Stochastic Programs}
\acrodef{SPOTA}{Simulation-based Policy Optimization with Transferability Assessment}
\acrodef{SVPG}{Stein Variational Policy Gradient}
\acrodef{TRPO}{Trust Region Policy Optimization}
\acrodef{UCB}{Upper Confidence Bound}
\acrodef{UDR}{Uniform Domain Randomization}
\acrodef{UCBOG}{Upper Confidence Bound on the Optimality Gap}
\acrodef{UCSOB}{Upper Confidence bound on the Simulation Optimization Bias}
\acrodef{UP-OSI}{Universal Policy - Online System Identification}

%% file: fMRT_macros.tex
\makeatletter
\@ifpackageloaded{xparse}{}{\usepackage{xparse}}
\@ifpackageloaded{tabularx}{}{\usepackage{tabularx}} 
\@ifpackageloaded{xifthen}{}{\usepackage{xifthen}}
\@ifpackageloaded{xspace}{}{\usepackage{xspace}} 
\@ifpackageloaded{amsopn}{}{\usepackage{amsopn}} 
\@ifpackageloaded{amssymb}{}{\usepackage{amssymb}} 
\@ifpackageloaded{mathtools}{}{\usepackage{mathtools}} 
\@ifpackageloaded{bm}{}{\usepackage{bm}} 
\@ifpackageloaded{siunitx}{}{%
\usepackage[detect-all,scientific-notation=false]{siunitx} 
\sisetup{%
	output-exponent-marker=\ensuremath{\mathrm{e}}, 
	per-mode=symbol, 
}%
}
\makeatother

\definecolor{fMRTblack}{HTML}{2B2E34}
\definecolor{fMRTwhite}{HTML}{F3EEE1}
\definecolor{fMRTyellow}{HTML}{FFD564}
\definecolor{fMRTorange}{HTML}{FF5500}
\colorlet{fMRTlightGray}{white!80!fMRTblack}
\colorlet{fMRTgray}{white!35!fMRTblack}
\colorlet{fMRTdarkGray}{white!20!fMRTblack}
\colorlet{fMRTlightOrange}{fMRTorange!60!fMRTwhite}
\definecolor{fMRTblue}{HTML}{3333B3}
\colorlet{fMRTlightBlue}{fMRTblue!30!white}
\definecolor{fMRTverylightRed}{HTML}{FF8E6B}
\colorlet{fMRTlightRed}{red!60!fMRTwhite}
\colorlet{fMRTdarkRed}{red!70!fMRTgray} 
\definecolor{fMRTlightGreen}{HTML}{8CDD81} 
\definecolor{fMRTborwn}{HTML}{8B4513} 
\colorlet{fMRTlightBorwn}{fMRTborwn!60!fMRTwhite}

\colorlet{fMRTdark@backgroundInner}{fMRTblack}
\colorlet{fMRTdark@backgroundOuter}{fMRTblack}
\colorlet{fMRTlight@backgroundInner}{fMRTwhite}
\colorlet{fMRTlight@backgroundOuter}{fMRTwhite}

\newcolumntype{L}{>{\raggedright\arraybackslash}X}
\newcolumntype{R}{>{\raggedleft\arraybackslash}X}
\newcolumntype{C}{>{\centering\arraybackslash}X}

\newcommand{\RR}{\mathbb{R}} 
\newcommand{\onehalf}{\frac{1}{2}} 
\newcommand{\onefourth}{\frac{1}{4}} 
\DeclareDocumentCommand{\set}{O{} O{} m}{\{#3\}_{#1}^{#2}} 
\DeclareDocumentCommand{\tuple}{O{} O{} m}{\left\langle#3\right\rangle_{#1}^{#2}} 

\newcommand{\data}{{\cal D}} 

\newcommand{\stateset}[1][]{\mathcal{S}_{#1}} 
\newcommand{\actionset}[1][]{\mathcal{A}_{#1}} 
\newcommand{\transprob}[1][]{\mathcal{P}_{#1}} 
\newcommand{\initstatedistr}[1][]{\mu_{0#1}} 
\newcommand{\mdp}[1][]{\mathcal{M}_{#1}} 
\newcommand{\dimstate}{{n_s}} 
\newcommand{\dimact}{{n_a}} 
\newcommand{\dimpolparam}{{n_{\polparam}}} 
\newcommand{\dimdomparam}{{n_{\domparam}}} 
\newcommand{\optsym}{\star} 
\newcommand{\opt}{^\optsym}

\newcommand{\eg}{e.g.\xspace}
\newcommand{\ie}{i.e.\xspace}

\newcommand{\wrt}{w.r.t.~}
\newcommand{\aka}{a.k.a.~}

\newcommand{\Simtosim}{{Sim-to-sim}\xspace}
\newcommand{\simtosim}{{sim-to-sim}\xspace}
\newcommand{\Simtoreal}{{Sim-to-real}\xspace}
\newcommand{\simtoreal}{{sim-to-real}\xspace}
\newcommand{\sota}{{state-of-the-art}\xspace}
\newcommand{\Sota}{{State-of-the-art}\xspace}

\newcommand{\sub}{swing-up and balance\xspace}
\newcommand{\bic}{ball-in-a-cup\xspace}

\DeclareMathOperator*{\argmax}{arg\,max}

\newcommand{\distr}[3]{#1\left( #2 \big| #3\right) } 
\newcommand{\distrnormal}[2]{\distr{\mathcal{N}}{#1}{#2}} 
\newcommand{\E}[1]{\mathbb{E}\! \left[ #1\right] } 
\newcommand{\Esub}[2]{\mathbb{E}_{#1} \! \left[ #2\right] } 
\newcommand{\EsubBig}[2]{\mathbb{E}_{#1} \! \Big[ #2\Big] } 
\newcommand{\V}[1]{\mathbb{V} \! \left[ #1\right] } 
\renewcommand{\exp}[1]{\mathrm{exp}\left( #1\right) } 

\renewcommand{\sin}[1]{\mathrm{sin}\!\left( #1\right) }
\renewcommand{\cos}[1]{\mathrm{cos}\!\left( #1\right) }
\makeatletter
\newcommand{\@givennostar}[2]{\left. #1\right| #2} 
\newcommand{\@givenstar}[3][]{#2 #1| #3} 
\newcommand{\given}{\@ifstar\@givenstar\@givennostar}
\makeatother

\newcommand{\ninit}{n_{\text{init}}}
\newcommand{\tholdsucc}{\edrsym^{\text{succ}}}
\newcommand{\tholdsuccQQ}{375}

\newcommand{\tran}{^\textsf{\upshape T}} 

\newcommand{\rewsym}{r}
\DeclareDocumentCommand{\rewfcn}{O{} O{} m}{\rewsym_{#1}^{#2}\!\left( #3\right) }

\newcommand{\trajssym}{\ftau}
\DeclareDocumentCommand{\traj}{O{} O{}}{\trajssym_{#1}^{#2}} 
\newcommand{\statedistrsym}{\mu}
\DeclareDocumentCommand{\statedistr}{O{} O{} m}{\statedistrsym_{#1}^{#2}\!\left( #3\right) } 

\newcommand{\domparam}{\xi}
\newcommand{\domparams}{\fxi}

\newcommand{\domparamdistrsym}{\nu} 
\newcommand{\domparamdistr}[1]{\domparamdistrsym\!\left( #1\right) }
\newcommand{\domdistrparams}{\fphi} 
\newcommand{\domdistrparamspace}{\Phi} 

\newcommand{\polsym}{\pi}
\DeclareDocumentCommand{\pol}{O{} O{} m}{\polsym_{#1}^{#2}\!\left( #3\right) } 
\newcommand{\polparam}{\theta}
\newcommand{\polparamspace}{\Theta}
\DeclareDocumentCommand{\polparams}{O{} O{}}{\ftheta_{#1}^{#2}} 
\DeclareDocumentCommand{\polparamsopt}{O{} O{}}{\ftheta_{#1}^{#2 \optsym}} 
\DeclareDocumentCommand{\polparamopt}{O{} O{}}{\polparam_{#1}^{#2 \optsym}} 

\newcommand{\GPsym}{\mathcal{GP}}
\newcommand{\GPmean}{m}
\newcommand{\GPcovar}{k}
\newcommand{\GP}[2]{\GPsym\!\left( #1,#2\right)}
\newcommand{\acqfcnsym}{a}
\DeclareDocumentCommand{\acqfcn}{O{} O{} m}{\acqfcnsym_{#1}^{#2}\!\left( #3\right) }
\newcommand{\boinputspace}[1][]{\mathcal{X}_{#1}} 

\newcommand{\optgapsym}{G}
\DeclareDocumentCommand{\optgap}{O{} O{} m}{\optgapsym_{#1}^{#2}\!\left( #3\right) } 
\newcommand{\estoptgapsym}{\hat{G}}
\DeclareDocumentCommand{\estoptgap}{O{} O{} m}{\estoptgapsym_{#1}^{#2}\!\left( #3\right) } 
\newcommand{\meanoptgapsym}{\bar{G}}
\DeclareDocumentCommand{\optgapmean}{O{} O{} m}{\meanoptgapsym_{#1}^{#2}\!\left( #3\right) } 
\newcommand{\optgapsampsetsym}{\mathcal{G}}
\DeclareDocumentCommand{\optgapsampset}{O{} O{}}{\optgapsampsetsym_{#1}^{#2}} 

\iftrue
	\newcommand{\bootsym}{B}
	\DeclareDocumentCommand{\bootestoptgap}{O{} O{} m}{\estoptgapsym_{#1}^{\bootsym #2}\!\left( #3\right) }
	\DeclareDocumentCommand{\bootmeanoptgap}{O{} O{} m}{\meanoptgapsym_{#1}^{\bootsym #2}\!\left( #3\right) }
	\DeclareDocumentCommand{\bootoptgapsampset}{O{} O{}}{\optgapsampsetsym_{#1}^{\bootsym #2}}
\fi

\newcommand{\edrsym}{J}
\DeclareDocumentCommand{\edr}{O{} O{} m}{\edrsym_{#1}^{#2}\!\left( #3\right)} 

\newcommand{\estedrsym}{\hat{J}}
\DeclareDocumentCommand{\estedr}{O{}  O{} m}{\estedrsym_{#1}^{#2}\!\left( #3\right)} 
\newcommand{\eedrsym}{J} 
\DeclareDocumentCommand{\eedr}{O{} O{} m}{\eedrsym_{#1}^{#2}\!\left( #3\right)} 
\DeclareDocumentCommand{\esteedr}{O{} O{} m}{\estedrsym_{#1}^{#2}\!\left( #3\right)} 
\newcommand{\objfunsym}{f}
\DeclareDocumentCommand{\objfun}{O{} m}{\objfunsym_{#1}\!\left( #2\right)} 

\newcommand{\dsdynsym}{f}
\DeclareDocumentCommand{\dsdyn}{O{} m}{\dsdynsym_{#1}\!\left( #2\right)} 

\newcommand{\potdynsym}{f}
\newcommand{\potsdynsym}{\ff}
\DeclareDocumentCommand{\potdyn}{O{} m}{\potdynsym_{#1}\!\left( #2\right)} 
\DeclareDocumentCommand{\potsdyn}{O{} m}{\potsdynsym_{#1}\!\left( #2\right)} 

\ExplSyntaxOn
\NewDocumentCommand{\eqsref}{m}{\quinn_eqsref:n {#1}}
\seq_new:N \l_quinn_eqsref_seq
\cs_new:Npn \quinn_eqsref:n #1
{
	\seq_set_split:Nnn \l_quinn_eqsref_seq { , } { #1 }
	\seq_pop_right:NN \l_quinn_eqsref_seq \l_tmpa_tl
	( 
	\seq_map_inline:Nn \l_quinn_eqsref_seq
	{ \ref{##1},\nobreakspace } 
	\exp_args:NV \ref \l_tmpa_tl 
	) 
}
\ExplSyntaxOff

\SetKw{kwNot}{not}
\SetKw{kwAnd}{and}
\SetKw{kwBreak}{break}
\SetKwInOut{Input}{input}
\SetKwInOut{Output}{output}
\SetKwInOut{Initialization}{init}
\SetKwComment{Comment}{$\triangleright$\ }{}
\SetKwRepeat{Do}{do}{while} 
\SetKwBlock{With}{with}{end} 
\newcommand{\llIf}[2]{{\let\par\relax\lIf{#1}{#2}}}
\newcommand{\llElse}[1]{{\let\par\relax\lElse{#1}}}
\newcommand{\llFor}[2]{{\let\par\relax\lFor{#1}{#2}}}

\DeclareBoldMathCommand{\fnull}{0}
\DeclareBoldMathCommand{\fO}{0}
\DeclareBoldMathCommand{\fA}{A}
\DeclareBoldMathCommand{\fa}{a}
\DeclareBoldMathCommand{\fB}{B}
\DeclareBoldMathCommand{\fb}{b}
\DeclareBoldMathCommand{\fC}{C}
\DeclareBoldMathCommand{\fc}{c}
\DeclareBoldMathCommand{\fD}{D}
\DeclareBoldMathCommand{\fd}{d}
\DeclareBoldMathCommand{\fE}{E}
\DeclareBoldMathCommand{\fe}{e}
\DeclareBoldMathCommand{\fF}{F}
\DeclareBoldMathCommand{\ff}{f}
\DeclareBoldMathCommand{\fG}{G}
\DeclareBoldMathCommand{\fg}{g}
\DeclareBoldMathCommand{\fH}{H}
\DeclareBoldMathCommand{\fh}{h}
\DeclareBoldMathCommand{\fI}{I}
\DeclareBoldMathCommand{\fJ}{J}
\DeclareBoldMathCommand{\fK}{K}
\DeclareBoldMathCommand{\fM}{M}
\DeclareBoldMathCommand{\fm}{m}
\DeclareBoldMathCommand{\fN}{N}
\DeclareBoldMathCommand{\fn}{n}
\DeclareBoldMathCommand{\fo}{o}
\DeclareBoldMathCommand{\fP}{P}
\DeclareBoldMathCommand{\fp}{p}
\DeclareBoldMathCommand{\fQ}{Q}
\DeclareBoldMathCommand{\fq}{q}
\DeclareBoldMathCommand{\fq}{q}

\DeclareBoldMathCommand{\fR}{R}
\DeclareBoldMathCommand{\fr}{r}

\DeclareBoldMathCommand{\fS}{S}
\DeclareBoldMathCommand{\fs}{s}

\DeclareBoldMathCommand{\ft}{t}
\DeclareBoldMathCommand{\fT}{T}
\DeclareBoldMathCommand{\fU}{U}
\DeclareBoldMathCommand{\fu}{u}
\DeclareBoldMathCommand{\fV}{V}
\DeclareBoldMathCommand{\fv}{v}
\DeclareBoldMathCommand{\fW}{W}
\DeclareBoldMathCommand{\fw}{w}
\DeclareBoldMathCommand{\fx}{x}

\DeclareBoldMathCommand{\fY}{Y}
\DeclareBoldMathCommand{\fy}{y}
\DeclareBoldMathCommand{\fZ}{Z}
\DeclareBoldMathCommand{\falpha}{\alpha}

\DeclareBoldMathCommand{\fbeta}{\beta}
\DeclareBoldMathCommand{\fchi}{\chi}
\DeclareBoldMathCommand{\fepsilon}{\epsilon}
\DeclareBoldMathCommand{\fvarepsilon}{\varepsilon}
\DeclareBoldMathCommand{\fgamma}{\gamma}
\DeclareBoldMathCommand{\fGamma}{\Gamma}
\DeclareBoldMathCommand{\flambda}{\lambda}
\DeclareBoldMathCommand{\fLambda}{\Lambda}
\DeclareBoldMathCommand{\fmu}{\mu}
\DeclareBoldMathCommand{\fnu}{\nu}
\DeclareBoldMathCommand{\fomega}{\omega}
\DeclareBoldMathCommand{\fpi}{\pi}
\DeclareBoldMathCommand{\fphi}{\phi}
\DeclareBoldMathCommand{\fPhi}{\Phi}
\DeclareBoldMathCommand{\fpsi}{\psi}
\DeclareBoldMathCommand{\fsigma}{\sigma}
\DeclareBoldMathCommand{\fSigma}{\Sigma}
\DeclareBoldMathCommand{\ftau}{\tau}
\DeclareBoldMathCommand{\ftheta}{\theta}
\DeclareBoldMathCommand{\fTheta}{\Theta}

\DeclareBoldMathCommand{\fxi}{\xi}

%% file: pseudocode/BayRn.tex
\DontPrintSemicolon
\Input{domain parameter distribution $\domparamdistr{\domparams;\domdistrparams}$, parameter space $\domdistrparamspace = [\domdistrparams_\text{min}, \domdistrparams_\text{max}]$, algorithm \texttt{PolOpt}, \acl{GP} $\GPsym$, acquisition function $\acqfcnsym$, \mbox{hyper-parameters} $\ninit$, $n_\tau$, $\edrsym^\text{succ}$ }
\Output{%
	maximum a posteriori domain distribution parameter $\domdistrparams\opt$ and policy $\pol{\polparamsopt}$ 
}

\Comment{Initialization phase}
Initialize empty data set and $\ninit$ policies randomly\label{line_BayRn_init}\;
$\data \gets \set{}$ ;
$ \pol{\polparams[1:n_{\text{init}}]} \gets \polparams[1:n_{\text{init}}] \sim \polparamspace$\;

Sample $\ninit$ source domain distribution parameter sets and train in randomized simulators\;

$\domdistrparams_{1:\ninit} \gets \domdistrparams_{1:\ninit} \sim \domdistrparamspace$\;

$\polparams[1:n_{\text{init}}][\optsym] \gets \texttt{PolOpt}\!\left[ \pol{\polparams[1:n_{\text{init}}]}, \domparamdistr{\domparams;\domdistrparams_{1:\ninit}} \right]$\;

Evaluate the $\ninit$ policies on the target domain for $n_\tau$ rollouts and estimate the return

$\estedr[][\text{real}]{\polparams[1:n_{\text{init}}][\optsym]} \gets 1/n_\tau \sum_{j=1}^{n_\tau}$ $\edr[j][\text{real}]{\polparams[1:\ninit][\optsym]}$\; 

Augment the data set and update the \ac{GP}'s posterior distribution\;

$\data \cup \set[i=1][\ninit]{\domdistrparams_i, \estedr[][\text{real}]{\polparams[i][\optsym]}}$ ;
$\GP{\GPmean}{\GPcovar} \gets \GPsym \big(\GPmean, \GPcovar \big| \data \big)$\;

\Do(\Comment*[f]{\Simtoreal loop}){$\estedr[][\text{real}]{\polparamsopt} < \edrsym^\text{succ}$ and $n_{\text{iter}} \le n_{\text{iter,max}}$} 
{
    Optimize the \acs{GP}'s acquisition function\;
    
	$\domdistrparams\opt \gets \argmax_{\domdistrparams \in \domdistrparamspace} \acqfcn{\domdistrparams, \data}$ \label{line_BayRn_optimize_acqf}\;
    
    Train a policy using the obtained domain distribution parameter set
	
	$\polparamsopt \gets \texttt{PolOpt}\!\left[ \pol{\polparams}, \domparamdistr{\domparams;\domdistrparams\opt} \right]$ \label{line_BayRn_optimize_policy}\;
    
    Evaluate the policy on the target domain for $n_\tau$ rollouts and estimate the return
	
	$\estedr[][\text{real}]{\polparamsopt} \gets 1/n_\tau \sum_{j=1}^{n_\tau}$ $\edr[j][\text{real}]{\polparams[][\optsym]}$\; 
	
    Augment the data set and update the \ac{GP}'s posterior distribution\;
    
    $\data \cup \set{\domdistrparams\opt, \estedr[][\text{real}]{\polparams[][\optsym]}}$ ;
    $\GP{\GPmean}{\GPcovar} \gets \GPsym \big(\GPmean, \GPcovar \big| \data \big)$\;
}\label{line_BayRn_stopping_criterion}
Train the maximum a posteriori policy (repeat the Lines~\ref{line_BayRn_optimize_acqf} and~\ref{line_BayRn_optimize_policy} once)

%% file: images/QQ_WAM_domain_param.pdf_tex
\begingroup%
  \makeatletter%
  \providecommand\color[2][]{%
    \errmessage{(Inkscape) Color is used for the text in Inkscape, but the package 'color.sty' is not loaded}%
    \renewcommand\color[2][]{}%
  }%
  \providecommand\transparent[1]{%
    \errmessage{(Inkscape) Transparency is used (non-zero) for the text in Inkscape, but the package 'transparent.sty' is not loaded}%
    \renewcommand\transparent[1]{}%
  }%
  \providecommand\rotatebox[2]{#2}%
  \newcommand*\fsize{\dimexpr\f@size pt\relax}%
  \newcommand*\lineheight[1]{\fontsize{\fsize}{#1\fsize}\selectfont}%
  \ifx\svgwidth\undefined%
    \setlength{\unitlength}{164.694066bp}%
    \ifx\svgscale\undefined%
      \relax%
    \else%
      \setlength{\unitlength}{\unitlength * \real{\svgscale}}%
    \fi%
  \else%
    \setlength{\unitlength}{\svgwidth}%
  \fi%
  \global\let\svgwidth\undefined%
  \global\let\svgscale\undefined%
  \makeatother%
  \begin{picture}(1,0.51917237)%
    \lineheight{1}%
    \setlength\tabcolsep{0pt}%
    \put(0,0){\includegraphics[width=\unitlength,page=1]{QQ_WAM_domain_param.pdf}}%
    \put(0.90691311,0.02398031){\color[rgb]{0.29803922,0,0}\makebox(0,0)[t]{\lineheight{1.25}\smash{\begin{tabular}[t]{c}\textcolor[RGB]{76,0,0}{$m_b$}\end{tabular}}}}%
    \put(0.72036777,0.0239804){\color[rgb]{0.29803922,0,0}\makebox(0,0)[t]{\lineheight{1.25}\smash{\begin{tabular}[t]{c}\textcolor[RGB]{76,0,0}{$d_s, l_s$}\end{tabular}}}}%
    \put(0.67190433,0.21126348){\color[rgb]{0.29803922,0,0}\makebox(0,0)[t]{\lineheight{1.25}\smash{\begin{tabular}[t]{c}\textcolor[RGB]{76,0,0}{$d_j, \mu_s$}\end{tabular}}}}%
    \put(0,0){\includegraphics[width=\unitlength,page=2]{QQ_WAM_domain_param.pdf}}%
    \put(0.08045551,0.1168112){\color[rgb]{1,1,1}\makebox(0,0)[t]{\lineheight{1.25}\smash{\begin{tabular}[t]{c}$l_r,$\end{tabular}}}}%
    \put(0.20821735,0.15793928){\color[rgb]{1,1,1}\makebox(0,0)[t]{\lineheight{1.25}\smash{\begin{tabular}[t]{c}$l_p,$\end{tabular}}}}%
    \put(0,0){\includegraphics[width=\unitlength,page=3]{QQ_WAM_domain_param.pdf}}%
    \put(0.20509747,0.07687929){\color[rgb]{1,1,1}\makebox(0,0)[t]{\lineheight{1.25}\smash{\begin{tabular}[t]{c}$m_p$\end{tabular}}}}%
    \put(0.07993931,0.04394991){\color[rgb]{1,1,1}\makebox(0,0)[t]{\lineheight{1.25}\smash{\begin{tabular}[t]{c}$m_r$\end{tabular}}}}%
  \end{picture}%
\endgroup%

%% file: tables/domain_distribution_parameters_QQ.tex
\begin{tabular}{ll@{ \ }ll} 
	\rowcolor{gray!50}
	\textbf{Parameter}        & \multicolumn{2}{l}{\textbf{Range}}              & \textbf{Unit}\\
	pendulum pole mass mean   & $\E{m_p}$ & $\in [\num{0.0192},\num{0.0288}]$    & $\si{kg}$\\
	pendulum pole mass var.   & $\V{m_p}$ & $\in [\num{5.76e-10},\num{5.76e-6}]$ & $\si{kg\squared}$\\
	\rowcolor{gray!25}
	rotary pole mass mean     & $\E{m_r}$ & $\in [\num{0.076},\num{0.114}]$     & $\si{kg}$\\
	\rowcolor{gray!25}
	rotary pole mass var.     & $\V{m_r}$ & $\in [\num{9.03e-9},\num{9.03e-5}]$ & $\si{kg\squared}$\\
	pendulum pole length mean & $\E{l_p}$ & $\in [\num{0.1032},\num{0.1548}]$    & $\si{m}$\\
	pendulum pole length var. & $\V{l_p}$ & $\in [\num{1.66e-8},\num{1.66e-4}]$  & $\si{m\squared}$\\
	\rowcolor{gray!25}
	rotary pole length mean   & $\E{l_r}$ & $\in [\num{0.068},\num{0.102}]$      & $\si{m}$\\
	\rowcolor{gray!25}
	rotary pole length var.   & $\V{l_r}$ & $\in [\num{7.23e-9},\num{7.23e-5}]$  & $\si{m\squared}$\\
\end{tabular}

%% file: tables/domain_distribution_parameters_WAM.tex
\begin{tabular}{ll@{ }ll} 
	\rowcolor{gray!50}
	\textbf{Parameter}      & \multicolumn{2}{l}{\textbf{Range}}             & \textbf{Unit}\\
	string length mean      & $\E{l_s}$   & $\in [\num{0.285},\num{0.315}]$   & $\si{m}$\\
	string length variance  & $\V{l_s}$   & $\in [\num{9e-8},\num{2.25e-4}]$  & $\si{m\squared}$\\
	\rowcolor{gray!25}
	string damping mean     & $\E{d_s}$   & $\in [\num{0},\num{2e-4}]$        & $\si{\newton\per\second}$\\
	\rowcolor{gray!25}
	string damping variance & $\V{d_s}$   & $\in [\num{3.33e-13},\num{8.33e-10}]$     & $\si{\newton\squared\per\second\squared}$\\
	ball mass mean          & $\E{m_b}$   & $\in [\num{0.0179},\num{0.0242}]$ & $\si{\kilogram}$\\
	ball mass variance      & $\V{m_b}$   & $\in [\num{4.41e-10},\num{4.41e-6}]$ & $\si{\kilogram\squared}$\\
	\rowcolor{gray!25}
	joint damping mean      & $\E{d_j}$   & $\in [\num{0.0},\num{0.1}]$       & $\si{\newton\per\second}$\\
	\rowcolor{gray!25}
	joint damping variance  & $\V{d_j}$   & $\in [\num{3.33e-8},\num{2.08e-4}]$    & $\si{\newton\squared\per\second\squared}$\\
	joint stiction mean     & $\E{\mu_s}$ & $\in [\num{0},\num{0.4}]$         & $-$\\
	joint stiction variance & $\V{\mu_s}$ & $\in [\num{1.33e-6},\num{3.33e-3}]$      & $-$\\
\end{tabular}

%% file: plots/eval_sim2sim_1row2col.tex
\begin{subfigure}[t]{0.498\columnwidth}
	\hspace*{38pt}\input{plots/gp_posterior_ret_mean_cb.pgf}\\[0pt]%
	\vspace*{-15pt}
	\input{plots/gp_posterior_ret_mean_hm.pgf}%
	\vspace{-5pt}
	\caption{posterior mean}
	\label{fig_sim_eval_QQ_mean}
\end{subfigure}%
\hfill
\begin{subfigure}[t]{0.498\columnwidth}
	\hspace*{38pt}\input{plots/gp_posterior_ret_std_cb.pgf}\\[0pt]%
	\vspace*{-15pt}
	\input{plots/gp_posterior_ret_std_hm.pgf}%
	\vspace*{-5pt}
	\caption{posterior std.}	
	\label{fig_sim_eval_QQ_std}
\end{subfigure}

%% file: plots/gp_posterior_ret_mean_cb.pgf
\begingroup%
\makeatletter%
\begin{pgfpicture}%
\pgfpathrectangle{\pgfpointorigin}{\pgfqpoint{1.155000in}{0.420000in}}%
\pgfusepath{use as bounding box, clip}%
\begin{pgfscope}%
\pgfsetbuttcap%
\pgfsetmiterjoin%
\definecolor{currentfill}{rgb}{1.000000,1.000000,1.000000}%
\pgfsetfillcolor{currentfill}%
\pgfsetlinewidth{0.000000pt}%
\definecolor{currentstroke}{rgb}{1.000000,1.000000,1.000000}%
\pgfsetstrokecolor{currentstroke}%
\pgfsetdash{}{0pt}%
\pgfpathmoveto{\pgfqpoint{0.000000in}{0.000000in}}%
\pgfpathlineto{\pgfqpoint{1.155000in}{0.000000in}}%
\pgfpathlineto{\pgfqpoint{1.155000in}{0.420000in}}%
\pgfpathlineto{\pgfqpoint{0.000000in}{0.420000in}}%
\pgfpathclose%
\pgfusepath{fill}%
\end{pgfscope}%
\begin{pgfscope}%
\pgfpathrectangle{\pgfqpoint{0.041670in}{0.041670in}}{\pgfqpoint{1.071660in}{0.128438in}}%
\pgfusepath{clip}%
\pgfsetbuttcap%
\pgfsetmiterjoin%
\definecolor{currentfill}{rgb}{1.000000,1.000000,1.000000}%
\pgfsetfillcolor{currentfill}%
\pgfsetlinewidth{0.010037pt}%
\definecolor{currentstroke}{rgb}{1.000000,1.000000,1.000000}%
\pgfsetstrokecolor{currentstroke}%
\pgfsetdash{}{0pt}%
\pgfpathmoveto{\pgfqpoint{0.041670in}{0.041670in}}%
\pgfpathlineto{\pgfqpoint{0.052387in}{0.041670in}}%
\pgfpathlineto{\pgfqpoint{1.102613in}{0.041670in}}%
\pgfpathlineto{\pgfqpoint{1.113330in}{0.041670in}}%
\pgfpathlineto{\pgfqpoint{1.113330in}{0.170108in}}%
\pgfpathlineto{\pgfqpoint{1.102613in}{0.170108in}}%
\pgfpathlineto{\pgfqpoint{0.052387in}{0.170108in}}%
\pgfpathlineto{\pgfqpoint{0.041670in}{0.170108in}}%
\pgfpathclose%
\pgfusepath{stroke,fill}%
\end{pgfscope}%
\begin{pgfscope}%
\pgfsys@transformshift{0.042000in}{0.042000in}%
\pgftext[left,bottom]{\includegraphics[interpolate=true,width=1.072000in,height=0.128000in]{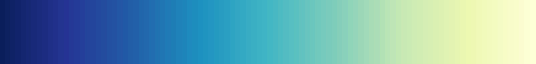}}%
\end{pgfscope}%
\begin{pgfscope}%
\pgfsetbuttcap%
\pgfsetroundjoin%
\definecolor{currentfill}{rgb}{0.000000,0.000000,0.000000}%
\pgfsetfillcolor{currentfill}%
\pgfsetlinewidth{0.803000pt}%
\definecolor{currentstroke}{rgb}{0.000000,0.000000,0.000000}%
\pgfsetstrokecolor{currentstroke}%
\pgfsetdash{}{0pt}%
\pgfsys@defobject{currentmarker}{\pgfqpoint{0.000000in}{0.000000in}}{\pgfqpoint{0.000000in}{0.048611in}}{%
\pgfpathmoveto{\pgfqpoint{0.000000in}{0.000000in}}%
\pgfpathlineto{\pgfqpoint{0.000000in}{0.048611in}}%
\pgfusepath{stroke,fill}%
}%
\begin{pgfscope}%
\pgfsys@transformshift{0.110598in}{0.170108in}%
\pgfsys@useobject{currentmarker}{}%
\end{pgfscope}%
\end{pgfscope}%
\begin{pgfscope}%
\definecolor{textcolor}{rgb}{0.000000,0.000000,0.000000}%
\pgfsetstrokecolor{textcolor}%
\pgfsetfillcolor{textcolor}%
\pgftext[x=0.110598in,y=0.267330in,,bottom]{\color{textcolor}\rmfamily\fontsize{9.000000}{10.800000}\selectfont \(\displaystyle {0}\)}%
\end{pgfscope}%
\begin{pgfscope}%
\pgfsetbuttcap%
\pgfsetroundjoin%
\definecolor{currentfill}{rgb}{0.000000,0.000000,0.000000}%
\pgfsetfillcolor{currentfill}%
\pgfsetlinewidth{0.803000pt}%
\definecolor{currentstroke}{rgb}{0.000000,0.000000,0.000000}%
\pgfsetstrokecolor{currentstroke}%
\pgfsetdash{}{0pt}%
\pgfsys@defobject{currentmarker}{\pgfqpoint{0.000000in}{0.000000in}}{\pgfqpoint{0.000000in}{0.048611in}}{%
\pgfpathmoveto{\pgfqpoint{0.000000in}{0.000000in}}%
\pgfpathlineto{\pgfqpoint{0.000000in}{0.048611in}}%
\pgfusepath{stroke,fill}%
}%
\begin{pgfscope}%
\pgfsys@transformshift{0.670062in}{0.170108in}%
\pgfsys@useobject{currentmarker}{}%
\end{pgfscope}%
\end{pgfscope}%
\begin{pgfscope}%
\definecolor{textcolor}{rgb}{0.000000,0.000000,0.000000}%
\pgfsetstrokecolor{textcolor}%
\pgfsetfillcolor{textcolor}%
\pgftext[x=0.670062in,y=0.267330in,,bottom]{\color{textcolor}\rmfamily\fontsize{9.000000}{10.800000}\selectfont \(\displaystyle {300}\)}%
\end{pgfscope}%
\begin{pgfscope}%
\pgfsetbuttcap%
\pgfsetmiterjoin%
\pgfsetlinewidth{0.803000pt}%
\definecolor{currentstroke}{rgb}{0.000000,0.000000,0.000000}%
\pgfsetstrokecolor{currentstroke}%
\pgfsetdash{}{0pt}%
\pgfpathmoveto{\pgfqpoint{0.041670in}{0.041670in}}%
\pgfpathlineto{\pgfqpoint{0.052387in}{0.041670in}}%
\pgfpathlineto{\pgfqpoint{1.102613in}{0.041670in}}%
\pgfpathlineto{\pgfqpoint{1.113330in}{0.041670in}}%
\pgfpathlineto{\pgfqpoint{1.113330in}{0.170108in}}%
\pgfpathlineto{\pgfqpoint{1.102613in}{0.170108in}}%
\pgfpathlineto{\pgfqpoint{0.052387in}{0.170108in}}%
\pgfpathlineto{\pgfqpoint{0.041670in}{0.170108in}}%
\pgfpathclose%
\pgfusepath{stroke}%
\end{pgfscope}%
\end{pgfpicture}%
\makeatother%
\endgroup%

%% file: plots/gp_posterior_ret_mean_hm.pgf
\begingroup%
\makeatletter%
\begin{pgfpicture}%
\pgfpathrectangle{\pgfpointorigin}{\pgfqpoint{1.750000in}{1.750000in}}%
\pgfusepath{use as bounding box, clip}%
\begin{pgfscope}%
\pgfsetbuttcap%
\pgfsetmiterjoin%
\definecolor{currentfill}{rgb}{1.000000,1.000000,1.000000}%
\pgfsetfillcolor{currentfill}%
\pgfsetlinewidth{0.000000pt}%
\definecolor{currentstroke}{rgb}{1.000000,1.000000,1.000000}%
\pgfsetstrokecolor{currentstroke}%
\pgfsetdash{}{0pt}%
\pgfpathmoveto{\pgfqpoint{0.000000in}{0.000000in}}%
\pgfpathlineto{\pgfqpoint{1.750000in}{0.000000in}}%
\pgfpathlineto{\pgfqpoint{1.750000in}{1.750000in}}%
\pgfpathlineto{\pgfqpoint{0.000000in}{1.750000in}}%
\pgfpathclose%
\pgfusepath{fill}%
\end{pgfscope}%
\begin{pgfscope}%
\pgfsetbuttcap%
\pgfsetmiterjoin%
\definecolor{currentfill}{rgb}{1.000000,1.000000,1.000000}%
\pgfsetfillcolor{currentfill}%
\pgfsetlinewidth{0.000000pt}%
\definecolor{currentstroke}{rgb}{0.000000,0.000000,0.000000}%
\pgfsetstrokecolor{currentstroke}%
\pgfsetstrokeopacity{0.000000}%
\pgfsetdash{}{0pt}%
\pgfpathmoveto{\pgfqpoint{0.533841in}{0.439824in}}%
\pgfpathlineto{\pgfqpoint{1.655502in}{0.439824in}}%
\pgfpathlineto{\pgfqpoint{1.655502in}{1.561484in}}%
\pgfpathlineto{\pgfqpoint{0.533841in}{1.561484in}}%
\pgfpathclose%
\pgfusepath{fill}%
\end{pgfscope}%
\begin{pgfscope}%
\pgfpathrectangle{\pgfqpoint{0.533841in}{0.439824in}}{\pgfqpoint{1.121661in}{1.121661in}}%
\pgfusepath{clip}%
\pgfsys@transformshift{0.533841in}{0.439824in}%
\pgftext[left,bottom]{\includegraphics[interpolate=true,width=1.122000in,height=1.122000in]{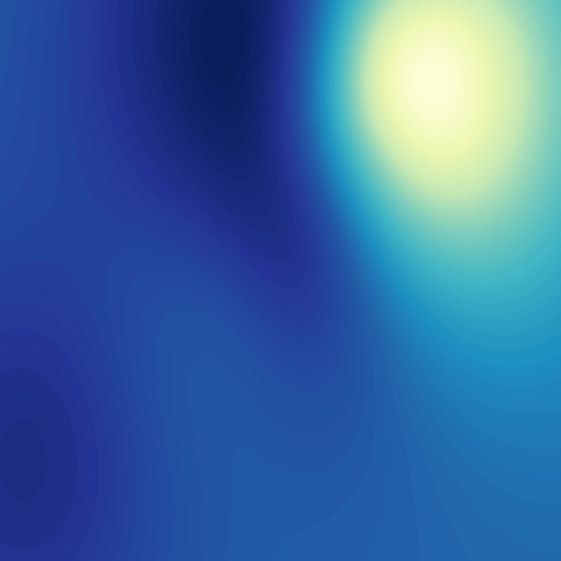}}%
\end{pgfscope}%
\begin{pgfscope}%
\pgfpathrectangle{\pgfqpoint{0.533841in}{0.439824in}}{\pgfqpoint{1.121661in}{1.121661in}}%
\pgfusepath{clip}%
\pgfsetbuttcap%
\pgfsetroundjoin%
\definecolor{currentfill}{rgb}{0.950000,0.950000,0.950000}%
\pgfsetfillcolor{currentfill}%
\pgfsetlinewidth{1.003750pt}%
\definecolor{currentstroke}{rgb}{0.950000,0.950000,0.950000}%
\pgfsetstrokecolor{currentstroke}%
\pgfsetdash{}{0pt}%
\pgfpathmoveto{\pgfqpoint{0.602322in}{0.675680in}}%
\pgfpathcurveto{\pgfqpoint{0.609455in}{0.675680in}}{\pgfqpoint{0.616297in}{0.678514in}}{\pgfqpoint{0.621340in}{0.683558in}}%
\pgfpathcurveto{\pgfqpoint{0.626384in}{0.688601in}}{\pgfqpoint{0.629218in}{0.695443in}}{\pgfqpoint{0.629218in}{0.702576in}}%
\pgfpathcurveto{\pgfqpoint{0.629218in}{0.709709in}}{\pgfqpoint{0.626384in}{0.716550in}}{\pgfqpoint{0.621340in}{0.721594in}}%
\pgfpathcurveto{\pgfqpoint{0.616297in}{0.726637in}}{\pgfqpoint{0.609455in}{0.729471in}}{\pgfqpoint{0.602322in}{0.729471in}}%
\pgfpathcurveto{\pgfqpoint{0.595189in}{0.729471in}}{\pgfqpoint{0.588348in}{0.726637in}}{\pgfqpoint{0.583304in}{0.721594in}}%
\pgfpathcurveto{\pgfqpoint{0.578260in}{0.716550in}}{\pgfqpoint{0.575427in}{0.709709in}}{\pgfqpoint{0.575427in}{0.702576in}}%
\pgfpathcurveto{\pgfqpoint{0.575427in}{0.695443in}}{\pgfqpoint{0.578260in}{0.688601in}}{\pgfqpoint{0.583304in}{0.683558in}}%
\pgfpathcurveto{\pgfqpoint{0.588348in}{0.678514in}}{\pgfqpoint{0.595189in}{0.675680in}}{\pgfqpoint{0.602322in}{0.675680in}}%
\pgfpathclose%
\pgfusepath{stroke,fill}%
\end{pgfscope}%
\begin{pgfscope}%
\pgfpathrectangle{\pgfqpoint{0.533841in}{0.439824in}}{\pgfqpoint{1.121661in}{1.121661in}}%
\pgfusepath{clip}%
\pgfsetbuttcap%
\pgfsetroundjoin%
\definecolor{currentfill}{rgb}{0.904118,0.904118,0.904118}%
\pgfsetfillcolor{currentfill}%
\pgfsetlinewidth{1.003750pt}%
\definecolor{currentstroke}{rgb}{0.904118,0.904118,0.904118}%
\pgfsetstrokecolor{currentstroke}%
\pgfsetdash{}{0pt}%
\pgfpathmoveto{\pgfqpoint{0.597175in}{0.533483in}}%
\pgfpathcurveto{\pgfqpoint{0.604308in}{0.533483in}}{\pgfqpoint{0.611150in}{0.536316in}}{\pgfqpoint{0.616194in}{0.541360in}}%
\pgfpathcurveto{\pgfqpoint{0.621237in}{0.546404in}}{\pgfqpoint{0.624071in}{0.553245in}}{\pgfqpoint{0.624071in}{0.560378in}}%
\pgfpathcurveto{\pgfqpoint{0.624071in}{0.567511in}}{\pgfqpoint{0.621237in}{0.574353in}}{\pgfqpoint{0.616194in}{0.579396in}}%
\pgfpathcurveto{\pgfqpoint{0.611150in}{0.584440in}}{\pgfqpoint{0.604308in}{0.587274in}}{\pgfqpoint{0.597175in}{0.587274in}}%
\pgfpathcurveto{\pgfqpoint{0.590043in}{0.587274in}}{\pgfqpoint{0.583201in}{0.584440in}}{\pgfqpoint{0.578157in}{0.579396in}}%
\pgfpathcurveto{\pgfqpoint{0.573114in}{0.574353in}}{\pgfqpoint{0.570280in}{0.567511in}}{\pgfqpoint{0.570280in}{0.560378in}}%
\pgfpathcurveto{\pgfqpoint{0.570280in}{0.553245in}}{\pgfqpoint{0.573114in}{0.546404in}}{\pgfqpoint{0.578157in}{0.541360in}}%
\pgfpathcurveto{\pgfqpoint{0.583201in}{0.536316in}}{\pgfqpoint{0.590043in}{0.533483in}}{\pgfqpoint{0.597175in}{0.533483in}}%
\pgfpathclose%
\pgfusepath{stroke,fill}%
\end{pgfscope}%
\begin{pgfscope}%
\pgfpathrectangle{\pgfqpoint{0.533841in}{0.439824in}}{\pgfqpoint{1.121661in}{1.121661in}}%
\pgfusepath{clip}%
\pgfsetbuttcap%
\pgfsetroundjoin%
\definecolor{currentfill}{rgb}{0.858235,0.858235,0.858235}%
\pgfsetfillcolor{currentfill}%
\pgfsetlinewidth{1.003750pt}%
\definecolor{currentstroke}{rgb}{0.858235,0.858235,0.858235}%
\pgfsetstrokecolor{currentstroke}%
\pgfsetdash{}{0pt}%
\pgfpathmoveto{\pgfqpoint{1.303412in}{1.364146in}}%
\pgfpathcurveto{\pgfqpoint{1.310545in}{1.364146in}}{\pgfqpoint{1.317386in}{1.366980in}}{\pgfqpoint{1.322430in}{1.372024in}}%
\pgfpathcurveto{\pgfqpoint{1.327474in}{1.377068in}}{\pgfqpoint{1.330308in}{1.383909in}}{\pgfqpoint{1.330308in}{1.391042in}}%
\pgfpathcurveto{\pgfqpoint{1.330308in}{1.398175in}}{\pgfqpoint{1.327474in}{1.405017in}}{\pgfqpoint{1.322430in}{1.410060in}}%
\pgfpathcurveto{\pgfqpoint{1.317386in}{1.415104in}}{\pgfqpoint{1.310545in}{1.417938in}}{\pgfqpoint{1.303412in}{1.417938in}}%
\pgfpathcurveto{\pgfqpoint{1.296279in}{1.417938in}}{\pgfqpoint{1.289437in}{1.415104in}}{\pgfqpoint{1.284394in}{1.410060in}}%
\pgfpathcurveto{\pgfqpoint{1.279350in}{1.405017in}}{\pgfqpoint{1.276516in}{1.398175in}}{\pgfqpoint{1.276516in}{1.391042in}}%
\pgfpathcurveto{\pgfqpoint{1.276516in}{1.383909in}}{\pgfqpoint{1.279350in}{1.377068in}}{\pgfqpoint{1.284394in}{1.372024in}}%
\pgfpathcurveto{\pgfqpoint{1.289437in}{1.366980in}}{\pgfqpoint{1.296279in}{1.364146in}}{\pgfqpoint{1.303412in}{1.364146in}}%
\pgfpathclose%
\pgfusepath{stroke,fill}%
\end{pgfscope}%
\begin{pgfscope}%
\pgfpathrectangle{\pgfqpoint{0.533841in}{0.439824in}}{\pgfqpoint{1.121661in}{1.121661in}}%
\pgfusepath{clip}%
\pgfsetbuttcap%
\pgfsetroundjoin%
\definecolor{currentfill}{rgb}{0.808824,0.808824,0.808824}%
\pgfsetfillcolor{currentfill}%
\pgfsetlinewidth{1.003750pt}%
\definecolor{currentstroke}{rgb}{0.808824,0.808824,0.808824}%
\pgfsetstrokecolor{currentstroke}%
\pgfsetdash{}{0pt}%
\pgfpathmoveto{\pgfqpoint{1.022847in}{0.967822in}}%
\pgfpathcurveto{\pgfqpoint{1.029980in}{0.967822in}}{\pgfqpoint{1.036822in}{0.970656in}}{\pgfqpoint{1.041866in}{0.975700in}}%
\pgfpathcurveto{\pgfqpoint{1.046909in}{0.980743in}}{\pgfqpoint{1.049743in}{0.987585in}}{\pgfqpoint{1.049743in}{0.994718in}}%
\pgfpathcurveto{\pgfqpoint{1.049743in}{1.001851in}}{\pgfqpoint{1.046909in}{1.008692in}}{\pgfqpoint{1.041866in}{1.013736in}}%
\pgfpathcurveto{\pgfqpoint{1.036822in}{1.018780in}}{\pgfqpoint{1.029980in}{1.021614in}}{\pgfqpoint{1.022847in}{1.021614in}}%
\pgfpathcurveto{\pgfqpoint{1.015715in}{1.021614in}}{\pgfqpoint{1.008873in}{1.018780in}}{\pgfqpoint{1.003829in}{1.013736in}}%
\pgfpathcurveto{\pgfqpoint{0.998786in}{1.008692in}}{\pgfqpoint{0.995952in}{1.001851in}}{\pgfqpoint{0.995952in}{0.994718in}}%
\pgfpathcurveto{\pgfqpoint{0.995952in}{0.987585in}}{\pgfqpoint{0.998786in}{0.980743in}}{\pgfqpoint{1.003829in}{0.975700in}}%
\pgfpathcurveto{\pgfqpoint{1.008873in}{0.970656in}}{\pgfqpoint{1.015715in}{0.967822in}}{\pgfqpoint{1.022847in}{0.967822in}}%
\pgfpathclose%
\pgfusepath{stroke,fill}%
\end{pgfscope}%
\begin{pgfscope}%
\pgfpathrectangle{\pgfqpoint{0.533841in}{0.439824in}}{\pgfqpoint{1.121661in}{1.121661in}}%
\pgfusepath{clip}%
\pgfsetbuttcap%
\pgfsetroundjoin%
\definecolor{currentfill}{rgb}{0.762941,0.762941,0.762941}%
\pgfsetfillcolor{currentfill}%
\pgfsetlinewidth{1.003750pt}%
\definecolor{currentstroke}{rgb}{0.762941,0.762941,0.762941}%
\pgfsetstrokecolor{currentstroke}%
\pgfsetdash{}{0pt}%
\pgfpathmoveto{\pgfqpoint{1.140620in}{1.072909in}}%
\pgfpathcurveto{\pgfqpoint{1.147753in}{1.072909in}}{\pgfqpoint{1.154595in}{1.075743in}}{\pgfqpoint{1.159638in}{1.080787in}}%
\pgfpathcurveto{\pgfqpoint{1.164682in}{1.085830in}}{\pgfqpoint{1.167516in}{1.092672in}}{\pgfqpoint{1.167516in}{1.099805in}}%
\pgfpathcurveto{\pgfqpoint{1.167516in}{1.106938in}}{\pgfqpoint{1.164682in}{1.113779in}}{\pgfqpoint{1.159638in}{1.118823in}}%
\pgfpathcurveto{\pgfqpoint{1.154595in}{1.123866in}}{\pgfqpoint{1.147753in}{1.126700in}}{\pgfqpoint{1.140620in}{1.126700in}}%
\pgfpathcurveto{\pgfqpoint{1.133487in}{1.126700in}}{\pgfqpoint{1.126646in}{1.123866in}}{\pgfqpoint{1.121602in}{1.118823in}}%
\pgfpathcurveto{\pgfqpoint{1.116558in}{1.113779in}}{\pgfqpoint{1.113724in}{1.106938in}}{\pgfqpoint{1.113724in}{1.099805in}}%
\pgfpathcurveto{\pgfqpoint{1.113724in}{1.092672in}}{\pgfqpoint{1.116558in}{1.085830in}}{\pgfqpoint{1.121602in}{1.080787in}}%
\pgfpathcurveto{\pgfqpoint{1.126646in}{1.075743in}}{\pgfqpoint{1.133487in}{1.072909in}}{\pgfqpoint{1.140620in}{1.072909in}}%
\pgfpathclose%
\pgfusepath{stroke,fill}%
\end{pgfscope}%
\begin{pgfscope}%
\pgfpathrectangle{\pgfqpoint{0.533841in}{0.439824in}}{\pgfqpoint{1.121661in}{1.121661in}}%
\pgfusepath{clip}%
\pgfsetbuttcap%
\pgfsetroundjoin%
\definecolor{currentfill}{rgb}{0.713529,0.713529,0.713529}%
\pgfsetfillcolor{currentfill}%
\pgfsetlinewidth{1.003750pt}%
\definecolor{currentstroke}{rgb}{0.713529,0.713529,0.713529}%
\pgfsetstrokecolor{currentstroke}%
\pgfsetdash{}{0pt}%
\pgfpathmoveto{\pgfqpoint{1.375108in}{1.480920in}}%
\pgfpathcurveto{\pgfqpoint{1.382241in}{1.480920in}}{\pgfqpoint{1.389083in}{1.483754in}}{\pgfqpoint{1.394127in}{1.488798in}}%
\pgfpathcurveto{\pgfqpoint{1.399170in}{1.493842in}}{\pgfqpoint{1.402004in}{1.500683in}}{\pgfqpoint{1.402004in}{1.507816in}}%
\pgfpathcurveto{\pgfqpoint{1.402004in}{1.514949in}}{\pgfqpoint{1.399170in}{1.521791in}}{\pgfqpoint{1.394127in}{1.526834in}}%
\pgfpathcurveto{\pgfqpoint{1.389083in}{1.531878in}}{\pgfqpoint{1.382241in}{1.534712in}}{\pgfqpoint{1.375108in}{1.534712in}}%
\pgfpathcurveto{\pgfqpoint{1.367976in}{1.534712in}}{\pgfqpoint{1.361134in}{1.531878in}}{\pgfqpoint{1.356090in}{1.526834in}}%
\pgfpathcurveto{\pgfqpoint{1.351047in}{1.521791in}}{\pgfqpoint{1.348213in}{1.514949in}}{\pgfqpoint{1.348213in}{1.507816in}}%
\pgfpathcurveto{\pgfqpoint{1.348213in}{1.500683in}}{\pgfqpoint{1.351047in}{1.493842in}}{\pgfqpoint{1.356090in}{1.488798in}}%
\pgfpathcurveto{\pgfqpoint{1.361134in}{1.483754in}}{\pgfqpoint{1.367976in}{1.480920in}}{\pgfqpoint{1.375108in}{1.480920in}}%
\pgfpathclose%
\pgfusepath{stroke,fill}%
\end{pgfscope}%
\begin{pgfscope}%
\pgfpathrectangle{\pgfqpoint{0.533841in}{0.439824in}}{\pgfqpoint{1.121661in}{1.121661in}}%
\pgfusepath{clip}%
\pgfsetbuttcap%
\pgfsetroundjoin%
\definecolor{currentfill}{rgb}{0.667647,0.667647,0.667647}%
\pgfsetfillcolor{currentfill}%
\pgfsetlinewidth{1.003750pt}%
\definecolor{currentstroke}{rgb}{0.667647,0.667647,0.667647}%
\pgfsetstrokecolor{currentstroke}%
\pgfsetdash{}{0pt}%
\pgfpathmoveto{\pgfqpoint{1.166055in}{1.460878in}}%
\pgfpathcurveto{\pgfqpoint{1.173188in}{1.460878in}}{\pgfqpoint{1.180030in}{1.463712in}}{\pgfqpoint{1.185073in}{1.468756in}}%
\pgfpathcurveto{\pgfqpoint{1.190117in}{1.473799in}}{\pgfqpoint{1.192951in}{1.480641in}}{\pgfqpoint{1.192951in}{1.487774in}}%
\pgfpathcurveto{\pgfqpoint{1.192951in}{1.494907in}}{\pgfqpoint{1.190117in}{1.501748in}}{\pgfqpoint{1.185073in}{1.506792in}}%
\pgfpathcurveto{\pgfqpoint{1.180030in}{1.511836in}}{\pgfqpoint{1.173188in}{1.514670in}}{\pgfqpoint{1.166055in}{1.514670in}}%
\pgfpathcurveto{\pgfqpoint{1.158922in}{1.514670in}}{\pgfqpoint{1.152081in}{1.511836in}}{\pgfqpoint{1.147037in}{1.506792in}}%
\pgfpathcurveto{\pgfqpoint{1.141993in}{1.501748in}}{\pgfqpoint{1.139159in}{1.494907in}}{\pgfqpoint{1.139159in}{1.487774in}}%
\pgfpathcurveto{\pgfqpoint{1.139159in}{1.480641in}}{\pgfqpoint{1.141993in}{1.473799in}}{\pgfqpoint{1.147037in}{1.468756in}}%
\pgfpathcurveto{\pgfqpoint{1.152081in}{1.463712in}}{\pgfqpoint{1.158922in}{1.460878in}}{\pgfqpoint{1.166055in}{1.460878in}}%
\pgfpathclose%
\pgfusepath{stroke,fill}%
\end{pgfscope}%
\begin{pgfscope}%
\pgfpathrectangle{\pgfqpoint{0.533841in}{0.439824in}}{\pgfqpoint{1.121661in}{1.121661in}}%
\pgfusepath{clip}%
\pgfsetbuttcap%
\pgfsetroundjoin%
\definecolor{currentfill}{rgb}{0.618235,0.618235,0.618235}%
\pgfsetfillcolor{currentfill}%
\pgfsetlinewidth{1.003750pt}%
\definecolor{currentstroke}{rgb}{0.618235,0.618235,0.618235}%
\pgfsetstrokecolor{currentstroke}%
\pgfsetdash{}{0pt}%
\pgfpathmoveto{\pgfqpoint{1.384576in}{1.293828in}}%
\pgfpathcurveto{\pgfqpoint{1.391709in}{1.293828in}}{\pgfqpoint{1.398551in}{1.296662in}}{\pgfqpoint{1.403594in}{1.301705in}}%
\pgfpathcurveto{\pgfqpoint{1.408638in}{1.306749in}}{\pgfqpoint{1.411472in}{1.313591in}}{\pgfqpoint{1.411472in}{1.320723in}}%
\pgfpathcurveto{\pgfqpoint{1.411472in}{1.327856in}}{\pgfqpoint{1.408638in}{1.334698in}}{\pgfqpoint{1.403594in}{1.339742in}}%
\pgfpathcurveto{\pgfqpoint{1.398551in}{1.344785in}}{\pgfqpoint{1.391709in}{1.347619in}}{\pgfqpoint{1.384576in}{1.347619in}}%
\pgfpathcurveto{\pgfqpoint{1.377443in}{1.347619in}}{\pgfqpoint{1.370602in}{1.344785in}}{\pgfqpoint{1.365558in}{1.339742in}}%
\pgfpathcurveto{\pgfqpoint{1.360514in}{1.334698in}}{\pgfqpoint{1.357681in}{1.327856in}}{\pgfqpoint{1.357681in}{1.320723in}}%
\pgfpathcurveto{\pgfqpoint{1.357681in}{1.313591in}}{\pgfqpoint{1.360514in}{1.306749in}}{\pgfqpoint{1.365558in}{1.301705in}}%
\pgfpathcurveto{\pgfqpoint{1.370602in}{1.296662in}}{\pgfqpoint{1.377443in}{1.293828in}}{\pgfqpoint{1.384576in}{1.293828in}}%
\pgfpathclose%
\pgfusepath{stroke,fill}%
\end{pgfscope}%
\begin{pgfscope}%
\pgfpathrectangle{\pgfqpoint{0.533841in}{0.439824in}}{\pgfqpoint{1.121661in}{1.121661in}}%
\pgfusepath{clip}%
\pgfsetbuttcap%
\pgfsetroundjoin%
\definecolor{currentfill}{rgb}{0.572353,0.572353,0.572353}%
\pgfsetfillcolor{currentfill}%
\pgfsetlinewidth{1.003750pt}%
\definecolor{currentstroke}{rgb}{0.572353,0.572353,0.572353}%
\pgfsetstrokecolor{currentstroke}%
\pgfsetdash{}{0pt}%
\pgfpathmoveto{\pgfqpoint{1.353022in}{1.441099in}}%
\pgfpathcurveto{\pgfqpoint{1.360155in}{1.441099in}}{\pgfqpoint{1.366997in}{1.443933in}}{\pgfqpoint{1.372041in}{1.448977in}}%
\pgfpathcurveto{\pgfqpoint{1.377084in}{1.454020in}}{\pgfqpoint{1.379918in}{1.460862in}}{\pgfqpoint{1.379918in}{1.467995in}}%
\pgfpathcurveto{\pgfqpoint{1.379918in}{1.475128in}}{\pgfqpoint{1.377084in}{1.481969in}}{\pgfqpoint{1.372041in}{1.487013in}}%
\pgfpathcurveto{\pgfqpoint{1.366997in}{1.492057in}}{\pgfqpoint{1.360155in}{1.494891in}}{\pgfqpoint{1.353022in}{1.494891in}}%
\pgfpathcurveto{\pgfqpoint{1.345890in}{1.494891in}}{\pgfqpoint{1.339048in}{1.492057in}}{\pgfqpoint{1.334004in}{1.487013in}}%
\pgfpathcurveto{\pgfqpoint{1.328961in}{1.481969in}}{\pgfqpoint{1.326127in}{1.475128in}}{\pgfqpoint{1.326127in}{1.467995in}}%
\pgfpathcurveto{\pgfqpoint{1.326127in}{1.460862in}}{\pgfqpoint{1.328961in}{1.454020in}}{\pgfqpoint{1.334004in}{1.448977in}}%
\pgfpathcurveto{\pgfqpoint{1.339048in}{1.443933in}}{\pgfqpoint{1.345890in}{1.441099in}}{\pgfqpoint{1.353022in}{1.441099in}}%
\pgfpathclose%
\pgfusepath{stroke,fill}%
\end{pgfscope}%
\begin{pgfscope}%
\pgfpathrectangle{\pgfqpoint{0.533841in}{0.439824in}}{\pgfqpoint{1.121661in}{1.121661in}}%
\pgfusepath{clip}%
\pgfsetbuttcap%
\pgfsetroundjoin%
\definecolor{currentfill}{rgb}{0.522941,0.522941,0.522941}%
\pgfsetfillcolor{currentfill}%
\pgfsetlinewidth{1.003750pt}%
\definecolor{currentstroke}{rgb}{0.522941,0.522941,0.522941}%
\pgfsetstrokecolor{currentstroke}%
\pgfsetdash{}{0pt}%
\pgfpathmoveto{\pgfqpoint{1.331989in}{1.290084in}}%
\pgfpathcurveto{\pgfqpoint{1.339122in}{1.290084in}}{\pgfqpoint{1.345963in}{1.292918in}}{\pgfqpoint{1.351007in}{1.297962in}}%
\pgfpathcurveto{\pgfqpoint{1.356051in}{1.303005in}}{\pgfqpoint{1.358885in}{1.309847in}}{\pgfqpoint{1.358885in}{1.316980in}}%
\pgfpathcurveto{\pgfqpoint{1.358885in}{1.324113in}}{\pgfqpoint{1.356051in}{1.330954in}}{\pgfqpoint{1.351007in}{1.335998in}}%
\pgfpathcurveto{\pgfqpoint{1.345963in}{1.341042in}}{\pgfqpoint{1.339122in}{1.343876in}}{\pgfqpoint{1.331989in}{1.343876in}}%
\pgfpathcurveto{\pgfqpoint{1.324856in}{1.343876in}}{\pgfqpoint{1.318014in}{1.341042in}}{\pgfqpoint{1.312971in}{1.335998in}}%
\pgfpathcurveto{\pgfqpoint{1.307927in}{1.330954in}}{\pgfqpoint{1.305093in}{1.324113in}}{\pgfqpoint{1.305093in}{1.316980in}}%
\pgfpathcurveto{\pgfqpoint{1.305093in}{1.309847in}}{\pgfqpoint{1.307927in}{1.303005in}}{\pgfqpoint{1.312971in}{1.297962in}}%
\pgfpathcurveto{\pgfqpoint{1.318014in}{1.292918in}}{\pgfqpoint{1.324856in}{1.290084in}}{\pgfqpoint{1.331989in}{1.290084in}}%
\pgfpathclose%
\pgfusepath{stroke,fill}%
\end{pgfscope}%
\begin{pgfscope}%
\pgfpathrectangle{\pgfqpoint{0.533841in}{0.439824in}}{\pgfqpoint{1.121661in}{1.121661in}}%
\pgfusepath{clip}%
\pgfsetbuttcap%
\pgfsetroundjoin%
\definecolor{currentfill}{rgb}{0.477059,0.477059,0.477059}%
\pgfsetfillcolor{currentfill}%
\pgfsetlinewidth{1.003750pt}%
\definecolor{currentstroke}{rgb}{0.477059,0.477059,0.477059}%
\pgfsetstrokecolor{currentstroke}%
\pgfsetdash{}{0pt}%
\pgfpathmoveto{\pgfqpoint{1.385635in}{1.341838in}}%
\pgfpathcurveto{\pgfqpoint{1.392768in}{1.341838in}}{\pgfqpoint{1.399609in}{1.344672in}}{\pgfqpoint{1.404653in}{1.349715in}}%
\pgfpathcurveto{\pgfqpoint{1.409697in}{1.354759in}}{\pgfqpoint{1.412531in}{1.361601in}}{\pgfqpoint{1.412531in}{1.368734in}}%
\pgfpathcurveto{\pgfqpoint{1.412531in}{1.375866in}}{\pgfqpoint{1.409697in}{1.382708in}}{\pgfqpoint{1.404653in}{1.387752in}}%
\pgfpathcurveto{\pgfqpoint{1.399609in}{1.392795in}}{\pgfqpoint{1.392768in}{1.395629in}}{\pgfqpoint{1.385635in}{1.395629in}}%
\pgfpathcurveto{\pgfqpoint{1.378502in}{1.395629in}}{\pgfqpoint{1.371661in}{1.392795in}}{\pgfqpoint{1.366617in}{1.387752in}}%
\pgfpathcurveto{\pgfqpoint{1.361573in}{1.382708in}}{\pgfqpoint{1.358739in}{1.375866in}}{\pgfqpoint{1.358739in}{1.368734in}}%
\pgfpathcurveto{\pgfqpoint{1.358739in}{1.361601in}}{\pgfqpoint{1.361573in}{1.354759in}}{\pgfqpoint{1.366617in}{1.349715in}}%
\pgfpathcurveto{\pgfqpoint{1.371661in}{1.344672in}}{\pgfqpoint{1.378502in}{1.341838in}}{\pgfqpoint{1.385635in}{1.341838in}}%
\pgfpathclose%
\pgfusepath{stroke,fill}%
\end{pgfscope}%
\begin{pgfscope}%
\pgfpathrectangle{\pgfqpoint{0.533841in}{0.439824in}}{\pgfqpoint{1.121661in}{1.121661in}}%
\pgfusepath{clip}%
\pgfsetbuttcap%
\pgfsetroundjoin%
\definecolor{currentfill}{rgb}{0.427647,0.427647,0.427647}%
\pgfsetfillcolor{currentfill}%
\pgfsetlinewidth{1.003750pt}%
\definecolor{currentstroke}{rgb}{0.427647,0.427647,0.427647}%
\pgfsetstrokecolor{currentstroke}%
\pgfsetdash{}{0pt}%
\pgfpathmoveto{\pgfqpoint{1.362098in}{1.361664in}}%
\pgfpathcurveto{\pgfqpoint{1.369230in}{1.361664in}}{\pgfqpoint{1.376072in}{1.364498in}}{\pgfqpoint{1.381116in}{1.369542in}}%
\pgfpathcurveto{\pgfqpoint{1.386159in}{1.374585in}}{\pgfqpoint{1.388993in}{1.381427in}}{\pgfqpoint{1.388993in}{1.388560in}}%
\pgfpathcurveto{\pgfqpoint{1.388993in}{1.395693in}}{\pgfqpoint{1.386159in}{1.402534in}}{\pgfqpoint{1.381116in}{1.407578in}}%
\pgfpathcurveto{\pgfqpoint{1.376072in}{1.412622in}}{\pgfqpoint{1.369230in}{1.415456in}}{\pgfqpoint{1.362098in}{1.415456in}}%
\pgfpathcurveto{\pgfqpoint{1.354965in}{1.415456in}}{\pgfqpoint{1.348123in}{1.412622in}}{\pgfqpoint{1.343079in}{1.407578in}}%
\pgfpathcurveto{\pgfqpoint{1.338036in}{1.402534in}}{\pgfqpoint{1.335202in}{1.395693in}}{\pgfqpoint{1.335202in}{1.388560in}}%
\pgfpathcurveto{\pgfqpoint{1.335202in}{1.381427in}}{\pgfqpoint{1.338036in}{1.374585in}}{\pgfqpoint{1.343079in}{1.369542in}}%
\pgfpathcurveto{\pgfqpoint{1.348123in}{1.364498in}}{\pgfqpoint{1.354965in}{1.361664in}}{\pgfqpoint{1.362098in}{1.361664in}}%
\pgfpathclose%
\pgfusepath{stroke,fill}%
\end{pgfscope}%
\begin{pgfscope}%
\pgfpathrectangle{\pgfqpoint{0.533841in}{0.439824in}}{\pgfqpoint{1.121661in}{1.121661in}}%
\pgfusepath{clip}%
\pgfsetbuttcap%
\pgfsetroundjoin%
\definecolor{currentfill}{rgb}{0.381765,0.381765,0.381765}%
\pgfsetfillcolor{currentfill}%
\pgfsetlinewidth{1.003750pt}%
\definecolor{currentstroke}{rgb}{0.381765,0.381765,0.381765}%
\pgfsetstrokecolor{currentstroke}%
\pgfsetdash{}{0pt}%
\pgfpathmoveto{\pgfqpoint{1.354785in}{1.363419in}}%
\pgfpathcurveto{\pgfqpoint{1.361918in}{1.363419in}}{\pgfqpoint{1.368760in}{1.366253in}}{\pgfqpoint{1.373803in}{1.371296in}}%
\pgfpathcurveto{\pgfqpoint{1.378847in}{1.376340in}}{\pgfqpoint{1.381681in}{1.383181in}}{\pgfqpoint{1.381681in}{1.390314in}}%
\pgfpathcurveto{\pgfqpoint{1.381681in}{1.397447in}}{\pgfqpoint{1.378847in}{1.404289in}}{\pgfqpoint{1.373803in}{1.409332in}}%
\pgfpathcurveto{\pgfqpoint{1.368760in}{1.414376in}}{\pgfqpoint{1.361918in}{1.417210in}}{\pgfqpoint{1.354785in}{1.417210in}}%
\pgfpathcurveto{\pgfqpoint{1.347652in}{1.417210in}}{\pgfqpoint{1.340811in}{1.414376in}}{\pgfqpoint{1.335767in}{1.409332in}}%
\pgfpathcurveto{\pgfqpoint{1.330723in}{1.404289in}}{\pgfqpoint{1.327889in}{1.397447in}}{\pgfqpoint{1.327889in}{1.390314in}}%
\pgfpathcurveto{\pgfqpoint{1.327889in}{1.383181in}}{\pgfqpoint{1.330723in}{1.376340in}}{\pgfqpoint{1.335767in}{1.371296in}}%
\pgfpathcurveto{\pgfqpoint{1.340811in}{1.366253in}}{\pgfqpoint{1.347652in}{1.363419in}}{\pgfqpoint{1.354785in}{1.363419in}}%
\pgfpathclose%
\pgfusepath{stroke,fill}%
\end{pgfscope}%
\begin{pgfscope}%
\pgfpathrectangle{\pgfqpoint{0.533841in}{0.439824in}}{\pgfqpoint{1.121661in}{1.121661in}}%
\pgfusepath{clip}%
\pgfsetbuttcap%
\pgfsetroundjoin%
\definecolor{currentfill}{rgb}{0.332353,0.332353,0.332353}%
\pgfsetfillcolor{currentfill}%
\pgfsetlinewidth{1.003750pt}%
\definecolor{currentstroke}{rgb}{0.332353,0.332353,0.332353}%
\pgfsetstrokecolor{currentstroke}%
\pgfsetdash{}{0pt}%
\pgfpathmoveto{\pgfqpoint{1.336500in}{1.265410in}}%
\pgfpathcurveto{\pgfqpoint{1.343633in}{1.265410in}}{\pgfqpoint{1.350475in}{1.268244in}}{\pgfqpoint{1.355518in}{1.273288in}}%
\pgfpathcurveto{\pgfqpoint{1.360562in}{1.278332in}}{\pgfqpoint{1.363396in}{1.285173in}}{\pgfqpoint{1.363396in}{1.292306in}}%
\pgfpathcurveto{\pgfqpoint{1.363396in}{1.299439in}}{\pgfqpoint{1.360562in}{1.306281in}}{\pgfqpoint{1.355518in}{1.311324in}}%
\pgfpathcurveto{\pgfqpoint{1.350475in}{1.316368in}}{\pgfqpoint{1.343633in}{1.319202in}}{\pgfqpoint{1.336500in}{1.319202in}}%
\pgfpathcurveto{\pgfqpoint{1.329367in}{1.319202in}}{\pgfqpoint{1.322526in}{1.316368in}}{\pgfqpoint{1.317482in}{1.311324in}}%
\pgfpathcurveto{\pgfqpoint{1.312438in}{1.306281in}}{\pgfqpoint{1.309604in}{1.299439in}}{\pgfqpoint{1.309604in}{1.292306in}}%
\pgfpathcurveto{\pgfqpoint{1.309604in}{1.285173in}}{\pgfqpoint{1.312438in}{1.278332in}}{\pgfqpoint{1.317482in}{1.273288in}}%
\pgfpathcurveto{\pgfqpoint{1.322526in}{1.268244in}}{\pgfqpoint{1.329367in}{1.265410in}}{\pgfqpoint{1.336500in}{1.265410in}}%
\pgfpathclose%
\pgfusepath{stroke,fill}%
\end{pgfscope}%
\begin{pgfscope}%
\pgfpathrectangle{\pgfqpoint{0.533841in}{0.439824in}}{\pgfqpoint{1.121661in}{1.121661in}}%
\pgfusepath{clip}%
\pgfsetbuttcap%
\pgfsetroundjoin%
\definecolor{currentfill}{rgb}{0.286471,0.286471,0.286471}%
\pgfsetfillcolor{currentfill}%
\pgfsetlinewidth{1.003750pt}%
\definecolor{currentstroke}{rgb}{0.286471,0.286471,0.286471}%
\pgfsetstrokecolor{currentstroke}%
\pgfsetdash{}{0pt}%
\pgfpathmoveto{\pgfqpoint{1.418524in}{1.389484in}}%
\pgfpathcurveto{\pgfqpoint{1.425657in}{1.389484in}}{\pgfqpoint{1.432498in}{1.392318in}}{\pgfqpoint{1.437542in}{1.397361in}}%
\pgfpathcurveto{\pgfqpoint{1.442586in}{1.402405in}}{\pgfqpoint{1.445420in}{1.409247in}}{\pgfqpoint{1.445420in}{1.416379in}}%
\pgfpathcurveto{\pgfqpoint{1.445420in}{1.423512in}}{\pgfqpoint{1.442586in}{1.430354in}}{\pgfqpoint{1.437542in}{1.435398in}}%
\pgfpathcurveto{\pgfqpoint{1.432498in}{1.440441in}}{\pgfqpoint{1.425657in}{1.443275in}}{\pgfqpoint{1.418524in}{1.443275in}}%
\pgfpathcurveto{\pgfqpoint{1.411391in}{1.443275in}}{\pgfqpoint{1.404549in}{1.440441in}}{\pgfqpoint{1.399506in}{1.435398in}}%
\pgfpathcurveto{\pgfqpoint{1.394462in}{1.430354in}}{\pgfqpoint{1.391628in}{1.423512in}}{\pgfqpoint{1.391628in}{1.416379in}}%
\pgfpathcurveto{\pgfqpoint{1.391628in}{1.409247in}}{\pgfqpoint{1.394462in}{1.402405in}}{\pgfqpoint{1.399506in}{1.397361in}}%
\pgfpathcurveto{\pgfqpoint{1.404549in}{1.392318in}}{\pgfqpoint{1.411391in}{1.389484in}}{\pgfqpoint{1.418524in}{1.389484in}}%
\pgfpathclose%
\pgfusepath{stroke,fill}%
\end{pgfscope}%
\begin{pgfscope}%
\pgfpathrectangle{\pgfqpoint{0.533841in}{0.439824in}}{\pgfqpoint{1.121661in}{1.121661in}}%
\pgfusepath{clip}%
\pgfsetbuttcap%
\pgfsetroundjoin%
\definecolor{currentfill}{rgb}{0.237059,0.237059,0.237059}%
\pgfsetfillcolor{currentfill}%
\pgfsetlinewidth{1.003750pt}%
\definecolor{currentstroke}{rgb}{0.237059,0.237059,0.237059}%
\pgfsetstrokecolor{currentstroke}%
\pgfsetdash{}{0pt}%
\pgfpathmoveto{\pgfqpoint{1.364340in}{1.420282in}}%
\pgfpathcurveto{\pgfqpoint{1.371473in}{1.420282in}}{\pgfqpoint{1.378314in}{1.423115in}}{\pgfqpoint{1.383358in}{1.428159in}}%
\pgfpathcurveto{\pgfqpoint{1.388402in}{1.433203in}}{\pgfqpoint{1.391235in}{1.440044in}}{\pgfqpoint{1.391235in}{1.447177in}}%
\pgfpathcurveto{\pgfqpoint{1.391235in}{1.454310in}}{\pgfqpoint{1.388402in}{1.461152in}}{\pgfqpoint{1.383358in}{1.466195in}}%
\pgfpathcurveto{\pgfqpoint{1.378314in}{1.471239in}}{\pgfqpoint{1.371473in}{1.474073in}}{\pgfqpoint{1.364340in}{1.474073in}}%
\pgfpathcurveto{\pgfqpoint{1.357207in}{1.474073in}}{\pgfqpoint{1.350365in}{1.471239in}}{\pgfqpoint{1.345322in}{1.466195in}}%
\pgfpathcurveto{\pgfqpoint{1.340278in}{1.461152in}}{\pgfqpoint{1.337444in}{1.454310in}}{\pgfqpoint{1.337444in}{1.447177in}}%
\pgfpathcurveto{\pgfqpoint{1.337444in}{1.440044in}}{\pgfqpoint{1.340278in}{1.433203in}}{\pgfqpoint{1.345322in}{1.428159in}}%
\pgfpathcurveto{\pgfqpoint{1.350365in}{1.423115in}}{\pgfqpoint{1.357207in}{1.420282in}}{\pgfqpoint{1.364340in}{1.420282in}}%
\pgfpathclose%
\pgfusepath{stroke,fill}%
\end{pgfscope}%
\begin{pgfscope}%
\pgfpathrectangle{\pgfqpoint{0.533841in}{0.439824in}}{\pgfqpoint{1.121661in}{1.121661in}}%
\pgfusepath{clip}%
\pgfsetbuttcap%
\pgfsetroundjoin%
\definecolor{currentfill}{rgb}{0.191176,0.191176,0.191176}%
\pgfsetfillcolor{currentfill}%
\pgfsetlinewidth{1.003750pt}%
\definecolor{currentstroke}{rgb}{0.191176,0.191176,0.191176}%
\pgfsetstrokecolor{currentstroke}%
\pgfsetdash{}{0pt}%
\pgfpathmoveto{\pgfqpoint{1.393312in}{1.283009in}}%
\pgfpathcurveto{\pgfqpoint{1.400445in}{1.283009in}}{\pgfqpoint{1.407287in}{1.285843in}}{\pgfqpoint{1.412330in}{1.290887in}}%
\pgfpathcurveto{\pgfqpoint{1.417374in}{1.295930in}}{\pgfqpoint{1.420208in}{1.302772in}}{\pgfqpoint{1.420208in}{1.309905in}}%
\pgfpathcurveto{\pgfqpoint{1.420208in}{1.317038in}}{\pgfqpoint{1.417374in}{1.323879in}}{\pgfqpoint{1.412330in}{1.328923in}}%
\pgfpathcurveto{\pgfqpoint{1.407287in}{1.333967in}}{\pgfqpoint{1.400445in}{1.336801in}}{\pgfqpoint{1.393312in}{1.336801in}}%
\pgfpathcurveto{\pgfqpoint{1.386179in}{1.336801in}}{\pgfqpoint{1.379338in}{1.333967in}}{\pgfqpoint{1.374294in}{1.328923in}}%
\pgfpathcurveto{\pgfqpoint{1.369250in}{1.323879in}}{\pgfqpoint{1.366417in}{1.317038in}}{\pgfqpoint{1.366417in}{1.309905in}}%
\pgfpathcurveto{\pgfqpoint{1.366417in}{1.302772in}}{\pgfqpoint{1.369250in}{1.295930in}}{\pgfqpoint{1.374294in}{1.290887in}}%
\pgfpathcurveto{\pgfqpoint{1.379338in}{1.285843in}}{\pgfqpoint{1.386179in}{1.283009in}}{\pgfqpoint{1.393312in}{1.283009in}}%
\pgfpathclose%
\pgfusepath{stroke,fill}%
\end{pgfscope}%
\begin{pgfscope}%
\pgfpathrectangle{\pgfqpoint{0.533841in}{0.439824in}}{\pgfqpoint{1.121661in}{1.121661in}}%
\pgfusepath{clip}%
\pgfsetbuttcap%
\pgfsetroundjoin%
\definecolor{currentfill}{rgb}{0.141765,0.141765,0.141765}%
\pgfsetfillcolor{currentfill}%
\pgfsetlinewidth{1.003750pt}%
\definecolor{currentstroke}{rgb}{0.141765,0.141765,0.141765}%
\pgfsetstrokecolor{currentstroke}%
\pgfsetdash{}{0pt}%
\pgfpathmoveto{\pgfqpoint{1.393917in}{1.416676in}}%
\pgfpathcurveto{\pgfqpoint{1.401050in}{1.416676in}}{\pgfqpoint{1.407892in}{1.419510in}}{\pgfqpoint{1.412936in}{1.424553in}}%
\pgfpathcurveto{\pgfqpoint{1.417979in}{1.429597in}}{\pgfqpoint{1.420813in}{1.436439in}}{\pgfqpoint{1.420813in}{1.443571in}}%
\pgfpathcurveto{\pgfqpoint{1.420813in}{1.450704in}}{\pgfqpoint{1.417979in}{1.457546in}}{\pgfqpoint{1.412936in}{1.462589in}}%
\pgfpathcurveto{\pgfqpoint{1.407892in}{1.467633in}}{\pgfqpoint{1.401050in}{1.470467in}}{\pgfqpoint{1.393917in}{1.470467in}}%
\pgfpathcurveto{\pgfqpoint{1.386785in}{1.470467in}}{\pgfqpoint{1.379943in}{1.467633in}}{\pgfqpoint{1.374899in}{1.462589in}}%
\pgfpathcurveto{\pgfqpoint{1.369856in}{1.457546in}}{\pgfqpoint{1.367022in}{1.450704in}}{\pgfqpoint{1.367022in}{1.443571in}}%
\pgfpathcurveto{\pgfqpoint{1.367022in}{1.436439in}}{\pgfqpoint{1.369856in}{1.429597in}}{\pgfqpoint{1.374899in}{1.424553in}}%
\pgfpathcurveto{\pgfqpoint{1.379943in}{1.419510in}}{\pgfqpoint{1.386785in}{1.416676in}}{\pgfqpoint{1.393917in}{1.416676in}}%
\pgfpathclose%
\pgfusepath{stroke,fill}%
\end{pgfscope}%
\begin{pgfscope}%
\pgfpathrectangle{\pgfqpoint{0.533841in}{0.439824in}}{\pgfqpoint{1.121661in}{1.121661in}}%
\pgfusepath{clip}%
\pgfsetbuttcap%
\pgfsetroundjoin%
\definecolor{currentfill}{rgb}{0.095882,0.095882,0.095882}%
\pgfsetfillcolor{currentfill}%
\pgfsetlinewidth{1.003750pt}%
\definecolor{currentstroke}{rgb}{0.095882,0.095882,0.095882}%
\pgfsetstrokecolor{currentstroke}%
\pgfsetdash{}{0pt}%
\pgfpathmoveto{\pgfqpoint{1.479725in}{1.303254in}}%
\pgfpathcurveto{\pgfqpoint{1.486858in}{1.303254in}}{\pgfqpoint{1.493699in}{1.306088in}}{\pgfqpoint{1.498743in}{1.311132in}}%
\pgfpathcurveto{\pgfqpoint{1.503787in}{1.316175in}}{\pgfqpoint{1.506620in}{1.323017in}}{\pgfqpoint{1.506620in}{1.330150in}}%
\pgfpathcurveto{\pgfqpoint{1.506620in}{1.337283in}}{\pgfqpoint{1.503787in}{1.344124in}}{\pgfqpoint{1.498743in}{1.349168in}}%
\pgfpathcurveto{\pgfqpoint{1.493699in}{1.354212in}}{\pgfqpoint{1.486858in}{1.357046in}}{\pgfqpoint{1.479725in}{1.357046in}}%
\pgfpathcurveto{\pgfqpoint{1.472592in}{1.357046in}}{\pgfqpoint{1.465750in}{1.354212in}}{\pgfqpoint{1.460707in}{1.349168in}}%
\pgfpathcurveto{\pgfqpoint{1.455663in}{1.344124in}}{\pgfqpoint{1.452829in}{1.337283in}}{\pgfqpoint{1.452829in}{1.330150in}}%
\pgfpathcurveto{\pgfqpoint{1.452829in}{1.323017in}}{\pgfqpoint{1.455663in}{1.316175in}}{\pgfqpoint{1.460707in}{1.311132in}}%
\pgfpathcurveto{\pgfqpoint{1.465750in}{1.306088in}}{\pgfqpoint{1.472592in}{1.303254in}}{\pgfqpoint{1.479725in}{1.303254in}}%
\pgfpathclose%
\pgfusepath{stroke,fill}%
\end{pgfscope}%
\begin{pgfscope}%
\pgfpathrectangle{\pgfqpoint{0.533841in}{0.439824in}}{\pgfqpoint{1.121661in}{1.121661in}}%
\pgfusepath{clip}%
\pgfsetbuttcap%
\pgfsetroundjoin%
\definecolor{currentfill}{rgb}{0.050000,0.050000,0.050000}%
\pgfsetfillcolor{currentfill}%
\pgfsetlinewidth{1.003750pt}%
\definecolor{currentstroke}{rgb}{0.050000,0.050000,0.050000}%
\pgfsetstrokecolor{currentstroke}%
\pgfsetdash{}{0pt}%
\pgfpathmoveto{\pgfqpoint{1.428598in}{1.458574in}}%
\pgfpathcurveto{\pgfqpoint{1.435731in}{1.458574in}}{\pgfqpoint{1.442573in}{1.461408in}}{\pgfqpoint{1.447616in}{1.466452in}}%
\pgfpathcurveto{\pgfqpoint{1.452660in}{1.471495in}}{\pgfqpoint{1.455494in}{1.478337in}}{\pgfqpoint{1.455494in}{1.485470in}}%
\pgfpathcurveto{\pgfqpoint{1.455494in}{1.492602in}}{\pgfqpoint{1.452660in}{1.499444in}}{\pgfqpoint{1.447616in}{1.504488in}}%
\pgfpathcurveto{\pgfqpoint{1.442573in}{1.509531in}}{\pgfqpoint{1.435731in}{1.512365in}}{\pgfqpoint{1.428598in}{1.512365in}}%
\pgfpathcurveto{\pgfqpoint{1.421465in}{1.512365in}}{\pgfqpoint{1.414624in}{1.509531in}}{\pgfqpoint{1.409580in}{1.504488in}}%
\pgfpathcurveto{\pgfqpoint{1.404536in}{1.499444in}}{\pgfqpoint{1.401702in}{1.492602in}}{\pgfqpoint{1.401702in}{1.485470in}}%
\pgfpathcurveto{\pgfqpoint{1.401702in}{1.478337in}}{\pgfqpoint{1.404536in}{1.471495in}}{\pgfqpoint{1.409580in}{1.466452in}}%
\pgfpathcurveto{\pgfqpoint{1.414624in}{1.461408in}}{\pgfqpoint{1.421465in}{1.458574in}}{\pgfqpoint{1.428598in}{1.458574in}}%
\pgfpathclose%
\pgfusepath{stroke,fill}%
\end{pgfscope}%
\begin{pgfscope}%
\pgfpathrectangle{\pgfqpoint{0.533841in}{0.439824in}}{\pgfqpoint{1.121661in}{1.121661in}}%
\pgfusepath{clip}%
\pgfsetbuttcap%
\pgfsetroundjoin%
\definecolor{currentfill}{rgb}{1.000000,0.549020,0.000000}%
\pgfsetfillcolor{currentfill}%
\pgfsetlinewidth{1.003750pt}%
\definecolor{currentstroke}{rgb}{1.000000,0.549020,0.000000}%
\pgfsetstrokecolor{currentstroke}%
\pgfsetdash{}{0pt}%
\pgfsys@defobject{currentmarker}{\pgfqpoint{-0.051159in}{-0.043518in}}{\pgfqpoint{0.051159in}{0.053791in}}{%
\pgfpathmoveto{\pgfqpoint{0.000000in}{0.053791in}}%
\pgfpathlineto{\pgfqpoint{-0.012077in}{0.016622in}}%
\pgfpathlineto{\pgfqpoint{-0.051159in}{0.016622in}}%
\pgfpathlineto{\pgfqpoint{-0.019541in}{-0.006349in}}%
\pgfpathlineto{\pgfqpoint{-0.031618in}{-0.043518in}}%
\pgfpathlineto{\pgfqpoint{-0.000000in}{-0.020546in}}%
\pgfpathlineto{\pgfqpoint{0.031618in}{-0.043518in}}%
\pgfpathlineto{\pgfqpoint{0.019541in}{-0.006349in}}%
\pgfpathlineto{\pgfqpoint{0.051159in}{0.016622in}}%
\pgfpathlineto{\pgfqpoint{0.012077in}{0.016622in}}%
\pgfpathclose%
\pgfusepath{stroke,fill}%
}%
\begin{pgfscope}%
\pgfsys@transformshift{1.399177in}{1.395400in}%
\pgfsys@useobject{currentmarker}{}%
\end{pgfscope}%
\end{pgfscope}%
\begin{pgfscope}%
\pgfpathrectangle{\pgfqpoint{0.533841in}{0.439824in}}{\pgfqpoint{1.121661in}{1.121661in}}%
\pgfusepath{clip}%
\pgfsetbuttcap%
\pgfsetroundjoin%
\definecolor{currentfill}{rgb}{0.698039,0.133333,0.133333}%
\pgfsetfillcolor{currentfill}%
\pgfsetlinewidth{1.003750pt}%
\definecolor{currentstroke}{rgb}{0.698039,0.133333,0.133333}%
\pgfsetstrokecolor{currentstroke}%
\pgfsetdash{}{0pt}%
\pgfsys@defobject{currentmarker}{\pgfqpoint{-0.051159in}{-0.043518in}}{\pgfqpoint{0.051159in}{0.053791in}}{%
\pgfpathmoveto{\pgfqpoint{0.000000in}{0.053791in}}%
\pgfpathlineto{\pgfqpoint{-0.012077in}{0.016622in}}%
\pgfpathlineto{\pgfqpoint{-0.051159in}{0.016622in}}%
\pgfpathlineto{\pgfqpoint{-0.019541in}{-0.006349in}}%
\pgfpathlineto{\pgfqpoint{-0.031618in}{-0.043518in}}%
\pgfpathlineto{\pgfqpoint{-0.000000in}{-0.020546in}}%
\pgfpathlineto{\pgfqpoint{0.031618in}{-0.043518in}}%
\pgfpathlineto{\pgfqpoint{0.019541in}{-0.006349in}}%
\pgfpathlineto{\pgfqpoint{0.051159in}{0.016622in}}%
\pgfpathlineto{\pgfqpoint{0.012077in}{0.016622in}}%
\pgfpathclose%
\pgfusepath{stroke,fill}%
}%
\begin{pgfscope}%
\pgfsys@transformshift{1.375087in}{1.281069in}%
\pgfsys@useobject{currentmarker}{}%
\end{pgfscope}%
\end{pgfscope}%
\begin{pgfscope}%
\pgfsetbuttcap%
\pgfsetroundjoin%
\definecolor{currentfill}{rgb}{0.000000,0.000000,0.000000}%
\pgfsetfillcolor{currentfill}%
\pgfsetlinewidth{0.803000pt}%
\definecolor{currentstroke}{rgb}{0.000000,0.000000,0.000000}%
\pgfsetstrokecolor{currentstroke}%
\pgfsetdash{}{0pt}%
\pgfsys@defobject{currentmarker}{\pgfqpoint{0.000000in}{-0.048611in}}{\pgfqpoint{0.000000in}{0.000000in}}{%
\pgfpathmoveto{\pgfqpoint{0.000000in}{0.000000in}}%
\pgfpathlineto{\pgfqpoint{0.000000in}{-0.048611in}}%
\pgfusepath{stroke,fill}%
}%
\begin{pgfscope}%
\pgfsys@transformshift{0.627313in}{0.439824in}%
\pgfsys@useobject{currentmarker}{}%
\end{pgfscope}%
\end{pgfscope}%
\begin{pgfscope}%
\definecolor{textcolor}{rgb}{0.000000,0.000000,0.000000}%
\pgfsetstrokecolor{textcolor}%
\pgfsetfillcolor{textcolor}%
\pgftext[x=0.627313in,y=0.342602in,,top]{\color{textcolor}\rmfamily\fontsize{9.000000}{10.800000}\selectfont \(\displaystyle {0.020}\)}%
\end{pgfscope}%
\begin{pgfscope}%
\pgfsetbuttcap%
\pgfsetroundjoin%
\definecolor{currentfill}{rgb}{0.000000,0.000000,0.000000}%
\pgfsetfillcolor{currentfill}%
\pgfsetlinewidth{0.803000pt}%
\definecolor{currentstroke}{rgb}{0.000000,0.000000,0.000000}%
\pgfsetstrokecolor{currentstroke}%
\pgfsetdash{}{0pt}%
\pgfsys@defobject{currentmarker}{\pgfqpoint{0.000000in}{-0.048611in}}{\pgfqpoint{0.000000in}{0.000000in}}{%
\pgfpathmoveto{\pgfqpoint{0.000000in}{0.000000in}}%
\pgfpathlineto{\pgfqpoint{0.000000in}{-0.048611in}}%
\pgfusepath{stroke,fill}%
}%
\begin{pgfscope}%
\pgfsys@transformshift{1.094672in}{0.439824in}%
\pgfsys@useobject{currentmarker}{}%
\end{pgfscope}%
\end{pgfscope}%
\begin{pgfscope}%
\definecolor{textcolor}{rgb}{0.000000,0.000000,0.000000}%
\pgfsetstrokecolor{textcolor}%
\pgfsetfillcolor{textcolor}%
\pgftext[x=1.094672in,y=0.342602in,,top]{\color{textcolor}\rmfamily\fontsize{9.000000}{10.800000}\selectfont \(\displaystyle {0.024}\)}%
\end{pgfscope}%
\begin{pgfscope}%
\pgfsetbuttcap%
\pgfsetroundjoin%
\definecolor{currentfill}{rgb}{0.000000,0.000000,0.000000}%
\pgfsetfillcolor{currentfill}%
\pgfsetlinewidth{0.803000pt}%
\definecolor{currentstroke}{rgb}{0.000000,0.000000,0.000000}%
\pgfsetstrokecolor{currentstroke}%
\pgfsetdash{}{0pt}%
\pgfsys@defobject{currentmarker}{\pgfqpoint{0.000000in}{-0.048611in}}{\pgfqpoint{0.000000in}{0.000000in}}{%
\pgfpathmoveto{\pgfqpoint{0.000000in}{0.000000in}}%
\pgfpathlineto{\pgfqpoint{0.000000in}{-0.048611in}}%
\pgfusepath{stroke,fill}%
}%
\begin{pgfscope}%
\pgfsys@transformshift{1.562030in}{0.439824in}%
\pgfsys@useobject{currentmarker}{}%
\end{pgfscope}%
\end{pgfscope}%
\begin{pgfscope}%
\definecolor{textcolor}{rgb}{0.000000,0.000000,0.000000}%
\pgfsetstrokecolor{textcolor}%
\pgfsetfillcolor{textcolor}%
\pgftext[x=1.562030in,y=0.342602in,,top]{\color{textcolor}\rmfamily\fontsize{9.000000}{10.800000}\selectfont \(\displaystyle {0.028}\)}%
\end{pgfscope}%
\begin{pgfscope}%
\definecolor{textcolor}{rgb}{0.000000,0.000000,0.000000}%
\pgfsetstrokecolor{textcolor}%
\pgfsetfillcolor{textcolor}%
\pgftext[x=1.094672in,y=0.176046in,,top]{\color{textcolor}\rmfamily\fontsize{9.000000}{10.800000}\selectfont pend. pole mass \(\displaystyle m_p\)}%
\end{pgfscope}%
\begin{pgfscope}%
\pgfsetbuttcap%
\pgfsetroundjoin%
\definecolor{currentfill}{rgb}{0.000000,0.000000,0.000000}%
\pgfsetfillcolor{currentfill}%
\pgfsetlinewidth{0.803000pt}%
\definecolor{currentstroke}{rgb}{0.000000,0.000000,0.000000}%
\pgfsetstrokecolor{currentstroke}%
\pgfsetdash{}{0pt}%
\pgfsys@defobject{currentmarker}{\pgfqpoint{-0.048611in}{0.000000in}}{\pgfqpoint{-0.000000in}{0.000000in}}{%
\pgfpathmoveto{\pgfqpoint{-0.000000in}{0.000000in}}%
\pgfpathlineto{\pgfqpoint{-0.048611in}{0.000000in}}%
\pgfusepath{stroke,fill}%
}%
\begin{pgfscope}%
\pgfsys@transformshift{0.533841in}{0.557893in}%
\pgfsys@useobject{currentmarker}{}%
\end{pgfscope}%
\end{pgfscope}%
\begin{pgfscope}%
\definecolor{textcolor}{rgb}{0.000000,0.000000,0.000000}%
\pgfsetstrokecolor{textcolor}%
\pgfsetfillcolor{textcolor}%
\pgftext[x=0.208225in, y=0.514518in, left, base]{\color{textcolor}\rmfamily\fontsize{9.000000}{10.800000}\selectfont \(\displaystyle {0.08}\)}%
\end{pgfscope}%
\begin{pgfscope}%
\pgfsetbuttcap%
\pgfsetroundjoin%
\definecolor{currentfill}{rgb}{0.000000,0.000000,0.000000}%
\pgfsetfillcolor{currentfill}%
\pgfsetlinewidth{0.803000pt}%
\definecolor{currentstroke}{rgb}{0.000000,0.000000,0.000000}%
\pgfsetstrokecolor{currentstroke}%
\pgfsetdash{}{0pt}%
\pgfsys@defobject{currentmarker}{\pgfqpoint{-0.048611in}{0.000000in}}{\pgfqpoint{-0.000000in}{0.000000in}}{%
\pgfpathmoveto{\pgfqpoint{-0.000000in}{0.000000in}}%
\pgfpathlineto{\pgfqpoint{-0.048611in}{0.000000in}}%
\pgfusepath{stroke,fill}%
}%
\begin{pgfscope}%
\pgfsys@transformshift{0.533841in}{0.853067in}%
\pgfsys@useobject{currentmarker}{}%
\end{pgfscope}%
\end{pgfscope}%
\begin{pgfscope}%
\definecolor{textcolor}{rgb}{0.000000,0.000000,0.000000}%
\pgfsetstrokecolor{textcolor}%
\pgfsetfillcolor{textcolor}%
\pgftext[x=0.208225in, y=0.809692in, left, base]{\color{textcolor}\rmfamily\fontsize{9.000000}{10.800000}\selectfont \(\displaystyle {0.09}\)}%
\end{pgfscope}%
\begin{pgfscope}%
\pgfsetbuttcap%
\pgfsetroundjoin%
\definecolor{currentfill}{rgb}{0.000000,0.000000,0.000000}%
\pgfsetfillcolor{currentfill}%
\pgfsetlinewidth{0.803000pt}%
\definecolor{currentstroke}{rgb}{0.000000,0.000000,0.000000}%
\pgfsetstrokecolor{currentstroke}%
\pgfsetdash{}{0pt}%
\pgfsys@defobject{currentmarker}{\pgfqpoint{-0.048611in}{0.000000in}}{\pgfqpoint{-0.000000in}{0.000000in}}{%
\pgfpathmoveto{\pgfqpoint{-0.000000in}{0.000000in}}%
\pgfpathlineto{\pgfqpoint{-0.048611in}{0.000000in}}%
\pgfusepath{stroke,fill}%
}%
\begin{pgfscope}%
\pgfsys@transformshift{0.533841in}{1.148241in}%
\pgfsys@useobject{currentmarker}{}%
\end{pgfscope}%
\end{pgfscope}%
\begin{pgfscope}%
\definecolor{textcolor}{rgb}{0.000000,0.000000,0.000000}%
\pgfsetstrokecolor{textcolor}%
\pgfsetfillcolor{textcolor}%
\pgftext[x=0.208225in, y=1.104866in, left, base]{\color{textcolor}\rmfamily\fontsize{9.000000}{10.800000}\selectfont \(\displaystyle {0.10}\)}%
\end{pgfscope}%
\begin{pgfscope}%
\pgfsetbuttcap%
\pgfsetroundjoin%
\definecolor{currentfill}{rgb}{0.000000,0.000000,0.000000}%
\pgfsetfillcolor{currentfill}%
\pgfsetlinewidth{0.803000pt}%
\definecolor{currentstroke}{rgb}{0.000000,0.000000,0.000000}%
\pgfsetstrokecolor{currentstroke}%
\pgfsetdash{}{0pt}%
\pgfsys@defobject{currentmarker}{\pgfqpoint{-0.048611in}{0.000000in}}{\pgfqpoint{-0.000000in}{0.000000in}}{%
\pgfpathmoveto{\pgfqpoint{-0.000000in}{0.000000in}}%
\pgfpathlineto{\pgfqpoint{-0.048611in}{0.000000in}}%
\pgfusepath{stroke,fill}%
}%
\begin{pgfscope}%
\pgfsys@transformshift{0.533841in}{1.443415in}%
\pgfsys@useobject{currentmarker}{}%
\end{pgfscope}%
\end{pgfscope}%
\begin{pgfscope}%
\definecolor{textcolor}{rgb}{0.000000,0.000000,0.000000}%
\pgfsetstrokecolor{textcolor}%
\pgfsetfillcolor{textcolor}%
\pgftext[x=0.208225in, y=1.400040in, left, base]{\color{textcolor}\rmfamily\fontsize{9.000000}{10.800000}\selectfont \(\displaystyle {0.11}\)}%
\end{pgfscope}%
\begin{pgfscope}%
\definecolor{textcolor}{rgb}{0.000000,0.000000,0.000000}%
\pgfsetstrokecolor{textcolor}%
\pgfsetfillcolor{textcolor}%
\pgftext[x=0.152670in,y=1.000654in,,bottom,rotate=90.000000]{\color{textcolor}\rmfamily\fontsize{9.000000}{10.800000}\selectfont rot. pole mass \(\displaystyle m_r\)}%
\end{pgfscope}%
\begin{pgfscope}%
\pgfsetrectcap%
\pgfsetmiterjoin%
\pgfsetlinewidth{0.803000pt}%
\definecolor{currentstroke}{rgb}{0.000000,0.000000,0.000000}%
\pgfsetstrokecolor{currentstroke}%
\pgfsetdash{}{0pt}%
\pgfpathmoveto{\pgfqpoint{0.533841in}{0.439824in}}%
\pgfpathlineto{\pgfqpoint{0.533841in}{1.561484in}}%
\pgfusepath{stroke}%
\end{pgfscope}%
\begin{pgfscope}%
\pgfsetrectcap%
\pgfsetmiterjoin%
\pgfsetlinewidth{0.803000pt}%
\definecolor{currentstroke}{rgb}{0.000000,0.000000,0.000000}%
\pgfsetstrokecolor{currentstroke}%
\pgfsetdash{}{0pt}%
\pgfpathmoveto{\pgfqpoint{1.655502in}{0.439824in}}%
\pgfpathlineto{\pgfqpoint{1.655502in}{1.561484in}}%
\pgfusepath{stroke}%
\end{pgfscope}%
\begin{pgfscope}%
\pgfsetrectcap%
\pgfsetmiterjoin%
\pgfsetlinewidth{0.803000pt}%
\definecolor{currentstroke}{rgb}{0.000000,0.000000,0.000000}%
\pgfsetstrokecolor{currentstroke}%
\pgfsetdash{}{0pt}%
\pgfpathmoveto{\pgfqpoint{0.533841in}{0.439824in}}%
\pgfpathlineto{\pgfqpoint{1.655502in}{0.439824in}}%
\pgfusepath{stroke}%
\end{pgfscope}%
\begin{pgfscope}%
\pgfsetrectcap%
\pgfsetmiterjoin%
\pgfsetlinewidth{0.803000pt}%
\definecolor{currentstroke}{rgb}{0.000000,0.000000,0.000000}%
\pgfsetstrokecolor{currentstroke}%
\pgfsetdash{}{0pt}%
\pgfpathmoveto{\pgfqpoint{0.533841in}{1.561484in}}%
\pgfpathlineto{\pgfqpoint{1.655502in}{1.561484in}}%
\pgfusepath{stroke}%
\end{pgfscope}%
\end{pgfpicture}%
\makeatother%
\endgroup%

%% file: plots/gp_posterior_ret_std_cb.pgf
\begingroup%
\makeatletter%
\begin{pgfpicture}%
\pgfpathrectangle{\pgfpointorigin}{\pgfqpoint{1.155000in}{0.420000in}}%
\pgfusepath{use as bounding box, clip}%
\begin{pgfscope}%
\pgfsetbuttcap%
\pgfsetmiterjoin%
\definecolor{currentfill}{rgb}{1.000000,1.000000,1.000000}%
\pgfsetfillcolor{currentfill}%
\pgfsetlinewidth{0.000000pt}%
\definecolor{currentstroke}{rgb}{1.000000,1.000000,1.000000}%
\pgfsetstrokecolor{currentstroke}%
\pgfsetdash{}{0pt}%
\pgfpathmoveto{\pgfqpoint{0.000000in}{0.000000in}}%
\pgfpathlineto{\pgfqpoint{1.155000in}{0.000000in}}%
\pgfpathlineto{\pgfqpoint{1.155000in}{0.420000in}}%
\pgfpathlineto{\pgfqpoint{0.000000in}{0.420000in}}%
\pgfpathclose%
\pgfusepath{fill}%
\end{pgfscope}%
\begin{pgfscope}%
\pgfpathrectangle{\pgfqpoint{0.041670in}{0.041670in}}{\pgfqpoint{1.071660in}{0.128438in}}%
\pgfusepath{clip}%
\pgfsetbuttcap%
\pgfsetmiterjoin%
\definecolor{currentfill}{rgb}{1.000000,1.000000,1.000000}%
\pgfsetfillcolor{currentfill}%
\pgfsetlinewidth{0.010037pt}%
\definecolor{currentstroke}{rgb}{1.000000,1.000000,1.000000}%
\pgfsetstrokecolor{currentstroke}%
\pgfsetdash{}{0pt}%
\pgfpathmoveto{\pgfqpoint{0.041670in}{0.041670in}}%
\pgfpathlineto{\pgfqpoint{0.052387in}{0.041670in}}%
\pgfpathlineto{\pgfqpoint{1.102613in}{0.041670in}}%
\pgfpathlineto{\pgfqpoint{1.113330in}{0.041670in}}%
\pgfpathlineto{\pgfqpoint{1.113330in}{0.170108in}}%
\pgfpathlineto{\pgfqpoint{1.102613in}{0.170108in}}%
\pgfpathlineto{\pgfqpoint{0.052387in}{0.170108in}}%
\pgfpathlineto{\pgfqpoint{0.041670in}{0.170108in}}%
\pgfpathclose%
\pgfusepath{stroke,fill}%
\end{pgfscope}%
\begin{pgfscope}%
\pgfsys@transformshift{0.042000in}{0.042000in}%
\pgftext[left,bottom]{\includegraphics[interpolate=true,width=1.072000in,height=0.128000in]{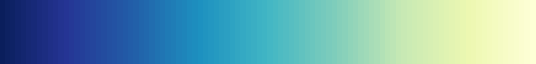}}%
\end{pgfscope}%
\begin{pgfscope}%
\pgfsetbuttcap%
\pgfsetroundjoin%
\definecolor{currentfill}{rgb}{0.000000,0.000000,0.000000}%
\pgfsetfillcolor{currentfill}%
\pgfsetlinewidth{0.803000pt}%
\definecolor{currentstroke}{rgb}{0.000000,0.000000,0.000000}%
\pgfsetstrokecolor{currentstroke}%
\pgfsetdash{}{0pt}%
\pgfsys@defobject{currentmarker}{\pgfqpoint{0.000000in}{0.000000in}}{\pgfqpoint{0.000000in}{0.048611in}}{%
\pgfpathmoveto{\pgfqpoint{0.000000in}{0.000000in}}%
\pgfpathlineto{\pgfqpoint{0.000000in}{0.048611in}}%
\pgfusepath{stroke,fill}%
}%
\begin{pgfscope}%
\pgfsys@transformshift{0.439436in}{0.170108in}%
\pgfsys@useobject{currentmarker}{}%
\end{pgfscope}%
\end{pgfscope}%
\begin{pgfscope}%
\definecolor{textcolor}{rgb}{0.000000,0.000000,0.000000}%
\pgfsetstrokecolor{textcolor}%
\pgfsetfillcolor{textcolor}%
\pgftext[x=0.439436in,y=0.267330in,,bottom]{\color{textcolor}\rmfamily\fontsize{9.000000}{10.800000}\selectfont \(\displaystyle {100}\)}%
\end{pgfscope}%
\begin{pgfscope}%
\pgfsetbuttcap%
\pgfsetroundjoin%
\definecolor{currentfill}{rgb}{0.000000,0.000000,0.000000}%
\pgfsetfillcolor{currentfill}%
\pgfsetlinewidth{0.803000pt}%
\definecolor{currentstroke}{rgb}{0.000000,0.000000,0.000000}%
\pgfsetstrokecolor{currentstroke}%
\pgfsetdash{}{0pt}%
\pgfsys@defobject{currentmarker}{\pgfqpoint{0.000000in}{0.000000in}}{\pgfqpoint{0.000000in}{0.048611in}}{%
\pgfpathmoveto{\pgfqpoint{0.000000in}{0.000000in}}%
\pgfpathlineto{\pgfqpoint{0.000000in}{0.048611in}}%
\pgfusepath{stroke,fill}%
}%
\begin{pgfscope}%
\pgfsys@transformshift{0.910366in}{0.170108in}%
\pgfsys@useobject{currentmarker}{}%
\end{pgfscope}%
\end{pgfscope}%
\begin{pgfscope}%
\definecolor{textcolor}{rgb}{0.000000,0.000000,0.000000}%
\pgfsetstrokecolor{textcolor}%
\pgfsetfillcolor{textcolor}%
\pgftext[x=0.910366in,y=0.267330in,,bottom]{\color{textcolor}\rmfamily\fontsize{9.000000}{10.800000}\selectfont \(\displaystyle {200}\)}%
\end{pgfscope}%
\begin{pgfscope}%
\pgfsetbuttcap%
\pgfsetmiterjoin%
\pgfsetlinewidth{0.803000pt}%
\definecolor{currentstroke}{rgb}{0.000000,0.000000,0.000000}%
\pgfsetstrokecolor{currentstroke}%
\pgfsetdash{}{0pt}%
\pgfpathmoveto{\pgfqpoint{0.041670in}{0.041670in}}%
\pgfpathlineto{\pgfqpoint{0.052387in}{0.041670in}}%
\pgfpathlineto{\pgfqpoint{1.102613in}{0.041670in}}%
\pgfpathlineto{\pgfqpoint{1.113330in}{0.041670in}}%
\pgfpathlineto{\pgfqpoint{1.113330in}{0.170108in}}%
\pgfpathlineto{\pgfqpoint{1.102613in}{0.170108in}}%
\pgfpathlineto{\pgfqpoint{0.052387in}{0.170108in}}%
\pgfpathlineto{\pgfqpoint{0.041670in}{0.170108in}}%
\pgfpathclose%
\pgfusepath{stroke}%
\end{pgfscope}%
\end{pgfpicture}%
\makeatother%
\endgroup%

%% file: plots/gp_posterior_ret_std_hm.pgf
\begingroup%
\makeatletter%
\begin{pgfpicture}%
\pgfpathrectangle{\pgfpointorigin}{\pgfqpoint{1.750000in}{1.750000in}}%
\pgfusepath{use as bounding box, clip}%
\begin{pgfscope}%
\pgfsetbuttcap%
\pgfsetmiterjoin%
\definecolor{currentfill}{rgb}{1.000000,1.000000,1.000000}%
\pgfsetfillcolor{currentfill}%
\pgfsetlinewidth{0.000000pt}%
\definecolor{currentstroke}{rgb}{1.000000,1.000000,1.000000}%
\pgfsetstrokecolor{currentstroke}%
\pgfsetdash{}{0pt}%
\pgfpathmoveto{\pgfqpoint{0.000000in}{0.000000in}}%
\pgfpathlineto{\pgfqpoint{1.750000in}{0.000000in}}%
\pgfpathlineto{\pgfqpoint{1.750000in}{1.750000in}}%
\pgfpathlineto{\pgfqpoint{0.000000in}{1.750000in}}%
\pgfpathclose%
\pgfusepath{fill}%
\end{pgfscope}%
\begin{pgfscope}%
\pgfsetbuttcap%
\pgfsetmiterjoin%
\definecolor{currentfill}{rgb}{1.000000,1.000000,1.000000}%
\pgfsetfillcolor{currentfill}%
\pgfsetlinewidth{0.000000pt}%
\definecolor{currentstroke}{rgb}{0.000000,0.000000,0.000000}%
\pgfsetstrokecolor{currentstroke}%
\pgfsetstrokeopacity{0.000000}%
\pgfsetdash{}{0pt}%
\pgfpathmoveto{\pgfqpoint{0.533841in}{0.439824in}}%
\pgfpathlineto{\pgfqpoint{1.655502in}{0.439824in}}%
\pgfpathlineto{\pgfqpoint{1.655502in}{1.561484in}}%
\pgfpathlineto{\pgfqpoint{0.533841in}{1.561484in}}%
\pgfpathclose%
\pgfusepath{fill}%
\end{pgfscope}%
\begin{pgfscope}%
\pgfpathrectangle{\pgfqpoint{0.533841in}{0.439824in}}{\pgfqpoint{1.121661in}{1.121661in}}%
\pgfusepath{clip}%
\pgfsys@transformshift{0.533841in}{0.439824in}%
\pgftext[left,bottom]{\includegraphics[interpolate=true,width=1.122000in,height=1.122000in]{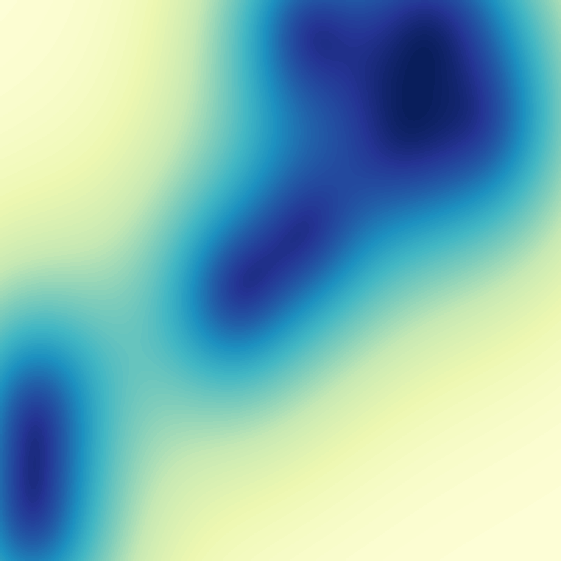}}%
\end{pgfscope}%
\begin{pgfscope}%
\pgfpathrectangle{\pgfqpoint{0.533841in}{0.439824in}}{\pgfqpoint{1.121661in}{1.121661in}}%
\pgfusepath{clip}%
\pgfsetbuttcap%
\pgfsetroundjoin%
\definecolor{currentfill}{rgb}{0.950000,0.950000,0.950000}%
\pgfsetfillcolor{currentfill}%
\pgfsetlinewidth{1.003750pt}%
\definecolor{currentstroke}{rgb}{0.950000,0.950000,0.950000}%
\pgfsetstrokecolor{currentstroke}%
\pgfsetdash{}{0pt}%
\pgfpathmoveto{\pgfqpoint{0.602322in}{0.675680in}}%
\pgfpathcurveto{\pgfqpoint{0.609455in}{0.675680in}}{\pgfqpoint{0.616297in}{0.678514in}}{\pgfqpoint{0.621340in}{0.683558in}}%
\pgfpathcurveto{\pgfqpoint{0.626384in}{0.688601in}}{\pgfqpoint{0.629218in}{0.695443in}}{\pgfqpoint{0.629218in}{0.702576in}}%
\pgfpathcurveto{\pgfqpoint{0.629218in}{0.709709in}}{\pgfqpoint{0.626384in}{0.716550in}}{\pgfqpoint{0.621340in}{0.721594in}}%
\pgfpathcurveto{\pgfqpoint{0.616297in}{0.726637in}}{\pgfqpoint{0.609455in}{0.729471in}}{\pgfqpoint{0.602322in}{0.729471in}}%
\pgfpathcurveto{\pgfqpoint{0.595189in}{0.729471in}}{\pgfqpoint{0.588348in}{0.726637in}}{\pgfqpoint{0.583304in}{0.721594in}}%
\pgfpathcurveto{\pgfqpoint{0.578260in}{0.716550in}}{\pgfqpoint{0.575427in}{0.709709in}}{\pgfqpoint{0.575427in}{0.702576in}}%
\pgfpathcurveto{\pgfqpoint{0.575427in}{0.695443in}}{\pgfqpoint{0.578260in}{0.688601in}}{\pgfqpoint{0.583304in}{0.683558in}}%
\pgfpathcurveto{\pgfqpoint{0.588348in}{0.678514in}}{\pgfqpoint{0.595189in}{0.675680in}}{\pgfqpoint{0.602322in}{0.675680in}}%
\pgfpathclose%
\pgfusepath{stroke,fill}%
\end{pgfscope}%
\begin{pgfscope}%
\pgfpathrectangle{\pgfqpoint{0.533841in}{0.439824in}}{\pgfqpoint{1.121661in}{1.121661in}}%
\pgfusepath{clip}%
\pgfsetbuttcap%
\pgfsetroundjoin%
\definecolor{currentfill}{rgb}{0.904118,0.904118,0.904118}%
\pgfsetfillcolor{currentfill}%
\pgfsetlinewidth{1.003750pt}%
\definecolor{currentstroke}{rgb}{0.904118,0.904118,0.904118}%
\pgfsetstrokecolor{currentstroke}%
\pgfsetdash{}{0pt}%
\pgfpathmoveto{\pgfqpoint{0.597175in}{0.533483in}}%
\pgfpathcurveto{\pgfqpoint{0.604308in}{0.533483in}}{\pgfqpoint{0.611150in}{0.536316in}}{\pgfqpoint{0.616194in}{0.541360in}}%
\pgfpathcurveto{\pgfqpoint{0.621237in}{0.546404in}}{\pgfqpoint{0.624071in}{0.553245in}}{\pgfqpoint{0.624071in}{0.560378in}}%
\pgfpathcurveto{\pgfqpoint{0.624071in}{0.567511in}}{\pgfqpoint{0.621237in}{0.574353in}}{\pgfqpoint{0.616194in}{0.579396in}}%
\pgfpathcurveto{\pgfqpoint{0.611150in}{0.584440in}}{\pgfqpoint{0.604308in}{0.587274in}}{\pgfqpoint{0.597175in}{0.587274in}}%
\pgfpathcurveto{\pgfqpoint{0.590043in}{0.587274in}}{\pgfqpoint{0.583201in}{0.584440in}}{\pgfqpoint{0.578157in}{0.579396in}}%
\pgfpathcurveto{\pgfqpoint{0.573114in}{0.574353in}}{\pgfqpoint{0.570280in}{0.567511in}}{\pgfqpoint{0.570280in}{0.560378in}}%
\pgfpathcurveto{\pgfqpoint{0.570280in}{0.553245in}}{\pgfqpoint{0.573114in}{0.546404in}}{\pgfqpoint{0.578157in}{0.541360in}}%
\pgfpathcurveto{\pgfqpoint{0.583201in}{0.536316in}}{\pgfqpoint{0.590043in}{0.533483in}}{\pgfqpoint{0.597175in}{0.533483in}}%
\pgfpathclose%
\pgfusepath{stroke,fill}%
\end{pgfscope}%
\begin{pgfscope}%
\pgfpathrectangle{\pgfqpoint{0.533841in}{0.439824in}}{\pgfqpoint{1.121661in}{1.121661in}}%
\pgfusepath{clip}%
\pgfsetbuttcap%
\pgfsetroundjoin%
\definecolor{currentfill}{rgb}{0.858235,0.858235,0.858235}%
\pgfsetfillcolor{currentfill}%
\pgfsetlinewidth{1.003750pt}%
\definecolor{currentstroke}{rgb}{0.858235,0.858235,0.858235}%
\pgfsetstrokecolor{currentstroke}%
\pgfsetdash{}{0pt}%
\pgfpathmoveto{\pgfqpoint{1.303412in}{1.364146in}}%
\pgfpathcurveto{\pgfqpoint{1.310545in}{1.364146in}}{\pgfqpoint{1.317386in}{1.366980in}}{\pgfqpoint{1.322430in}{1.372024in}}%
\pgfpathcurveto{\pgfqpoint{1.327474in}{1.377068in}}{\pgfqpoint{1.330308in}{1.383909in}}{\pgfqpoint{1.330308in}{1.391042in}}%
\pgfpathcurveto{\pgfqpoint{1.330308in}{1.398175in}}{\pgfqpoint{1.327474in}{1.405017in}}{\pgfqpoint{1.322430in}{1.410060in}}%
\pgfpathcurveto{\pgfqpoint{1.317386in}{1.415104in}}{\pgfqpoint{1.310545in}{1.417938in}}{\pgfqpoint{1.303412in}{1.417938in}}%
\pgfpathcurveto{\pgfqpoint{1.296279in}{1.417938in}}{\pgfqpoint{1.289437in}{1.415104in}}{\pgfqpoint{1.284394in}{1.410060in}}%
\pgfpathcurveto{\pgfqpoint{1.279350in}{1.405017in}}{\pgfqpoint{1.276516in}{1.398175in}}{\pgfqpoint{1.276516in}{1.391042in}}%
\pgfpathcurveto{\pgfqpoint{1.276516in}{1.383909in}}{\pgfqpoint{1.279350in}{1.377068in}}{\pgfqpoint{1.284394in}{1.372024in}}%
\pgfpathcurveto{\pgfqpoint{1.289437in}{1.366980in}}{\pgfqpoint{1.296279in}{1.364146in}}{\pgfqpoint{1.303412in}{1.364146in}}%
\pgfpathclose%
\pgfusepath{stroke,fill}%
\end{pgfscope}%
\begin{pgfscope}%
\pgfpathrectangle{\pgfqpoint{0.533841in}{0.439824in}}{\pgfqpoint{1.121661in}{1.121661in}}%
\pgfusepath{clip}%
\pgfsetbuttcap%
\pgfsetroundjoin%
\definecolor{currentfill}{rgb}{0.808824,0.808824,0.808824}%
\pgfsetfillcolor{currentfill}%
\pgfsetlinewidth{1.003750pt}%
\definecolor{currentstroke}{rgb}{0.808824,0.808824,0.808824}%
\pgfsetstrokecolor{currentstroke}%
\pgfsetdash{}{0pt}%
\pgfpathmoveto{\pgfqpoint{1.022847in}{0.967822in}}%
\pgfpathcurveto{\pgfqpoint{1.029980in}{0.967822in}}{\pgfqpoint{1.036822in}{0.970656in}}{\pgfqpoint{1.041866in}{0.975700in}}%
\pgfpathcurveto{\pgfqpoint{1.046909in}{0.980743in}}{\pgfqpoint{1.049743in}{0.987585in}}{\pgfqpoint{1.049743in}{0.994718in}}%
\pgfpathcurveto{\pgfqpoint{1.049743in}{1.001851in}}{\pgfqpoint{1.046909in}{1.008692in}}{\pgfqpoint{1.041866in}{1.013736in}}%
\pgfpathcurveto{\pgfqpoint{1.036822in}{1.018780in}}{\pgfqpoint{1.029980in}{1.021614in}}{\pgfqpoint{1.022847in}{1.021614in}}%
\pgfpathcurveto{\pgfqpoint{1.015715in}{1.021614in}}{\pgfqpoint{1.008873in}{1.018780in}}{\pgfqpoint{1.003829in}{1.013736in}}%
\pgfpathcurveto{\pgfqpoint{0.998786in}{1.008692in}}{\pgfqpoint{0.995952in}{1.001851in}}{\pgfqpoint{0.995952in}{0.994718in}}%
\pgfpathcurveto{\pgfqpoint{0.995952in}{0.987585in}}{\pgfqpoint{0.998786in}{0.980743in}}{\pgfqpoint{1.003829in}{0.975700in}}%
\pgfpathcurveto{\pgfqpoint{1.008873in}{0.970656in}}{\pgfqpoint{1.015715in}{0.967822in}}{\pgfqpoint{1.022847in}{0.967822in}}%
\pgfpathclose%
\pgfusepath{stroke,fill}%
\end{pgfscope}%
\begin{pgfscope}%
\pgfpathrectangle{\pgfqpoint{0.533841in}{0.439824in}}{\pgfqpoint{1.121661in}{1.121661in}}%
\pgfusepath{clip}%
\pgfsetbuttcap%
\pgfsetroundjoin%
\definecolor{currentfill}{rgb}{0.762941,0.762941,0.762941}%
\pgfsetfillcolor{currentfill}%
\pgfsetlinewidth{1.003750pt}%
\definecolor{currentstroke}{rgb}{0.762941,0.762941,0.762941}%
\pgfsetstrokecolor{currentstroke}%
\pgfsetdash{}{0pt}%
\pgfpathmoveto{\pgfqpoint{1.140620in}{1.072909in}}%
\pgfpathcurveto{\pgfqpoint{1.147753in}{1.072909in}}{\pgfqpoint{1.154595in}{1.075743in}}{\pgfqpoint{1.159638in}{1.080787in}}%
\pgfpathcurveto{\pgfqpoint{1.164682in}{1.085830in}}{\pgfqpoint{1.167516in}{1.092672in}}{\pgfqpoint{1.167516in}{1.099805in}}%
\pgfpathcurveto{\pgfqpoint{1.167516in}{1.106938in}}{\pgfqpoint{1.164682in}{1.113779in}}{\pgfqpoint{1.159638in}{1.118823in}}%
\pgfpathcurveto{\pgfqpoint{1.154595in}{1.123866in}}{\pgfqpoint{1.147753in}{1.126700in}}{\pgfqpoint{1.140620in}{1.126700in}}%
\pgfpathcurveto{\pgfqpoint{1.133487in}{1.126700in}}{\pgfqpoint{1.126646in}{1.123866in}}{\pgfqpoint{1.121602in}{1.118823in}}%
\pgfpathcurveto{\pgfqpoint{1.116558in}{1.113779in}}{\pgfqpoint{1.113724in}{1.106938in}}{\pgfqpoint{1.113724in}{1.099805in}}%
\pgfpathcurveto{\pgfqpoint{1.113724in}{1.092672in}}{\pgfqpoint{1.116558in}{1.085830in}}{\pgfqpoint{1.121602in}{1.080787in}}%
\pgfpathcurveto{\pgfqpoint{1.126646in}{1.075743in}}{\pgfqpoint{1.133487in}{1.072909in}}{\pgfqpoint{1.140620in}{1.072909in}}%
\pgfpathclose%
\pgfusepath{stroke,fill}%
\end{pgfscope}%
\begin{pgfscope}%
\pgfpathrectangle{\pgfqpoint{0.533841in}{0.439824in}}{\pgfqpoint{1.121661in}{1.121661in}}%
\pgfusepath{clip}%
\pgfsetbuttcap%
\pgfsetroundjoin%
\definecolor{currentfill}{rgb}{0.713529,0.713529,0.713529}%
\pgfsetfillcolor{currentfill}%
\pgfsetlinewidth{1.003750pt}%
\definecolor{currentstroke}{rgb}{0.713529,0.713529,0.713529}%
\pgfsetstrokecolor{currentstroke}%
\pgfsetdash{}{0pt}%
\pgfpathmoveto{\pgfqpoint{1.375108in}{1.480920in}}%
\pgfpathcurveto{\pgfqpoint{1.382241in}{1.480920in}}{\pgfqpoint{1.389083in}{1.483754in}}{\pgfqpoint{1.394127in}{1.488798in}}%
\pgfpathcurveto{\pgfqpoint{1.399170in}{1.493842in}}{\pgfqpoint{1.402004in}{1.500683in}}{\pgfqpoint{1.402004in}{1.507816in}}%
\pgfpathcurveto{\pgfqpoint{1.402004in}{1.514949in}}{\pgfqpoint{1.399170in}{1.521791in}}{\pgfqpoint{1.394127in}{1.526834in}}%
\pgfpathcurveto{\pgfqpoint{1.389083in}{1.531878in}}{\pgfqpoint{1.382241in}{1.534712in}}{\pgfqpoint{1.375108in}{1.534712in}}%
\pgfpathcurveto{\pgfqpoint{1.367976in}{1.534712in}}{\pgfqpoint{1.361134in}{1.531878in}}{\pgfqpoint{1.356090in}{1.526834in}}%
\pgfpathcurveto{\pgfqpoint{1.351047in}{1.521791in}}{\pgfqpoint{1.348213in}{1.514949in}}{\pgfqpoint{1.348213in}{1.507816in}}%
\pgfpathcurveto{\pgfqpoint{1.348213in}{1.500683in}}{\pgfqpoint{1.351047in}{1.493842in}}{\pgfqpoint{1.356090in}{1.488798in}}%
\pgfpathcurveto{\pgfqpoint{1.361134in}{1.483754in}}{\pgfqpoint{1.367976in}{1.480920in}}{\pgfqpoint{1.375108in}{1.480920in}}%
\pgfpathclose%
\pgfusepath{stroke,fill}%
\end{pgfscope}%
\begin{pgfscope}%
\pgfpathrectangle{\pgfqpoint{0.533841in}{0.439824in}}{\pgfqpoint{1.121661in}{1.121661in}}%
\pgfusepath{clip}%
\pgfsetbuttcap%
\pgfsetroundjoin%
\definecolor{currentfill}{rgb}{0.667647,0.667647,0.667647}%
\pgfsetfillcolor{currentfill}%
\pgfsetlinewidth{1.003750pt}%
\definecolor{currentstroke}{rgb}{0.667647,0.667647,0.667647}%
\pgfsetstrokecolor{currentstroke}%
\pgfsetdash{}{0pt}%
\pgfpathmoveto{\pgfqpoint{1.166055in}{1.460878in}}%
\pgfpathcurveto{\pgfqpoint{1.173188in}{1.460878in}}{\pgfqpoint{1.180030in}{1.463712in}}{\pgfqpoint{1.185073in}{1.468756in}}%
\pgfpathcurveto{\pgfqpoint{1.190117in}{1.473799in}}{\pgfqpoint{1.192951in}{1.480641in}}{\pgfqpoint{1.192951in}{1.487774in}}%
\pgfpathcurveto{\pgfqpoint{1.192951in}{1.494907in}}{\pgfqpoint{1.190117in}{1.501748in}}{\pgfqpoint{1.185073in}{1.506792in}}%
\pgfpathcurveto{\pgfqpoint{1.180030in}{1.511836in}}{\pgfqpoint{1.173188in}{1.514670in}}{\pgfqpoint{1.166055in}{1.514670in}}%
\pgfpathcurveto{\pgfqpoint{1.158922in}{1.514670in}}{\pgfqpoint{1.152081in}{1.511836in}}{\pgfqpoint{1.147037in}{1.506792in}}%
\pgfpathcurveto{\pgfqpoint{1.141993in}{1.501748in}}{\pgfqpoint{1.139159in}{1.494907in}}{\pgfqpoint{1.139159in}{1.487774in}}%
\pgfpathcurveto{\pgfqpoint{1.139159in}{1.480641in}}{\pgfqpoint{1.141993in}{1.473799in}}{\pgfqpoint{1.147037in}{1.468756in}}%
\pgfpathcurveto{\pgfqpoint{1.152081in}{1.463712in}}{\pgfqpoint{1.158922in}{1.460878in}}{\pgfqpoint{1.166055in}{1.460878in}}%
\pgfpathclose%
\pgfusepath{stroke,fill}%
\end{pgfscope}%
\begin{pgfscope}%
\pgfpathrectangle{\pgfqpoint{0.533841in}{0.439824in}}{\pgfqpoint{1.121661in}{1.121661in}}%
\pgfusepath{clip}%
\pgfsetbuttcap%
\pgfsetroundjoin%
\definecolor{currentfill}{rgb}{0.618235,0.618235,0.618235}%
\pgfsetfillcolor{currentfill}%
\pgfsetlinewidth{1.003750pt}%
\definecolor{currentstroke}{rgb}{0.618235,0.618235,0.618235}%
\pgfsetstrokecolor{currentstroke}%
\pgfsetdash{}{0pt}%
\pgfpathmoveto{\pgfqpoint{1.384576in}{1.293828in}}%
\pgfpathcurveto{\pgfqpoint{1.391709in}{1.293828in}}{\pgfqpoint{1.398551in}{1.296662in}}{\pgfqpoint{1.403594in}{1.301705in}}%
\pgfpathcurveto{\pgfqpoint{1.408638in}{1.306749in}}{\pgfqpoint{1.411472in}{1.313591in}}{\pgfqpoint{1.411472in}{1.320723in}}%
\pgfpathcurveto{\pgfqpoint{1.411472in}{1.327856in}}{\pgfqpoint{1.408638in}{1.334698in}}{\pgfqpoint{1.403594in}{1.339742in}}%
\pgfpathcurveto{\pgfqpoint{1.398551in}{1.344785in}}{\pgfqpoint{1.391709in}{1.347619in}}{\pgfqpoint{1.384576in}{1.347619in}}%
\pgfpathcurveto{\pgfqpoint{1.377443in}{1.347619in}}{\pgfqpoint{1.370602in}{1.344785in}}{\pgfqpoint{1.365558in}{1.339742in}}%
\pgfpathcurveto{\pgfqpoint{1.360514in}{1.334698in}}{\pgfqpoint{1.357681in}{1.327856in}}{\pgfqpoint{1.357681in}{1.320723in}}%
\pgfpathcurveto{\pgfqpoint{1.357681in}{1.313591in}}{\pgfqpoint{1.360514in}{1.306749in}}{\pgfqpoint{1.365558in}{1.301705in}}%
\pgfpathcurveto{\pgfqpoint{1.370602in}{1.296662in}}{\pgfqpoint{1.377443in}{1.293828in}}{\pgfqpoint{1.384576in}{1.293828in}}%
\pgfpathclose%
\pgfusepath{stroke,fill}%
\end{pgfscope}%
\begin{pgfscope}%
\pgfpathrectangle{\pgfqpoint{0.533841in}{0.439824in}}{\pgfqpoint{1.121661in}{1.121661in}}%
\pgfusepath{clip}%
\pgfsetbuttcap%
\pgfsetroundjoin%
\definecolor{currentfill}{rgb}{0.572353,0.572353,0.572353}%
\pgfsetfillcolor{currentfill}%
\pgfsetlinewidth{1.003750pt}%
\definecolor{currentstroke}{rgb}{0.572353,0.572353,0.572353}%
\pgfsetstrokecolor{currentstroke}%
\pgfsetdash{}{0pt}%
\pgfpathmoveto{\pgfqpoint{1.353022in}{1.441099in}}%
\pgfpathcurveto{\pgfqpoint{1.360155in}{1.441099in}}{\pgfqpoint{1.366997in}{1.443933in}}{\pgfqpoint{1.372041in}{1.448977in}}%
\pgfpathcurveto{\pgfqpoint{1.377084in}{1.454020in}}{\pgfqpoint{1.379918in}{1.460862in}}{\pgfqpoint{1.379918in}{1.467995in}}%
\pgfpathcurveto{\pgfqpoint{1.379918in}{1.475128in}}{\pgfqpoint{1.377084in}{1.481969in}}{\pgfqpoint{1.372041in}{1.487013in}}%
\pgfpathcurveto{\pgfqpoint{1.366997in}{1.492057in}}{\pgfqpoint{1.360155in}{1.494891in}}{\pgfqpoint{1.353022in}{1.494891in}}%
\pgfpathcurveto{\pgfqpoint{1.345890in}{1.494891in}}{\pgfqpoint{1.339048in}{1.492057in}}{\pgfqpoint{1.334004in}{1.487013in}}%
\pgfpathcurveto{\pgfqpoint{1.328961in}{1.481969in}}{\pgfqpoint{1.326127in}{1.475128in}}{\pgfqpoint{1.326127in}{1.467995in}}%
\pgfpathcurveto{\pgfqpoint{1.326127in}{1.460862in}}{\pgfqpoint{1.328961in}{1.454020in}}{\pgfqpoint{1.334004in}{1.448977in}}%
\pgfpathcurveto{\pgfqpoint{1.339048in}{1.443933in}}{\pgfqpoint{1.345890in}{1.441099in}}{\pgfqpoint{1.353022in}{1.441099in}}%
\pgfpathclose%
\pgfusepath{stroke,fill}%
\end{pgfscope}%
\begin{pgfscope}%
\pgfpathrectangle{\pgfqpoint{0.533841in}{0.439824in}}{\pgfqpoint{1.121661in}{1.121661in}}%
\pgfusepath{clip}%
\pgfsetbuttcap%
\pgfsetroundjoin%
\definecolor{currentfill}{rgb}{0.522941,0.522941,0.522941}%
\pgfsetfillcolor{currentfill}%
\pgfsetlinewidth{1.003750pt}%
\definecolor{currentstroke}{rgb}{0.522941,0.522941,0.522941}%
\pgfsetstrokecolor{currentstroke}%
\pgfsetdash{}{0pt}%
\pgfpathmoveto{\pgfqpoint{1.331989in}{1.290084in}}%
\pgfpathcurveto{\pgfqpoint{1.339122in}{1.290084in}}{\pgfqpoint{1.345963in}{1.292918in}}{\pgfqpoint{1.351007in}{1.297962in}}%
\pgfpathcurveto{\pgfqpoint{1.356051in}{1.303005in}}{\pgfqpoint{1.358885in}{1.309847in}}{\pgfqpoint{1.358885in}{1.316980in}}%
\pgfpathcurveto{\pgfqpoint{1.358885in}{1.324113in}}{\pgfqpoint{1.356051in}{1.330954in}}{\pgfqpoint{1.351007in}{1.335998in}}%
\pgfpathcurveto{\pgfqpoint{1.345963in}{1.341042in}}{\pgfqpoint{1.339122in}{1.343876in}}{\pgfqpoint{1.331989in}{1.343876in}}%
\pgfpathcurveto{\pgfqpoint{1.324856in}{1.343876in}}{\pgfqpoint{1.318014in}{1.341042in}}{\pgfqpoint{1.312971in}{1.335998in}}%
\pgfpathcurveto{\pgfqpoint{1.307927in}{1.330954in}}{\pgfqpoint{1.305093in}{1.324113in}}{\pgfqpoint{1.305093in}{1.316980in}}%
\pgfpathcurveto{\pgfqpoint{1.305093in}{1.309847in}}{\pgfqpoint{1.307927in}{1.303005in}}{\pgfqpoint{1.312971in}{1.297962in}}%
\pgfpathcurveto{\pgfqpoint{1.318014in}{1.292918in}}{\pgfqpoint{1.324856in}{1.290084in}}{\pgfqpoint{1.331989in}{1.290084in}}%
\pgfpathclose%
\pgfusepath{stroke,fill}%
\end{pgfscope}%
\begin{pgfscope}%
\pgfpathrectangle{\pgfqpoint{0.533841in}{0.439824in}}{\pgfqpoint{1.121661in}{1.121661in}}%
\pgfusepath{clip}%
\pgfsetbuttcap%
\pgfsetroundjoin%
\definecolor{currentfill}{rgb}{0.477059,0.477059,0.477059}%
\pgfsetfillcolor{currentfill}%
\pgfsetlinewidth{1.003750pt}%
\definecolor{currentstroke}{rgb}{0.477059,0.477059,0.477059}%
\pgfsetstrokecolor{currentstroke}%
\pgfsetdash{}{0pt}%
\pgfpathmoveto{\pgfqpoint{1.385635in}{1.341838in}}%
\pgfpathcurveto{\pgfqpoint{1.392768in}{1.341838in}}{\pgfqpoint{1.399609in}{1.344672in}}{\pgfqpoint{1.404653in}{1.349715in}}%
\pgfpathcurveto{\pgfqpoint{1.409697in}{1.354759in}}{\pgfqpoint{1.412531in}{1.361601in}}{\pgfqpoint{1.412531in}{1.368734in}}%
\pgfpathcurveto{\pgfqpoint{1.412531in}{1.375866in}}{\pgfqpoint{1.409697in}{1.382708in}}{\pgfqpoint{1.404653in}{1.387752in}}%
\pgfpathcurveto{\pgfqpoint{1.399609in}{1.392795in}}{\pgfqpoint{1.392768in}{1.395629in}}{\pgfqpoint{1.385635in}{1.395629in}}%
\pgfpathcurveto{\pgfqpoint{1.378502in}{1.395629in}}{\pgfqpoint{1.371661in}{1.392795in}}{\pgfqpoint{1.366617in}{1.387752in}}%
\pgfpathcurveto{\pgfqpoint{1.361573in}{1.382708in}}{\pgfqpoint{1.358739in}{1.375866in}}{\pgfqpoint{1.358739in}{1.368734in}}%
\pgfpathcurveto{\pgfqpoint{1.358739in}{1.361601in}}{\pgfqpoint{1.361573in}{1.354759in}}{\pgfqpoint{1.366617in}{1.349715in}}%
\pgfpathcurveto{\pgfqpoint{1.371661in}{1.344672in}}{\pgfqpoint{1.378502in}{1.341838in}}{\pgfqpoint{1.385635in}{1.341838in}}%
\pgfpathclose%
\pgfusepath{stroke,fill}%
\end{pgfscope}%
\begin{pgfscope}%
\pgfpathrectangle{\pgfqpoint{0.533841in}{0.439824in}}{\pgfqpoint{1.121661in}{1.121661in}}%
\pgfusepath{clip}%
\pgfsetbuttcap%
\pgfsetroundjoin%
\definecolor{currentfill}{rgb}{0.427647,0.427647,0.427647}%
\pgfsetfillcolor{currentfill}%
\pgfsetlinewidth{1.003750pt}%
\definecolor{currentstroke}{rgb}{0.427647,0.427647,0.427647}%
\pgfsetstrokecolor{currentstroke}%
\pgfsetdash{}{0pt}%
\pgfpathmoveto{\pgfqpoint{1.362098in}{1.361664in}}%
\pgfpathcurveto{\pgfqpoint{1.369230in}{1.361664in}}{\pgfqpoint{1.376072in}{1.364498in}}{\pgfqpoint{1.381116in}{1.369542in}}%
\pgfpathcurveto{\pgfqpoint{1.386159in}{1.374585in}}{\pgfqpoint{1.388993in}{1.381427in}}{\pgfqpoint{1.388993in}{1.388560in}}%
\pgfpathcurveto{\pgfqpoint{1.388993in}{1.395693in}}{\pgfqpoint{1.386159in}{1.402534in}}{\pgfqpoint{1.381116in}{1.407578in}}%
\pgfpathcurveto{\pgfqpoint{1.376072in}{1.412622in}}{\pgfqpoint{1.369230in}{1.415456in}}{\pgfqpoint{1.362098in}{1.415456in}}%
\pgfpathcurveto{\pgfqpoint{1.354965in}{1.415456in}}{\pgfqpoint{1.348123in}{1.412622in}}{\pgfqpoint{1.343079in}{1.407578in}}%
\pgfpathcurveto{\pgfqpoint{1.338036in}{1.402534in}}{\pgfqpoint{1.335202in}{1.395693in}}{\pgfqpoint{1.335202in}{1.388560in}}%
\pgfpathcurveto{\pgfqpoint{1.335202in}{1.381427in}}{\pgfqpoint{1.338036in}{1.374585in}}{\pgfqpoint{1.343079in}{1.369542in}}%
\pgfpathcurveto{\pgfqpoint{1.348123in}{1.364498in}}{\pgfqpoint{1.354965in}{1.361664in}}{\pgfqpoint{1.362098in}{1.361664in}}%
\pgfpathclose%
\pgfusepath{stroke,fill}%
\end{pgfscope}%
\begin{pgfscope}%
\pgfpathrectangle{\pgfqpoint{0.533841in}{0.439824in}}{\pgfqpoint{1.121661in}{1.121661in}}%
\pgfusepath{clip}%
\pgfsetbuttcap%
\pgfsetroundjoin%
\definecolor{currentfill}{rgb}{0.381765,0.381765,0.381765}%
\pgfsetfillcolor{currentfill}%
\pgfsetlinewidth{1.003750pt}%
\definecolor{currentstroke}{rgb}{0.381765,0.381765,0.381765}%
\pgfsetstrokecolor{currentstroke}%
\pgfsetdash{}{0pt}%
\pgfpathmoveto{\pgfqpoint{1.354785in}{1.363419in}}%
\pgfpathcurveto{\pgfqpoint{1.361918in}{1.363419in}}{\pgfqpoint{1.368760in}{1.366253in}}{\pgfqpoint{1.373803in}{1.371296in}}%
\pgfpathcurveto{\pgfqpoint{1.378847in}{1.376340in}}{\pgfqpoint{1.381681in}{1.383181in}}{\pgfqpoint{1.381681in}{1.390314in}}%
\pgfpathcurveto{\pgfqpoint{1.381681in}{1.397447in}}{\pgfqpoint{1.378847in}{1.404289in}}{\pgfqpoint{1.373803in}{1.409332in}}%
\pgfpathcurveto{\pgfqpoint{1.368760in}{1.414376in}}{\pgfqpoint{1.361918in}{1.417210in}}{\pgfqpoint{1.354785in}{1.417210in}}%
\pgfpathcurveto{\pgfqpoint{1.347652in}{1.417210in}}{\pgfqpoint{1.340811in}{1.414376in}}{\pgfqpoint{1.335767in}{1.409332in}}%
\pgfpathcurveto{\pgfqpoint{1.330723in}{1.404289in}}{\pgfqpoint{1.327889in}{1.397447in}}{\pgfqpoint{1.327889in}{1.390314in}}%
\pgfpathcurveto{\pgfqpoint{1.327889in}{1.383181in}}{\pgfqpoint{1.330723in}{1.376340in}}{\pgfqpoint{1.335767in}{1.371296in}}%
\pgfpathcurveto{\pgfqpoint{1.340811in}{1.366253in}}{\pgfqpoint{1.347652in}{1.363419in}}{\pgfqpoint{1.354785in}{1.363419in}}%
\pgfpathclose%
\pgfusepath{stroke,fill}%
\end{pgfscope}%
\begin{pgfscope}%
\pgfpathrectangle{\pgfqpoint{0.533841in}{0.439824in}}{\pgfqpoint{1.121661in}{1.121661in}}%
\pgfusepath{clip}%
\pgfsetbuttcap%
\pgfsetroundjoin%
\definecolor{currentfill}{rgb}{0.332353,0.332353,0.332353}%
\pgfsetfillcolor{currentfill}%
\pgfsetlinewidth{1.003750pt}%
\definecolor{currentstroke}{rgb}{0.332353,0.332353,0.332353}%
\pgfsetstrokecolor{currentstroke}%
\pgfsetdash{}{0pt}%
\pgfpathmoveto{\pgfqpoint{1.336500in}{1.265410in}}%
\pgfpathcurveto{\pgfqpoint{1.343633in}{1.265410in}}{\pgfqpoint{1.350475in}{1.268244in}}{\pgfqpoint{1.355518in}{1.273288in}}%
\pgfpathcurveto{\pgfqpoint{1.360562in}{1.278332in}}{\pgfqpoint{1.363396in}{1.285173in}}{\pgfqpoint{1.363396in}{1.292306in}}%
\pgfpathcurveto{\pgfqpoint{1.363396in}{1.299439in}}{\pgfqpoint{1.360562in}{1.306281in}}{\pgfqpoint{1.355518in}{1.311324in}}%
\pgfpathcurveto{\pgfqpoint{1.350475in}{1.316368in}}{\pgfqpoint{1.343633in}{1.319202in}}{\pgfqpoint{1.336500in}{1.319202in}}%
\pgfpathcurveto{\pgfqpoint{1.329367in}{1.319202in}}{\pgfqpoint{1.322526in}{1.316368in}}{\pgfqpoint{1.317482in}{1.311324in}}%
\pgfpathcurveto{\pgfqpoint{1.312438in}{1.306281in}}{\pgfqpoint{1.309604in}{1.299439in}}{\pgfqpoint{1.309604in}{1.292306in}}%
\pgfpathcurveto{\pgfqpoint{1.309604in}{1.285173in}}{\pgfqpoint{1.312438in}{1.278332in}}{\pgfqpoint{1.317482in}{1.273288in}}%
\pgfpathcurveto{\pgfqpoint{1.322526in}{1.268244in}}{\pgfqpoint{1.329367in}{1.265410in}}{\pgfqpoint{1.336500in}{1.265410in}}%
\pgfpathclose%
\pgfusepath{stroke,fill}%
\end{pgfscope}%
\begin{pgfscope}%
\pgfpathrectangle{\pgfqpoint{0.533841in}{0.439824in}}{\pgfqpoint{1.121661in}{1.121661in}}%
\pgfusepath{clip}%
\pgfsetbuttcap%
\pgfsetroundjoin%
\definecolor{currentfill}{rgb}{0.286471,0.286471,0.286471}%
\pgfsetfillcolor{currentfill}%
\pgfsetlinewidth{1.003750pt}%
\definecolor{currentstroke}{rgb}{0.286471,0.286471,0.286471}%
\pgfsetstrokecolor{currentstroke}%
\pgfsetdash{}{0pt}%
\pgfpathmoveto{\pgfqpoint{1.418524in}{1.389484in}}%
\pgfpathcurveto{\pgfqpoint{1.425657in}{1.389484in}}{\pgfqpoint{1.432498in}{1.392318in}}{\pgfqpoint{1.437542in}{1.397361in}}%
\pgfpathcurveto{\pgfqpoint{1.442586in}{1.402405in}}{\pgfqpoint{1.445420in}{1.409247in}}{\pgfqpoint{1.445420in}{1.416379in}}%
\pgfpathcurveto{\pgfqpoint{1.445420in}{1.423512in}}{\pgfqpoint{1.442586in}{1.430354in}}{\pgfqpoint{1.437542in}{1.435398in}}%
\pgfpathcurveto{\pgfqpoint{1.432498in}{1.440441in}}{\pgfqpoint{1.425657in}{1.443275in}}{\pgfqpoint{1.418524in}{1.443275in}}%
\pgfpathcurveto{\pgfqpoint{1.411391in}{1.443275in}}{\pgfqpoint{1.404549in}{1.440441in}}{\pgfqpoint{1.399506in}{1.435398in}}%
\pgfpathcurveto{\pgfqpoint{1.394462in}{1.430354in}}{\pgfqpoint{1.391628in}{1.423512in}}{\pgfqpoint{1.391628in}{1.416379in}}%
\pgfpathcurveto{\pgfqpoint{1.391628in}{1.409247in}}{\pgfqpoint{1.394462in}{1.402405in}}{\pgfqpoint{1.399506in}{1.397361in}}%
\pgfpathcurveto{\pgfqpoint{1.404549in}{1.392318in}}{\pgfqpoint{1.411391in}{1.389484in}}{\pgfqpoint{1.418524in}{1.389484in}}%
\pgfpathclose%
\pgfusepath{stroke,fill}%
\end{pgfscope}%
\begin{pgfscope}%
\pgfpathrectangle{\pgfqpoint{0.533841in}{0.439824in}}{\pgfqpoint{1.121661in}{1.121661in}}%
\pgfusepath{clip}%
\pgfsetbuttcap%
\pgfsetroundjoin%
\definecolor{currentfill}{rgb}{0.237059,0.237059,0.237059}%
\pgfsetfillcolor{currentfill}%
\pgfsetlinewidth{1.003750pt}%
\definecolor{currentstroke}{rgb}{0.237059,0.237059,0.237059}%
\pgfsetstrokecolor{currentstroke}%
\pgfsetdash{}{0pt}%
\pgfpathmoveto{\pgfqpoint{1.364340in}{1.420282in}}%
\pgfpathcurveto{\pgfqpoint{1.371473in}{1.420282in}}{\pgfqpoint{1.378314in}{1.423115in}}{\pgfqpoint{1.383358in}{1.428159in}}%
\pgfpathcurveto{\pgfqpoint{1.388402in}{1.433203in}}{\pgfqpoint{1.391235in}{1.440044in}}{\pgfqpoint{1.391235in}{1.447177in}}%
\pgfpathcurveto{\pgfqpoint{1.391235in}{1.454310in}}{\pgfqpoint{1.388402in}{1.461152in}}{\pgfqpoint{1.383358in}{1.466195in}}%
\pgfpathcurveto{\pgfqpoint{1.378314in}{1.471239in}}{\pgfqpoint{1.371473in}{1.474073in}}{\pgfqpoint{1.364340in}{1.474073in}}%
\pgfpathcurveto{\pgfqpoint{1.357207in}{1.474073in}}{\pgfqpoint{1.350365in}{1.471239in}}{\pgfqpoint{1.345322in}{1.466195in}}%
\pgfpathcurveto{\pgfqpoint{1.340278in}{1.461152in}}{\pgfqpoint{1.337444in}{1.454310in}}{\pgfqpoint{1.337444in}{1.447177in}}%
\pgfpathcurveto{\pgfqpoint{1.337444in}{1.440044in}}{\pgfqpoint{1.340278in}{1.433203in}}{\pgfqpoint{1.345322in}{1.428159in}}%
\pgfpathcurveto{\pgfqpoint{1.350365in}{1.423115in}}{\pgfqpoint{1.357207in}{1.420282in}}{\pgfqpoint{1.364340in}{1.420282in}}%
\pgfpathclose%
\pgfusepath{stroke,fill}%
\end{pgfscope}%
\begin{pgfscope}%
\pgfpathrectangle{\pgfqpoint{0.533841in}{0.439824in}}{\pgfqpoint{1.121661in}{1.121661in}}%
\pgfusepath{clip}%
\pgfsetbuttcap%
\pgfsetroundjoin%
\definecolor{currentfill}{rgb}{0.191176,0.191176,0.191176}%
\pgfsetfillcolor{currentfill}%
\pgfsetlinewidth{1.003750pt}%
\definecolor{currentstroke}{rgb}{0.191176,0.191176,0.191176}%
\pgfsetstrokecolor{currentstroke}%
\pgfsetdash{}{0pt}%
\pgfpathmoveto{\pgfqpoint{1.393312in}{1.283009in}}%
\pgfpathcurveto{\pgfqpoint{1.400445in}{1.283009in}}{\pgfqpoint{1.407287in}{1.285843in}}{\pgfqpoint{1.412330in}{1.290887in}}%
\pgfpathcurveto{\pgfqpoint{1.417374in}{1.295930in}}{\pgfqpoint{1.420208in}{1.302772in}}{\pgfqpoint{1.420208in}{1.309905in}}%
\pgfpathcurveto{\pgfqpoint{1.420208in}{1.317038in}}{\pgfqpoint{1.417374in}{1.323879in}}{\pgfqpoint{1.412330in}{1.328923in}}%
\pgfpathcurveto{\pgfqpoint{1.407287in}{1.333967in}}{\pgfqpoint{1.400445in}{1.336801in}}{\pgfqpoint{1.393312in}{1.336801in}}%
\pgfpathcurveto{\pgfqpoint{1.386179in}{1.336801in}}{\pgfqpoint{1.379338in}{1.333967in}}{\pgfqpoint{1.374294in}{1.328923in}}%
\pgfpathcurveto{\pgfqpoint{1.369250in}{1.323879in}}{\pgfqpoint{1.366417in}{1.317038in}}{\pgfqpoint{1.366417in}{1.309905in}}%
\pgfpathcurveto{\pgfqpoint{1.366417in}{1.302772in}}{\pgfqpoint{1.369250in}{1.295930in}}{\pgfqpoint{1.374294in}{1.290887in}}%
\pgfpathcurveto{\pgfqpoint{1.379338in}{1.285843in}}{\pgfqpoint{1.386179in}{1.283009in}}{\pgfqpoint{1.393312in}{1.283009in}}%
\pgfpathclose%
\pgfusepath{stroke,fill}%
\end{pgfscope}%
\begin{pgfscope}%
\pgfpathrectangle{\pgfqpoint{0.533841in}{0.439824in}}{\pgfqpoint{1.121661in}{1.121661in}}%
\pgfusepath{clip}%
\pgfsetbuttcap%
\pgfsetroundjoin%
\definecolor{currentfill}{rgb}{0.141765,0.141765,0.141765}%
\pgfsetfillcolor{currentfill}%
\pgfsetlinewidth{1.003750pt}%
\definecolor{currentstroke}{rgb}{0.141765,0.141765,0.141765}%
\pgfsetstrokecolor{currentstroke}%
\pgfsetdash{}{0pt}%
\pgfpathmoveto{\pgfqpoint{1.393917in}{1.416676in}}%
\pgfpathcurveto{\pgfqpoint{1.401050in}{1.416676in}}{\pgfqpoint{1.407892in}{1.419510in}}{\pgfqpoint{1.412936in}{1.424553in}}%
\pgfpathcurveto{\pgfqpoint{1.417979in}{1.429597in}}{\pgfqpoint{1.420813in}{1.436439in}}{\pgfqpoint{1.420813in}{1.443571in}}%
\pgfpathcurveto{\pgfqpoint{1.420813in}{1.450704in}}{\pgfqpoint{1.417979in}{1.457546in}}{\pgfqpoint{1.412936in}{1.462589in}}%
\pgfpathcurveto{\pgfqpoint{1.407892in}{1.467633in}}{\pgfqpoint{1.401050in}{1.470467in}}{\pgfqpoint{1.393917in}{1.470467in}}%
\pgfpathcurveto{\pgfqpoint{1.386785in}{1.470467in}}{\pgfqpoint{1.379943in}{1.467633in}}{\pgfqpoint{1.374899in}{1.462589in}}%
\pgfpathcurveto{\pgfqpoint{1.369856in}{1.457546in}}{\pgfqpoint{1.367022in}{1.450704in}}{\pgfqpoint{1.367022in}{1.443571in}}%
\pgfpathcurveto{\pgfqpoint{1.367022in}{1.436439in}}{\pgfqpoint{1.369856in}{1.429597in}}{\pgfqpoint{1.374899in}{1.424553in}}%
\pgfpathcurveto{\pgfqpoint{1.379943in}{1.419510in}}{\pgfqpoint{1.386785in}{1.416676in}}{\pgfqpoint{1.393917in}{1.416676in}}%
\pgfpathclose%
\pgfusepath{stroke,fill}%
\end{pgfscope}%
\begin{pgfscope}%
\pgfpathrectangle{\pgfqpoint{0.533841in}{0.439824in}}{\pgfqpoint{1.121661in}{1.121661in}}%
\pgfusepath{clip}%
\pgfsetbuttcap%
\pgfsetroundjoin%
\definecolor{currentfill}{rgb}{0.095882,0.095882,0.095882}%
\pgfsetfillcolor{currentfill}%
\pgfsetlinewidth{1.003750pt}%
\definecolor{currentstroke}{rgb}{0.095882,0.095882,0.095882}%
\pgfsetstrokecolor{currentstroke}%
\pgfsetdash{}{0pt}%
\pgfpathmoveto{\pgfqpoint{1.479725in}{1.303254in}}%
\pgfpathcurveto{\pgfqpoint{1.486858in}{1.303254in}}{\pgfqpoint{1.493699in}{1.306088in}}{\pgfqpoint{1.498743in}{1.311132in}}%
\pgfpathcurveto{\pgfqpoint{1.503787in}{1.316175in}}{\pgfqpoint{1.506620in}{1.323017in}}{\pgfqpoint{1.506620in}{1.330150in}}%
\pgfpathcurveto{\pgfqpoint{1.506620in}{1.337283in}}{\pgfqpoint{1.503787in}{1.344124in}}{\pgfqpoint{1.498743in}{1.349168in}}%
\pgfpathcurveto{\pgfqpoint{1.493699in}{1.354212in}}{\pgfqpoint{1.486858in}{1.357046in}}{\pgfqpoint{1.479725in}{1.357046in}}%
\pgfpathcurveto{\pgfqpoint{1.472592in}{1.357046in}}{\pgfqpoint{1.465750in}{1.354212in}}{\pgfqpoint{1.460707in}{1.349168in}}%
\pgfpathcurveto{\pgfqpoint{1.455663in}{1.344124in}}{\pgfqpoint{1.452829in}{1.337283in}}{\pgfqpoint{1.452829in}{1.330150in}}%
\pgfpathcurveto{\pgfqpoint{1.452829in}{1.323017in}}{\pgfqpoint{1.455663in}{1.316175in}}{\pgfqpoint{1.460707in}{1.311132in}}%
\pgfpathcurveto{\pgfqpoint{1.465750in}{1.306088in}}{\pgfqpoint{1.472592in}{1.303254in}}{\pgfqpoint{1.479725in}{1.303254in}}%
\pgfpathclose%
\pgfusepath{stroke,fill}%
\end{pgfscope}%
\begin{pgfscope}%
\pgfpathrectangle{\pgfqpoint{0.533841in}{0.439824in}}{\pgfqpoint{1.121661in}{1.121661in}}%
\pgfusepath{clip}%
\pgfsetbuttcap%
\pgfsetroundjoin%
\definecolor{currentfill}{rgb}{0.050000,0.050000,0.050000}%
\pgfsetfillcolor{currentfill}%
\pgfsetlinewidth{1.003750pt}%
\definecolor{currentstroke}{rgb}{0.050000,0.050000,0.050000}%
\pgfsetstrokecolor{currentstroke}%
\pgfsetdash{}{0pt}%
\pgfpathmoveto{\pgfqpoint{1.428598in}{1.458574in}}%
\pgfpathcurveto{\pgfqpoint{1.435731in}{1.458574in}}{\pgfqpoint{1.442573in}{1.461408in}}{\pgfqpoint{1.447616in}{1.466452in}}%
\pgfpathcurveto{\pgfqpoint{1.452660in}{1.471495in}}{\pgfqpoint{1.455494in}{1.478337in}}{\pgfqpoint{1.455494in}{1.485470in}}%
\pgfpathcurveto{\pgfqpoint{1.455494in}{1.492602in}}{\pgfqpoint{1.452660in}{1.499444in}}{\pgfqpoint{1.447616in}{1.504488in}}%
\pgfpathcurveto{\pgfqpoint{1.442573in}{1.509531in}}{\pgfqpoint{1.435731in}{1.512365in}}{\pgfqpoint{1.428598in}{1.512365in}}%
\pgfpathcurveto{\pgfqpoint{1.421465in}{1.512365in}}{\pgfqpoint{1.414624in}{1.509531in}}{\pgfqpoint{1.409580in}{1.504488in}}%
\pgfpathcurveto{\pgfqpoint{1.404536in}{1.499444in}}{\pgfqpoint{1.401702in}{1.492602in}}{\pgfqpoint{1.401702in}{1.485470in}}%
\pgfpathcurveto{\pgfqpoint{1.401702in}{1.478337in}}{\pgfqpoint{1.404536in}{1.471495in}}{\pgfqpoint{1.409580in}{1.466452in}}%
\pgfpathcurveto{\pgfqpoint{1.414624in}{1.461408in}}{\pgfqpoint{1.421465in}{1.458574in}}{\pgfqpoint{1.428598in}{1.458574in}}%
\pgfpathclose%
\pgfusepath{stroke,fill}%
\end{pgfscope}%
\begin{pgfscope}%
\pgfpathrectangle{\pgfqpoint{0.533841in}{0.439824in}}{\pgfqpoint{1.121661in}{1.121661in}}%
\pgfusepath{clip}%
\pgfsetbuttcap%
\pgfsetroundjoin%
\definecolor{currentfill}{rgb}{1.000000,0.549020,0.000000}%
\pgfsetfillcolor{currentfill}%
\pgfsetlinewidth{1.003750pt}%
\definecolor{currentstroke}{rgb}{1.000000,0.549020,0.000000}%
\pgfsetstrokecolor{currentstroke}%
\pgfsetdash{}{0pt}%
\pgfsys@defobject{currentmarker}{\pgfqpoint{-0.051159in}{-0.043518in}}{\pgfqpoint{0.051159in}{0.053791in}}{%
\pgfpathmoveto{\pgfqpoint{0.000000in}{0.053791in}}%
\pgfpathlineto{\pgfqpoint{-0.012077in}{0.016622in}}%
\pgfpathlineto{\pgfqpoint{-0.051159in}{0.016622in}}%
\pgfpathlineto{\pgfqpoint{-0.019541in}{-0.006349in}}%
\pgfpathlineto{\pgfqpoint{-0.031618in}{-0.043518in}}%
\pgfpathlineto{\pgfqpoint{-0.000000in}{-0.020546in}}%
\pgfpathlineto{\pgfqpoint{0.031618in}{-0.043518in}}%
\pgfpathlineto{\pgfqpoint{0.019541in}{-0.006349in}}%
\pgfpathlineto{\pgfqpoint{0.051159in}{0.016622in}}%
\pgfpathlineto{\pgfqpoint{0.012077in}{0.016622in}}%
\pgfpathclose%
\pgfusepath{stroke,fill}%
}%
\begin{pgfscope}%
\pgfsys@transformshift{1.399177in}{1.395400in}%
\pgfsys@useobject{currentmarker}{}%
\end{pgfscope}%
\end{pgfscope}%
\begin{pgfscope}%
\pgfpathrectangle{\pgfqpoint{0.533841in}{0.439824in}}{\pgfqpoint{1.121661in}{1.121661in}}%
\pgfusepath{clip}%
\pgfsetbuttcap%
\pgfsetroundjoin%
\definecolor{currentfill}{rgb}{0.698039,0.133333,0.133333}%
\pgfsetfillcolor{currentfill}%
\pgfsetlinewidth{1.003750pt}%
\definecolor{currentstroke}{rgb}{0.698039,0.133333,0.133333}%
\pgfsetstrokecolor{currentstroke}%
\pgfsetdash{}{0pt}%
\pgfsys@defobject{currentmarker}{\pgfqpoint{-0.051159in}{-0.043518in}}{\pgfqpoint{0.051159in}{0.053791in}}{%
\pgfpathmoveto{\pgfqpoint{0.000000in}{0.053791in}}%
\pgfpathlineto{\pgfqpoint{-0.012077in}{0.016622in}}%
\pgfpathlineto{\pgfqpoint{-0.051159in}{0.016622in}}%
\pgfpathlineto{\pgfqpoint{-0.019541in}{-0.006349in}}%
\pgfpathlineto{\pgfqpoint{-0.031618in}{-0.043518in}}%
\pgfpathlineto{\pgfqpoint{-0.000000in}{-0.020546in}}%
\pgfpathlineto{\pgfqpoint{0.031618in}{-0.043518in}}%
\pgfpathlineto{\pgfqpoint{0.019541in}{-0.006349in}}%
\pgfpathlineto{\pgfqpoint{0.051159in}{0.016622in}}%
\pgfpathlineto{\pgfqpoint{0.012077in}{0.016622in}}%
\pgfpathclose%
\pgfusepath{stroke,fill}%
}%
\begin{pgfscope}%
\pgfsys@transformshift{1.375087in}{1.281069in}%
\pgfsys@useobject{currentmarker}{}%
\end{pgfscope}%
\end{pgfscope}%
\begin{pgfscope}%
\pgfsetbuttcap%
\pgfsetroundjoin%
\definecolor{currentfill}{rgb}{0.000000,0.000000,0.000000}%
\pgfsetfillcolor{currentfill}%
\pgfsetlinewidth{0.803000pt}%
\definecolor{currentstroke}{rgb}{0.000000,0.000000,0.000000}%
\pgfsetstrokecolor{currentstroke}%
\pgfsetdash{}{0pt}%
\pgfsys@defobject{currentmarker}{\pgfqpoint{0.000000in}{-0.048611in}}{\pgfqpoint{0.000000in}{0.000000in}}{%
\pgfpathmoveto{\pgfqpoint{0.000000in}{0.000000in}}%
\pgfpathlineto{\pgfqpoint{0.000000in}{-0.048611in}}%
\pgfusepath{stroke,fill}%
}%
\begin{pgfscope}%
\pgfsys@transformshift{0.627313in}{0.439824in}%
\pgfsys@useobject{currentmarker}{}%
\end{pgfscope}%
\end{pgfscope}%
\begin{pgfscope}%
\definecolor{textcolor}{rgb}{0.000000,0.000000,0.000000}%
\pgfsetstrokecolor{textcolor}%
\pgfsetfillcolor{textcolor}%
\pgftext[x=0.627313in,y=0.342602in,,top]{\color{textcolor}\rmfamily\fontsize{9.000000}{10.800000}\selectfont \(\displaystyle {0.020}\)}%
\end{pgfscope}%
\begin{pgfscope}%
\pgfsetbuttcap%
\pgfsetroundjoin%
\definecolor{currentfill}{rgb}{0.000000,0.000000,0.000000}%
\pgfsetfillcolor{currentfill}%
\pgfsetlinewidth{0.803000pt}%
\definecolor{currentstroke}{rgb}{0.000000,0.000000,0.000000}%
\pgfsetstrokecolor{currentstroke}%
\pgfsetdash{}{0pt}%
\pgfsys@defobject{currentmarker}{\pgfqpoint{0.000000in}{-0.048611in}}{\pgfqpoint{0.000000in}{0.000000in}}{%
\pgfpathmoveto{\pgfqpoint{0.000000in}{0.000000in}}%
\pgfpathlineto{\pgfqpoint{0.000000in}{-0.048611in}}%
\pgfusepath{stroke,fill}%
}%
\begin{pgfscope}%
\pgfsys@transformshift{1.094672in}{0.439824in}%
\pgfsys@useobject{currentmarker}{}%
\end{pgfscope}%
\end{pgfscope}%
\begin{pgfscope}%
\definecolor{textcolor}{rgb}{0.000000,0.000000,0.000000}%
\pgfsetstrokecolor{textcolor}%
\pgfsetfillcolor{textcolor}%
\pgftext[x=1.094672in,y=0.342602in,,top]{\color{textcolor}\rmfamily\fontsize{9.000000}{10.800000}\selectfont \(\displaystyle {0.024}\)}%
\end{pgfscope}%
\begin{pgfscope}%
\pgfsetbuttcap%
\pgfsetroundjoin%
\definecolor{currentfill}{rgb}{0.000000,0.000000,0.000000}%
\pgfsetfillcolor{currentfill}%
\pgfsetlinewidth{0.803000pt}%
\definecolor{currentstroke}{rgb}{0.000000,0.000000,0.000000}%
\pgfsetstrokecolor{currentstroke}%
\pgfsetdash{}{0pt}%
\pgfsys@defobject{currentmarker}{\pgfqpoint{0.000000in}{-0.048611in}}{\pgfqpoint{0.000000in}{0.000000in}}{%
\pgfpathmoveto{\pgfqpoint{0.000000in}{0.000000in}}%
\pgfpathlineto{\pgfqpoint{0.000000in}{-0.048611in}}%
\pgfusepath{stroke,fill}%
}%
\begin{pgfscope}%
\pgfsys@transformshift{1.562030in}{0.439824in}%
\pgfsys@useobject{currentmarker}{}%
\end{pgfscope}%
\end{pgfscope}%
\begin{pgfscope}%
\definecolor{textcolor}{rgb}{0.000000,0.000000,0.000000}%
\pgfsetstrokecolor{textcolor}%
\pgfsetfillcolor{textcolor}%
\pgftext[x=1.562030in,y=0.342602in,,top]{\color{textcolor}\rmfamily\fontsize{9.000000}{10.800000}\selectfont \(\displaystyle {0.028}\)}%
\end{pgfscope}%
\begin{pgfscope}%
\definecolor{textcolor}{rgb}{0.000000,0.000000,0.000000}%
\pgfsetstrokecolor{textcolor}%
\pgfsetfillcolor{textcolor}%
\pgftext[x=1.094672in,y=0.176046in,,top]{\color{textcolor}\rmfamily\fontsize{9.000000}{10.800000}\selectfont pend. pole mass \(\displaystyle m_p\)}%
\end{pgfscope}%
\begin{pgfscope}%
\pgfsetbuttcap%
\pgfsetroundjoin%
\definecolor{currentfill}{rgb}{0.000000,0.000000,0.000000}%
\pgfsetfillcolor{currentfill}%
\pgfsetlinewidth{0.803000pt}%
\definecolor{currentstroke}{rgb}{0.000000,0.000000,0.000000}%
\pgfsetstrokecolor{currentstroke}%
\pgfsetdash{}{0pt}%
\pgfsys@defobject{currentmarker}{\pgfqpoint{-0.048611in}{0.000000in}}{\pgfqpoint{-0.000000in}{0.000000in}}{%
\pgfpathmoveto{\pgfqpoint{-0.000000in}{0.000000in}}%
\pgfpathlineto{\pgfqpoint{-0.048611in}{0.000000in}}%
\pgfusepath{stroke,fill}%
}%
\begin{pgfscope}%
\pgfsys@transformshift{0.533841in}{0.557893in}%
\pgfsys@useobject{currentmarker}{}%
\end{pgfscope}%
\end{pgfscope}%
\begin{pgfscope}%
\definecolor{textcolor}{rgb}{0.000000,0.000000,0.000000}%
\pgfsetstrokecolor{textcolor}%
\pgfsetfillcolor{textcolor}%
\pgftext[x=0.208225in, y=0.514518in, left, base]{\color{textcolor}\rmfamily\fontsize{9.000000}{10.800000}\selectfont \(\displaystyle {0.08}\)}%
\end{pgfscope}%
\begin{pgfscope}%
\pgfsetbuttcap%
\pgfsetroundjoin%
\definecolor{currentfill}{rgb}{0.000000,0.000000,0.000000}%
\pgfsetfillcolor{currentfill}%
\pgfsetlinewidth{0.803000pt}%
\definecolor{currentstroke}{rgb}{0.000000,0.000000,0.000000}%
\pgfsetstrokecolor{currentstroke}%
\pgfsetdash{}{0pt}%
\pgfsys@defobject{currentmarker}{\pgfqpoint{-0.048611in}{0.000000in}}{\pgfqpoint{-0.000000in}{0.000000in}}{%
\pgfpathmoveto{\pgfqpoint{-0.000000in}{0.000000in}}%
\pgfpathlineto{\pgfqpoint{-0.048611in}{0.000000in}}%
\pgfusepath{stroke,fill}%
}%
\begin{pgfscope}%
\pgfsys@transformshift{0.533841in}{0.853067in}%
\pgfsys@useobject{currentmarker}{}%
\end{pgfscope}%
\end{pgfscope}%
\begin{pgfscope}%
\definecolor{textcolor}{rgb}{0.000000,0.000000,0.000000}%
\pgfsetstrokecolor{textcolor}%
\pgfsetfillcolor{textcolor}%
\pgftext[x=0.208225in, y=0.809692in, left, base]{\color{textcolor}\rmfamily\fontsize{9.000000}{10.800000}\selectfont \(\displaystyle {0.09}\)}%
\end{pgfscope}%
\begin{pgfscope}%
\pgfsetbuttcap%
\pgfsetroundjoin%
\definecolor{currentfill}{rgb}{0.000000,0.000000,0.000000}%
\pgfsetfillcolor{currentfill}%
\pgfsetlinewidth{0.803000pt}%
\definecolor{currentstroke}{rgb}{0.000000,0.000000,0.000000}%
\pgfsetstrokecolor{currentstroke}%
\pgfsetdash{}{0pt}%
\pgfsys@defobject{currentmarker}{\pgfqpoint{-0.048611in}{0.000000in}}{\pgfqpoint{-0.000000in}{0.000000in}}{%
\pgfpathmoveto{\pgfqpoint{-0.000000in}{0.000000in}}%
\pgfpathlineto{\pgfqpoint{-0.048611in}{0.000000in}}%
\pgfusepath{stroke,fill}%
}%
\begin{pgfscope}%
\pgfsys@transformshift{0.533841in}{1.148241in}%
\pgfsys@useobject{currentmarker}{}%
\end{pgfscope}%
\end{pgfscope}%
\begin{pgfscope}%
\definecolor{textcolor}{rgb}{0.000000,0.000000,0.000000}%
\pgfsetstrokecolor{textcolor}%
\pgfsetfillcolor{textcolor}%
\pgftext[x=0.208225in, y=1.104866in, left, base]{\color{textcolor}\rmfamily\fontsize{9.000000}{10.800000}\selectfont \(\displaystyle {0.10}\)}%
\end{pgfscope}%
\begin{pgfscope}%
\pgfsetbuttcap%
\pgfsetroundjoin%
\definecolor{currentfill}{rgb}{0.000000,0.000000,0.000000}%
\pgfsetfillcolor{currentfill}%
\pgfsetlinewidth{0.803000pt}%
\definecolor{currentstroke}{rgb}{0.000000,0.000000,0.000000}%
\pgfsetstrokecolor{currentstroke}%
\pgfsetdash{}{0pt}%
\pgfsys@defobject{currentmarker}{\pgfqpoint{-0.048611in}{0.000000in}}{\pgfqpoint{-0.000000in}{0.000000in}}{%
\pgfpathmoveto{\pgfqpoint{-0.000000in}{0.000000in}}%
\pgfpathlineto{\pgfqpoint{-0.048611in}{0.000000in}}%
\pgfusepath{stroke,fill}%
}%
\begin{pgfscope}%
\pgfsys@transformshift{0.533841in}{1.443415in}%
\pgfsys@useobject{currentmarker}{}%
\end{pgfscope}%
\end{pgfscope}%
\begin{pgfscope}%
\definecolor{textcolor}{rgb}{0.000000,0.000000,0.000000}%
\pgfsetstrokecolor{textcolor}%
\pgfsetfillcolor{textcolor}%
\pgftext[x=0.208225in, y=1.400040in, left, base]{\color{textcolor}\rmfamily\fontsize{9.000000}{10.800000}\selectfont \(\displaystyle {0.11}\)}%
\end{pgfscope}%
\begin{pgfscope}%
\definecolor{textcolor}{rgb}{0.000000,0.000000,0.000000}%
\pgfsetstrokecolor{textcolor}%
\pgfsetfillcolor{textcolor}%
\pgftext[x=0.152670in,y=1.000654in,,bottom,rotate=90.000000]{\color{textcolor}\rmfamily\fontsize{9.000000}{10.800000}\selectfont rot. pole mass \(\displaystyle m_r\)}%
\end{pgfscope}%
\begin{pgfscope}%
\pgfsetrectcap%
\pgfsetmiterjoin%
\pgfsetlinewidth{0.803000pt}%
\definecolor{currentstroke}{rgb}{0.000000,0.000000,0.000000}%
\pgfsetstrokecolor{currentstroke}%
\pgfsetdash{}{0pt}%
\pgfpathmoveto{\pgfqpoint{0.533841in}{0.439824in}}%
\pgfpathlineto{\pgfqpoint{0.533841in}{1.561484in}}%
\pgfusepath{stroke}%
\end{pgfscope}%
\begin{pgfscope}%
\pgfsetrectcap%
\pgfsetmiterjoin%
\pgfsetlinewidth{0.803000pt}%
\definecolor{currentstroke}{rgb}{0.000000,0.000000,0.000000}%
\pgfsetstrokecolor{currentstroke}%
\pgfsetdash{}{0pt}%
\pgfpathmoveto{\pgfqpoint{1.655502in}{0.439824in}}%
\pgfpathlineto{\pgfqpoint{1.655502in}{1.561484in}}%
\pgfusepath{stroke}%
\end{pgfscope}%
\begin{pgfscope}%
\pgfsetrectcap%
\pgfsetmiterjoin%
\pgfsetlinewidth{0.803000pt}%
\definecolor{currentstroke}{rgb}{0.000000,0.000000,0.000000}%
\pgfsetstrokecolor{currentstroke}%
\pgfsetdash{}{0pt}%
\pgfpathmoveto{\pgfqpoint{0.533841in}{0.439824in}}%
\pgfpathlineto{\pgfqpoint{1.655502in}{0.439824in}}%
\pgfusepath{stroke}%
\end{pgfscope}%
\begin{pgfscope}%
\pgfsetrectcap%
\pgfsetmiterjoin%
\pgfsetlinewidth{0.803000pt}%
\definecolor{currentstroke}{rgb}{0.000000,0.000000,0.000000}%
\pgfsetstrokecolor{currentstroke}%
\pgfsetdash{}{0pt}%
\pgfpathmoveto{\pgfqpoint{0.533841in}{1.561484in}}%
\pgfpathlineto{\pgfqpoint{1.655502in}{1.561484in}}%
\pgfusepath{stroke}%
\end{pgfscope}%
\end{pgfpicture}%
\makeatother%
\endgroup%

%% file: tables/experiment_hparam_QQ.tex
\rowcolors{2}{gray!25}{white}
\begin{tabular}{ll}
	\rowcolor{gray!50}
	\textbf{Hyper-parameter}                 & \textbf{Value}\\
	\multicolumn{2}{c}{\textbf{common}}\\
	\texttt{PolOpt}                          & \acs{PPO}\\
	policy / critic architecture             & \acs{FNN} \num{64}-\num{64} with tan-h\\
	optimizer                                & Adam\\
	learning rate policy                     & \num{5.97e-4}\\
	learning rate critic                     & \num{3.44e-4}\\
    \acs{PPO} clipping ratio $\epsilon$      & \num{0.1}\\
	iterations $n_{\text{iter}}$             & \num{300}\\
   	step size $\Delta t$                     & \SI{0.01}{\second}\\
	max. steps per episode $T$               & \num{600}\\
	min. steps per iteration                 & \num{20}$T$\\
	temporal discount $\gamma$               & \num{0.9885}\\
	adv. est. trade-off factor $\lambda$     & \num{0.965}\\
	success threshold $\tholdsucc$           & \num{\tholdsuccQQ} \\
	$\fQ$                                    & $\textrm{diag} (\num{2e-1}, \num{1.}, \num{2e-2}, \num{5e-3})$ \\
	$R$                                      & $\num{3e-3}$\\
    real-world rollouts $n_\tau$             & \num{5}\\
    \multicolumn{2}{c}{\textbf{\acs{UDR} specific}}\\
    min. steps per iteration                 & \num{30}$T$\\
    \multicolumn{2}{c}{\textbf{SimOpt specific}}\\
    max. iterations $n_{\text{iter}}$    & \num{15}\\
    \texttt{DistrOpt} population size        & \num{500}\\
    \texttt{DistrOpt} \acs{KL} bound $\epsilon$ & \num{1.0}\\
    \texttt{DistrOpt} learning rate          & \num{5e-4}\\
    \multicolumn{2}{c}{\textbf{\acs{BayRn} specific}}\\
    max. iterations $n_{\text{iter,max}}$    & \num{15}\\
    initial solutions $\ninit$               & \num{5}\\
\end{tabular}

%% file: tables/experiment_hparam_WAM.tex
\rowcolors{2}{gray!25}{white}
\begin{tabular}{ll}
	\rowcolor{gray!50}
	\textbf{Hyper-parameter}                 & \textbf{Value}\\
	\multicolumn{2}{c}{\textbf{common}}\\
	\texttt{PolOpt}                          & \acs{PoWER}\\
	policy architecture                      & \acs{RBF} with 16 basis functions\\
	iterations $n_{\text{iter}}$             & \num{20}\\
	population size $n_{\text{pop}}$         & \num{100}\\
	num. importance samples $n_{\text{is}}$  & \num{10}\\
	init. exploration std $\sigma_{\text{init}}$  & $\pi/12$\\
	min. rollouts per iteration              & \num{20}\\
	max. steps per episode $T$               & \num{1750}\\
   	step size $\Delta t$                     & \SI{0.002}{\second}\\
	temporal discount $\gamma$               & \num{1}\\
    real-world rollouts $n_\tau$             & \num{5}\\
    \multicolumn{2}{c}{\textbf{\acs{UDR} specific}}\\
    min. steps per iteration                 & \num{30}$T$\\
    \multicolumn{2}{c}{\textbf{\acs{BayRn} specific}}\\
    max. iterations $n_{\text{iter}}$    & \num{15}\\
    initial solutions $\ninit$               & \num{5}\\
\end{tabular}